\documentclass[fleqn]{article} 

\usepackage{iclr2018_conference,times}
\usepackage{hyperref}
\usepackage{url}
\usepackage[pdftex]{graphicx}
\usepackage{subfig}
\usepackage{colortbl}
\usepackage{wrapfig}
\usepackage{amsmath}
\usepackage{sidecap}
\usepackage{amssymb}
\usepackage{pifont}
\usepackage[export]{adjustbox}

\newcommand{\figref}[1]{figure~\ref{fig:#1}}

\newcommand{\tabref}[1]{table~\ref{tab:#1}}

\newcommand{\secref}[1]{section~\ref{sec:#1}}
\newcommand{\appref}[1]{appendix~\ref{sec:#1}}

\definecolor{gray}{RGB}{87, 87, 87}
\definecolor{red}{RGB}{173, 35, 35}
\definecolor{blue}{RGB}{42, 75, 215}
\definecolor{green}{RGB}{29, 105, 20}
\definecolor{brown}{RGB}{129, 74, 25}
\definecolor{purple}{RGB}{129, 38, 192}
\definecolor{cyan}{RGB}{41, 208, 208}
\definecolor{yellow}{RGB}{189, 167, 0}
\definecolor{Red}{rgb}{0.68, 0.05, 0.0}
\definecolor{Blue}{rgb}{0.0, 0.0, 0.61}
\definecolor{Blue1}{RGB}{214, 235, 245}
\definecolor{Blue2}{RGB}{235, 245, 250}
\definecolor{lime}{RGB}{60,179,113}

\iclrfinalcopy

\usepackage[font=small,labelfont=bf]{caption}
\title{Compositional Attention Networks \\ for Machine Reasoning}

\author{Drew A. Hudson \\
Department of Computer Science\\
Stanford University\\
\texttt{dorarad@cs.stanford.edu} \\
\And
Christopher D. Manning \\
Department of Computer Science\\
Stanford University\\
\texttt{manning@cs.stanford.edu} \\
}

\begin{document}

\maketitle

\begin{abstract}
We present the MAC network, a novel fully differentiable neural network architecture, designed to facilitate explicit and expressive reasoning. MAC moves away from monolithic black-box neural architectures towards a design that encourages both transparency and versatility. The model approaches problems by decomposing them into a series of attention-based reasoning steps, each performed by a novel recurrent \textbf{M}emory, \textbf{A}ttention, and \textbf{C}omposition (MAC) cell that maintains a separation between control and memory. By stringing the cells together and imposing structural constraints that regulate their interaction, MAC effectively learns to perform iterative reasoning processes that are directly inferred from the data in an end-to-end approach. We demonstrate the model's strength, robustness and interpretability on the challenging CLEVR dataset for visual reasoning, achieving a new state-of-the-art 98.9\% accuracy, halving the error rate of the previous best model. More importantly, we show that the model is computationally-efficient and data-efficient, in particular requiring 5x less data than existing models to achieve strong results. 
\end{abstract}

\setlength{\textfloatsep}{12pt}

\section{Introduction}
Reasoning, the ability to manipulate previously acquired knowledge to draw novel inferences or answer new questions, is one of the fundamental building blocks of the intelligent mind. As we seek to advance neural networks beyond their current great success with sensory perception  towards tasks that require more deliberate thinking, conferring them with the ability to move from facts to conclusions is thus of crucial importance. To this end, we consider here how best to design a neural network to perform the structured and iterative reasoning necessary for complex problem solving.

\begin{wrapfigure}[14]{r}{0.28\textwidth}
{
\centering
\vspace{-4mm}
\includegraphics[width=1.0\linewidth]{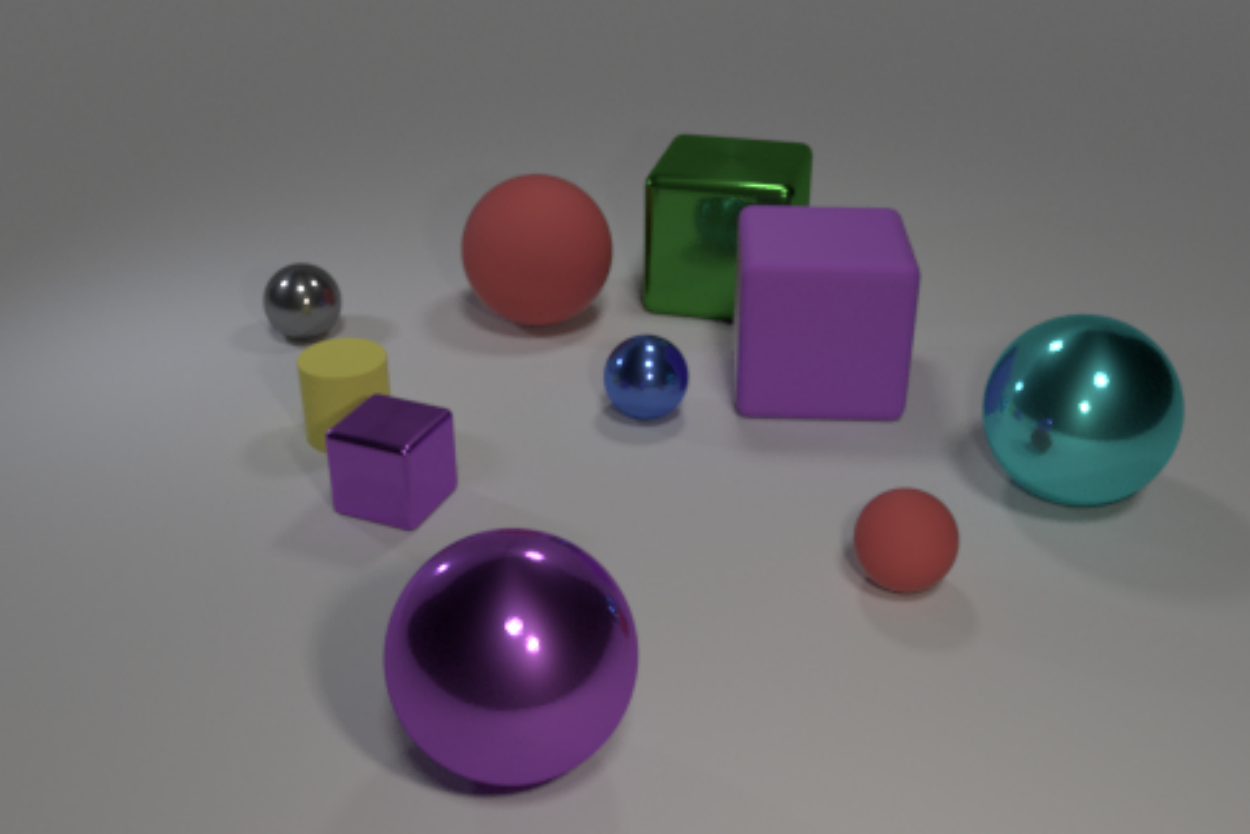}
}
\footnotesize 
\textit{
{{\normalfont\textbf{Q:}} Do \textbf{\textcolor{purple}{the block}} in \textbf{front} of \textbf{\textcolor{yellow}{the tiny yellow cylinder}} and \textbf{\textcolor{red}{the tiny thing}} that is to the \textbf{right} of \textbf{\textcolor{green}{the large green shiny object}} have the \textbf{same color}? {\normalfont\textbf{A:}} No}
}
\vspace*{-1mm}
\caption{A CLEVR example. Color added for illustration.} 
\label{fig:clevrPhoto}
\end{wrapfigure}

Concretely, we develop a novel model that we apply to the CLEVR task \citep{clevr} of visual question answering (VQA). VQA \citep{vqa,vqaSurv} is a challenging multimodal task that requires responding to natural language questions about images. However, \citet{clevr1} show how the first generation of successful VQA models tends to acquire only superficial comprehension of both the image and the question, exploiting dataset biases rather than capturing a sound perception and reasoning process that would lead to the correct answer (cf. \citet{clevr35}). CLEVR was created to address this problem. As illustrated in \figref{clevrPhoto}, the dataset features unbiased, highly compositional questions that require an array of challenging reasoning skills, such as transitive and logical relations, counting and comparisons, without allowing any shortcuts around such reasoning.  

However, deep learning approaches often struggle to perform well on tasks with a compositional and structured nature \citep{rl7, rl23}. Most neural networks are essentially very large correlation engines that will hone in on any statistical, potentially spurious pattern that allows them to model the observed data more accurately. The depth, size and statistical nature that allows them to cope with noisy and diverse data often limits their interpretability and hinders their capacity to perform explicit and sound inference procedures that are vital for problem solving tasks. To mitigate this issue, some recent approaches adopt symbolic structures, resembling the expression trees of programming languages, that compose neural \textit{modules} from a fixed predefined collection \citep{nmn,pgee}. However, they consequently rely on externally provided structured representations and functional programs, brittle handcrafted parsers or expert demonstrations, and require relatively complex multi-stage reinforcement learning training schemes. The rigidity of these models' structure and the use of an inventory of specialized operation-specific modules ultimately undermines their robustness and generalization capacities. 

\begin{SCfigure}

\centering
\vspace*{0mm}
\includegraphics[width=0.59\linewidth]{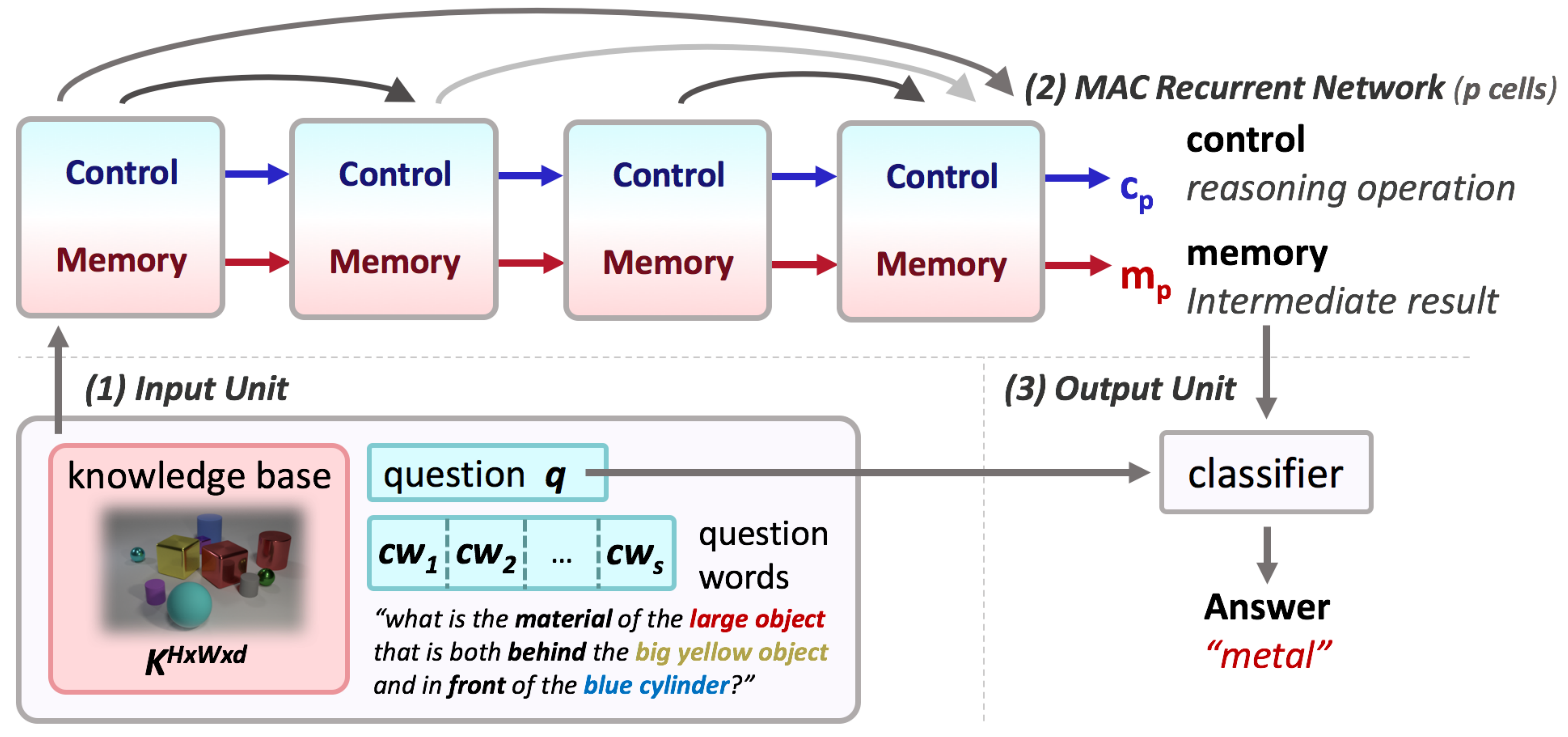}

\hspace*{-3.5mm}
\caption{\textbf{Model Overview.} The MAC network consists of an input unit, a core recurrent network and an output unit. (1) The input unit transforms the raw image and question into distributed vector representations. (2) The core recurrent network reasons sequentially over the question by decomposing it into a series of operations (\textit{control}) that retrieve information from the image (knowledge base) and aggregate the results into a recurrent \textit{memory}. (3) The output classifier computes the final answer using the question and the final memory state.}

\label{fig:model}
\end{SCfigure}

Seeking a balance between the versatility and robustness of end-to-end neural approaches on the one hand and the need to support more explicit and structured reasoning on the other, we propose the MAC network, a novel fully differentiable architecture for reasoning tasks. Our model performs structured and explicit reasoning by sequencing a new recurrent \textbf{Memory, Attention and Composition} (MAC) cell. The MAC cell was deliberately designed to capture the inner workings of an elementary, yet general-purpose reasoning step, drawing inspiration from the design principles of computer architectures. The cell explicitly separates out memory from control, both represented recurrently, and consists of three operational units that work in tandem to perform a reasoning step: the control unit updates the control state to attend at each iteration to some aspect of a given question; the read unit extracts information out of a knowledge base, guided by the control and memory states; and the write unit integrates the retrieved information into the memory state, iteratively computing the answer. This universal design of the MAC cell serves as a structural prior that encourages the network to solve problems by decomposing them into a sequence of attention-based reasoning operations that are directly inferred from the data, without resorting to any strong supervision. With self-attention connections between the cells, the MAC network is capable of representing arbitrarily complex acyclic reasoning graphs in a soft manner, while still featuring a physically sequential structure and end-to-end differentiabillity, amenable to training simply by backpropagation.

We demonstrate the model's quantitative and qualitative performance on the CLEVR task and its associated datasets. The model achieves state-of-the-art accuracy across a variety of reasoning tasks and settings, both for the primary dataset as well as the more difficult human-authored questions. Notably, it performs particularly well on questions that involve counting and aggregation skills, which tend to be remarkably challenging for other VQA models \citep{rn, nmn3, pgee}. Moreover, we show that the MAC network learns rapidly and generalizes effectively from an order of magnitude less data than other approaches. Finally, extensive ablation studies and error analysis demonstrate MAC's robustness, versatility and generalization capacity. These results highlight the significance and value of imposing strong structural priors to guide the network towards compositional reasoning.
The model contains structures that encourage it to explicitly perform a chain of operations that build upon each other, allowing MAC to develop reasoning skills from the ground up, realizing the vision of an algebraic, compositional model of inference as proposed by \citet{bottou}. Although each cell's functionality has only a limited range of possible continuous behaviors, geared to perform a simple reasoning operation, when chained together in a MAC network, the whole system becomes expressive and powerful. TensorFlow implementation of the model is available at https://github.com/stanfordnlp/mac-network.

\section{The MAC Network}
\label{sec:model}

\begin{figure}[t]

\centering
\includegraphics[width=0.4\linewidth]{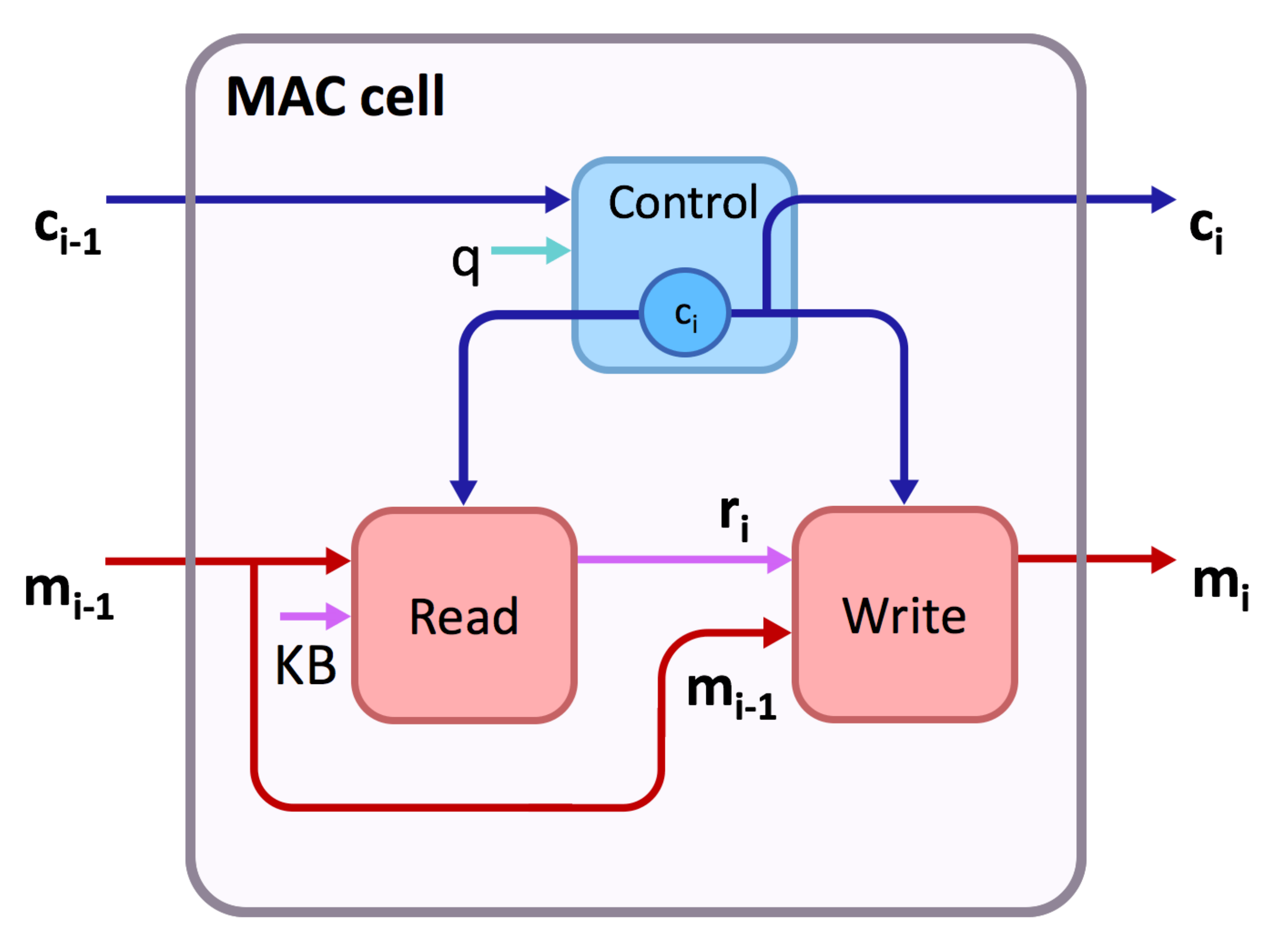}

\caption{\textbf{The MAC cell architecture.} The MAC recurrent cell consists of a control unit, read unit, and write unit, that operate over dual \textit{\textbf{\textcolor{Blue}{control}}} and \textit{\textbf{\textcolor{Red}{memory}}} hidden states. The \textbf{control unit} successively attends to different parts of the task description (question), updating the control state to represent at each timestep the reasoning operation the cell intends to perform. The \textbf{read unit} extracts information out of a knowledge base (here, image), guided by the control state. The  \textbf{write unit} integrates the retrieved information into the memory state, yielding the new intermediate result that follows from applying the current reasoning operation.}
\label{fig:mac}
\end{figure}

A MAC network is an end-to-end differentiable architecture primed to perform an explicit multi-step reasoning process, by stringing together \(p\) recurrent MAC cells, each responsible for performing one reasoning step. Given a knowledge base \(K\) (for VQA, an image) and a task description \(q\) (for VQA, a question), the model infers a decomposition into a series of \(p\) reasoning operations that interact with the knowledge base, iteratively aggregating and manipulating information to perform the task at hand. It consists of three components: (1) an input unit, (2) the core recurrent network, composed out of \(p\) MAC cells, and (3) an output unit, all described below.

\subsection{The Input Unit}

The input unit transforms the raw inputs given to the model into distributed vector representations. Naturally, this unit is tied to the specifics of the task we seek to perform. For the particular case of VQA, it receives a  question and an image and processes each of them respectively: 

\textbf{The question} string, of length \(S\), is converted into a sequence of learned word embeddings that is further processed by a \(d\)-dimensional biLSTM yielding: (1) \textit{contextual words}: a series of output states \({{\boldsymbol{cw}_1},\ldots,{\boldsymbol{cw}_S}}\) that represent each word in the context of the question, and (2) the question representation: \({q} = \left[ {\overleftarrow {\boldsymbol{cw}_1}} ,\overrightarrow {\boldsymbol{cw}_S} \right] \), the concatenation of the final hidden states from the backward and forward LSTM passes. Subsequently, for each step \(i=1,\ldots,p\), the question \({q}\) is transformed through a learned linear transformation into a position-aware vector \(\boldsymbol{q_i} ={{W}_{i}^{d\times 2d}}{q} + {{b}_{i}^d} \), representing the aspects of the question that are relevant to the \(i^{th}\) reasoning step.

\textbf{The image} is first processed by a fixed feature extractor pre-trained on ImageNet \citep{imagenet} that outputs \textit{conv4} features from ResNet101 \citep{imp1}, matching prior work for CLEVR \citep{nmn3,rn,film}. The resulting tensor is then passed through two CNN layers with \(d\) output channels to obtain a final image representation, the \textit{knowledge base} \(\boldsymbol{K}^{H{\times}W{\times}d}={\{\boldsymbol{k_{h,w}^d}|_{h,w=1,1}^{H,W}\}}\), where \(H = W = 14 \) are the height and width of the processed image, corresponding to each of its regions.

\subsection{The MAC cell}
\label{sec:MACcell}

The MAC cell is a recurrent cell designed to capture the notion of an atomic and universal reasoning operation and formulate its mechanics. For each step \(i=1,\ldots,p\) in the reasoning process, the \(i^{th}\) cell maintains dual hidden states: control \(\boldsymbol{c_i}\) and memory \(\boldsymbol{m_i}\), of dimension \(d\), initialized to learned parameters \(\boldsymbol{m_0}\) and \(\boldsymbol{c_0}\), respectively.

\textbf{The control} \(\boldsymbol{c_i}\) represents the \textit{reasoning operation} the cell should accomplish in the \(i^{th}\) step, selectively focusing on some aspect of the question. Concretely, it is represented by a soft attention-based weighted average of the question words \({\boldsymbol{cw}_s}; s=1,\ldots,S\).

\textbf{The memory} \(\boldsymbol{m_i}\) holds the \textit{intermediate result} obtained from the reasoning process up to the \(i^{th}\) step, computed recurrently by integrating the preceding hidden state \(\boldsymbol{m_{i-1}}\)
with new information \(\boldsymbol{r_{i}}\) retrieved from the image, performing the \(i^{th}\) reasoning operation \(\boldsymbol{c_i}\). Analogously to the control, \(\boldsymbol{r_{i}}\) is  a weighted average over its regions \(\{\boldsymbol{k_{h,w}}|_{h,w=1,1}^{H,W}\}\).

Building on the design principles of computer organization, the MAC cell consists of three operational units: control unit CU, read unit RU and write unit WU, that work together to accomplish tasks by performing an iterative reasoning process: The control unit identifies a series of operations, represented by a recurrent control state; the read unit extracts relevant information from a given knowledge base to perform each operation, and the write unit iteratively integrates the information into the cell's memory state, producing a new intermediate result.

Through their operation, the three units together impose an interface that regulates the interaction between the control and memory states. Specifically, the control state, which is a function of the question, guides the integration of content from the image into the memory state only through indirect means: soft-attention maps and sigmoidal gating mechanisms. Consequently, the interaction between these two modalities -- visual and textual, or knowledge base and query -- is mediated through probability distributions only. This stands in stark contrast to common approaches that fuse the question and image together into the same vector space through linear combinations, multiplication, or concatenation. As we will see in \secref{experiments}, maintaining a strict separation between the representational spaces of question and image, which can interact only through interpretable discrete distributions, greatly enhances the generalizability of the network and improves its transparency.

In the following, we describe the cell's three components: control, read and write units, and detail their formal specification. Unless otherwise stated, all the vectors are of dimension \(d\). 

\subsubsection{The Control Unit}
\label{sec:CU}

\begin{figure}[t]
\centering

\begin{minipage}{0.59\textwidth}
\noindent
\vspace*{-8pt}
\includegraphics[valign = c, width=1.0\linewidth]{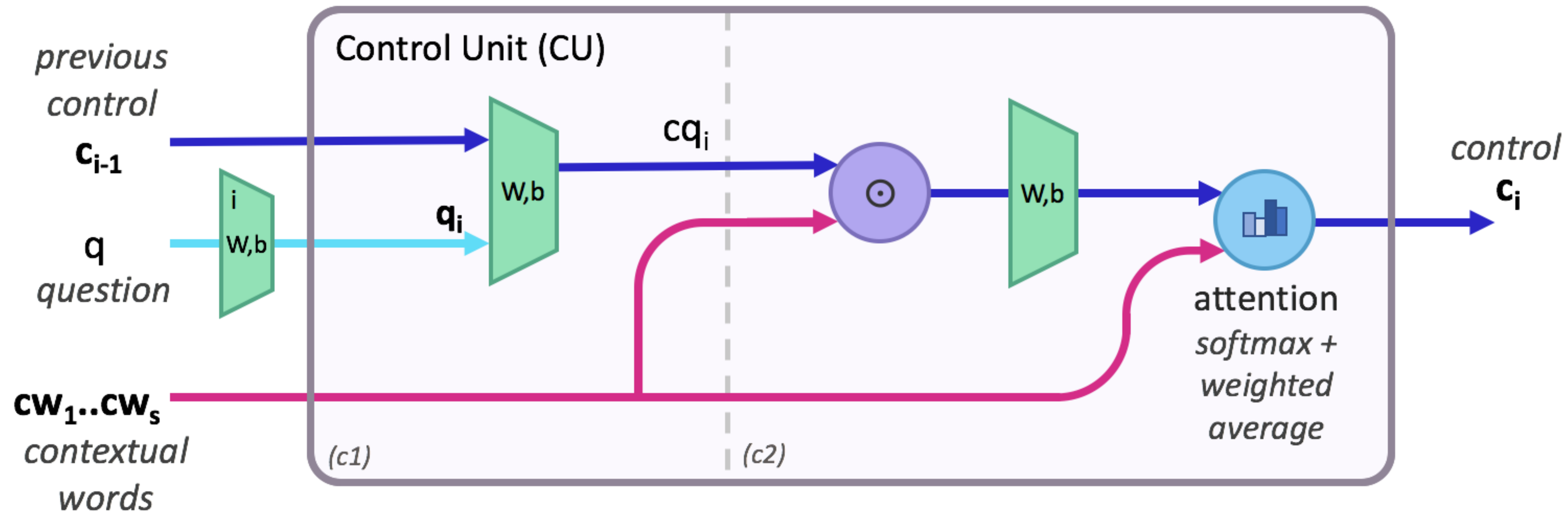}
\end{minipage}
\hspace*{-35pt}
\begin{minipage}{0.48\textwidth}
\noindent
\footnotesize
\begin{subequations}
\begin{gather}
{\mathit{cq}_i} = {W}^{d\times 2d}\left[\boldsymbol{c_{i-1}},\boldsymbol{q_i}\right] + {b}^d\tag{c1} \\[2pt]
{\mathit{ca}_{i,s}} = W^{1\times d}({\mathit{cq}_i} \odot \boldsymbol{\mathit{cw}_s}) + b^1\tag{c2.1} \\[5pt]
{\mathit{cv}_{i,s}} = \textrm{softmax}({{ca}_{i,s}})\tag{c2.2} \\[-0.5pt]
\boldsymbol{c_{i}} = \sum\limits_{s = 1}^S {\mathit{cv}_{i,s}}  \cdot {\boldsymbol{\mathit{cw}}_s}\tag{c2.3}
\end{gather}
\end{subequations}
\end{minipage}

\caption{\textbf{The Control Unit (CU) architecture.} The control unit attends at each iteration to some part of the question, by applying soft attention over the question words, and updates the control state accordingly. The unit's inputs and outputs are in \textbf{bold}. See \secref{CU} for details.}
\label{fig:control}
\end{figure}

The control unit (see \figref{control}) determines the reasoning operation that should be performed at each step \(i\), attending to some part of the question and updating the control state \(\boldsymbol{c_i}\) accordingly. It receives the contextual question words \({\boldsymbol{\mathit{cw}}_1,\ldots,\boldsymbol{\mathit{cw}}_S}\), the question position-aware representation \(\boldsymbol{q_i}\), and the control state from the preceding step \(\boldsymbol{c_{i-1}}\) and consists of two stages:

\begin{enumerate}
  
\item First, we combine \(\boldsymbol{q_i}\) and  \(\boldsymbol{c_{i-1}}\) through a linear transformation into \({{cq}_i}\), taking into account both the overall question representation \(\boldsymbol{q_i}\), biased towards the \(i^{th}\) reasoning step, as well as the preceding reasoning operation \(\boldsymbol{c_{i-1}}\). This allows the cell to base its decision for the \(i^{th}\) reasoning operation \(\boldsymbol{c_i}\) on the previously performed operation \(\boldsymbol{c_{i-1}}\).
\item Subsequently, we \textit{cast} \({\mathit{cq}_i}\) onto the space of the question words. Specifically, this is achieved by measuring the similarity between \({{cq}_i}\) and each question word \({\boldsymbol{\mathit{cw}}_s}\) and passing the result through a softmax layer, yielding an attention distribution over the question words \({\boldsymbol{\mathit{cw}}_1,\ldots,\boldsymbol{\mathit{cw}}_S}\). Finally, we sum the words according to this distribution to produce the reasoning operation \(\boldsymbol{c_i}\), represented \textit{in terms of} the question words. 
\end{enumerate}  
The casting of \({\mathit{cq}_i}\) onto question words serves as a form of regularization that restricts the space of the valid reasoning operations by anchoring them back in the original question words, and due to the use of soft attention may also improve the MAC cell transparency, since we can interpret the control state content and the cell's consequent behavior based on the words it attends to.

\subsubsection{The Read Unit}
\label{sec:RU}

\begin{figure}[t]
\centering

\begin{minipage}{0.6\textwidth}
\noindent
\vspace*{0pt}
\hspace*{-12pt}
\includegraphics[valign=c,width=1.0\linewidth]{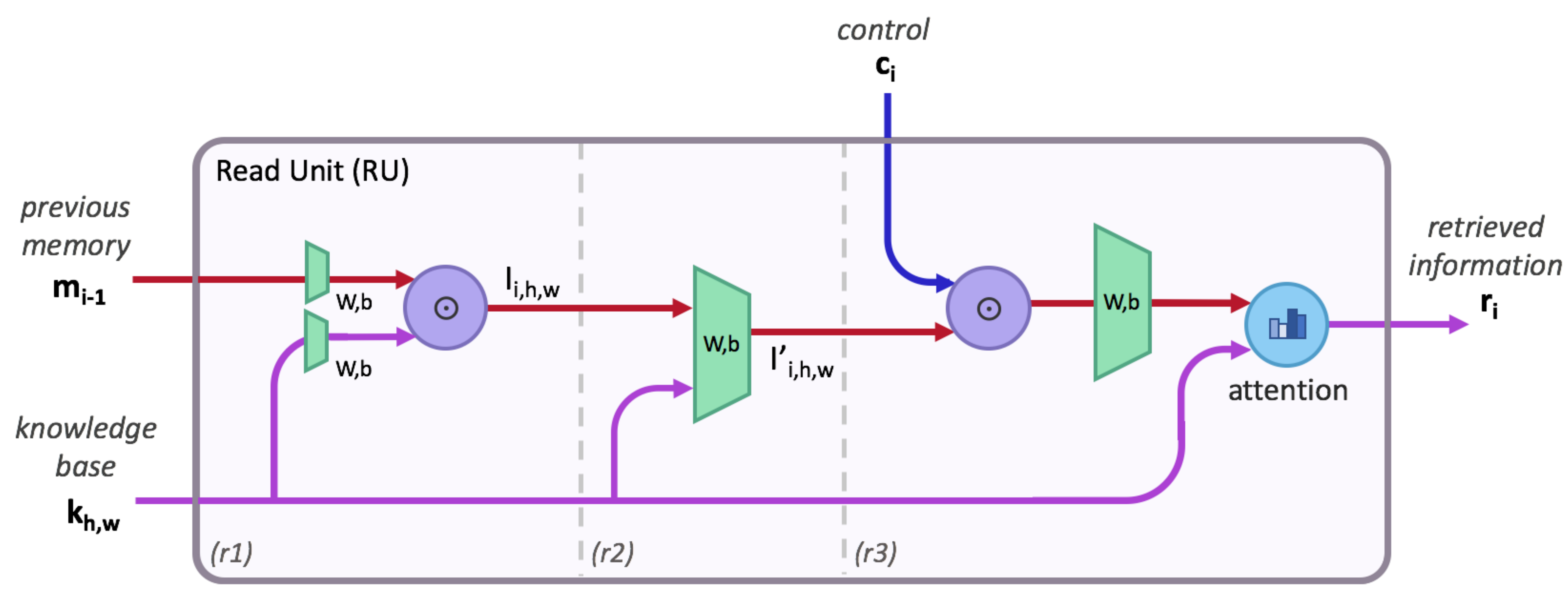}
\end{minipage}
\hspace*{-45pt}
\begin{minipage}{0.5\textwidth}
\noindent
\footnotesize
\begin{subequations}
\begin{gather}
\begin{split}
I_{i,h,w} = & [W_m^{d\times d}\boldsymbol{m_{i-1}} + b_m^d] \; \odot \\[2pt]
& [W_k^{d\times d}{{\boldsymbol{k}_{h,w}}} + b_k^d]
\end{split}\tag{r1} \\[5pt]
{{{I'_{i,h,w}}}} = W^{d\times 2d}\left[ {{I_{i,h,w}}},{\boldsymbol{k}_{h,w}} \right] + b^d\tag{r2} \\[3pt]
{{\mathit{ra}_{i,h,w}}} = W^{d\times d}({\boldsymbol{c_i} \odot {{{I'_{i,h,w}}}}}) + b^d\tag{r3.1}  \\[5pt]
{{\mathit{rv}_{i,h,w}}} = \textrm{softmax}({{\mathit{ra}_{i,h,w}}})\tag{r3.2} \\[-0.5pt]
\boldsymbol{r_i} = \sum\limits_{h,w = 1,1}^{H,W} {\mathit{rv}_{i,h,w}}\tag{r3.3}  \cdot {\boldsymbol{k}_{h,w}}  
\end{gather}
\end{subequations}
\end{minipage}

\caption{\textbf{The Read Unit (RU) architecture.} The read unit retrieves information from the knowledge base that is necessary for performing the current reasoning operation (control) and potentially related to previously obtained intermediate results (memory). It extracts the information by performing a two-stage attention process over the knowledge base elements. See \secref{RU} for details.}

\label{fig:read}
\end{figure}


For the \(i^{th}\) step, the read unit (see \figref{read}) inspects the knowledge base (the image) and retrieves the information \(\boldsymbol{r_i}\) that is required for performing the \(i^{th}\) reasoning operation \(\boldsymbol{c_i}\). The content's relevance is measured by an attention distribution \(\boldsymbol{\mathit{rv}_i}\) that assigns a probability to each element in the knowledge base \({\boldsymbol{k}_{h,w}^d}\), taking into account the current reasoning operation \(\boldsymbol{c_i}\) and the prior memory \(\boldsymbol{m_{i-1}}\), the intermediate result produced by the preceding reasoning step. The attention distribution is computed in several stages:

\begin{enumerate}  
\item First, we compute the direct interaction between the knowledge-base element \({\boldsymbol{k}_{h,w}}\) and the memory \(\boldsymbol{m_{i-1}}\), resulting in \(I_{i,h,w}\). This term measures the relevance of the element to the preceding intermediate result, allowing the model to perform transitive reasoning by considering content that now seems important in light of information obtained from the prior computation step. 

\item Then, we concatenate the element \({\boldsymbol{k}_{h,w}}\) to \(I_{i,h,w}\) and pass the result through a linear transformation, yielding \(I'_{i,h,w}\). This allows us to also consider new information that is \textit{not} directly related to the prior intermediate result, as sometimes a cogent reasoning process has to combine together \textit{independent} facts to arrive at the answer (e.g., for a logical OR operation, set union and counting). 

\item Finally, aiming to retrieve information that is relevant for the reasoning operation \(\boldsymbol{c_i}\), we measure its similarity to each of the interactions \(I_{i,h,w}\) and pass the result through a softmax layer. This produces an attention distribution over the knowledge base elements, which we then use to compute a weighted average over them -- \(\boldsymbol{r_i}\).

\end{enumerate}  

\begin{wrapfigure}[15]{r}{0.32\textwidth}
\vspace*{-8mm}
\begin{minipage}{0.4\linewidth}
\centering
\vspace*{3mm}
\subfloat{\includegraphics[width=1.0\linewidth]{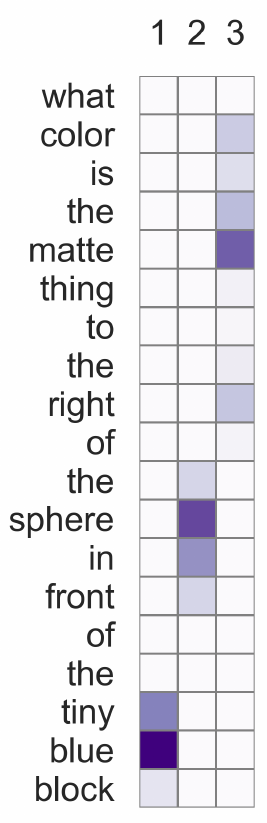}}
\end{minipage}
\begin{minipage}{0.58\linewidth}
\noindent
\centering
\subfloat{\includegraphics[width=1.0\linewidth]{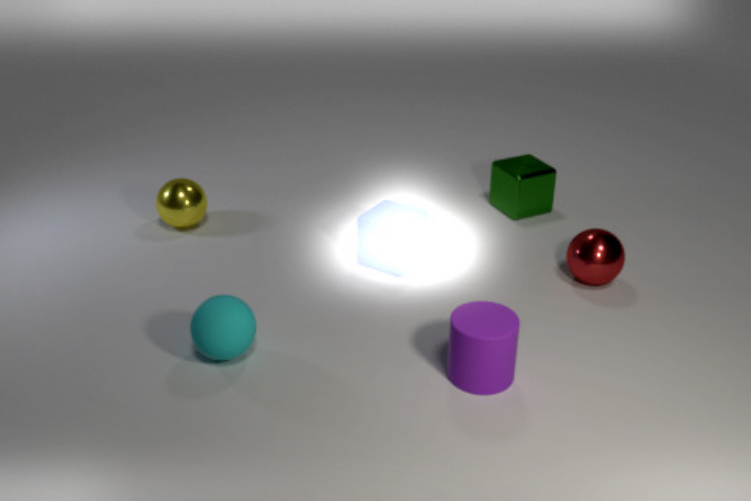}}

\vspace*{-3.5mm}
\subfloat{\includegraphics[width=1.0\linewidth]{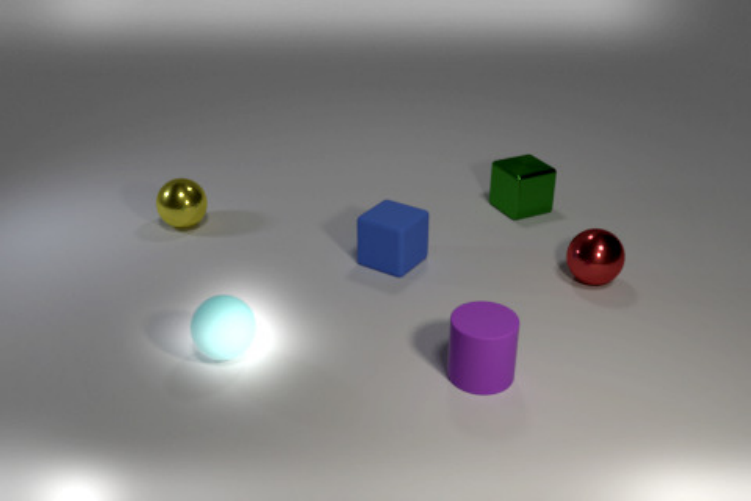}}

\vspace*{-3.5mm}
\subfloat{\includegraphics[width=1.0\linewidth]{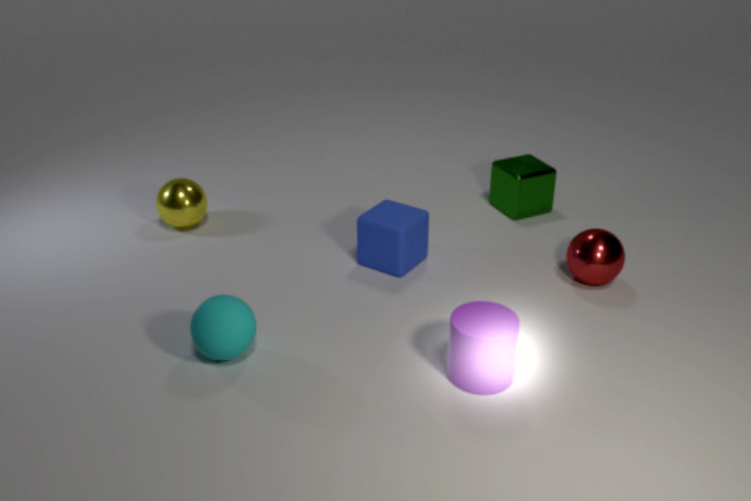}}
\end{minipage}

\scriptsize 

\vspace*{-1mm}
\caption{Attention maps produced by a MAC network of length 3.}
\label{fig:rexample}
\end{wrapfigure}

To give an example of the read unit operation, consider the question in \figref{rexample}, which refers to the purple cylinder in the image. Initially, no cue is provided to the model to attend to the cylinder, since no direct mention of it is given in the question. Instead, the model approaches the question in steps: in the first iteration it attends to the \textcolor{blue}{\textit{\textbf{``tiny blue block"}}}, updating \(\boldsymbol{m_1}\) accordingly to the visual representation of the block. At the following step, the control unit realizes it should now look for \textcolor{cyan}{\textit{\textbf{``the sphere in front"}}} of the block, storing that in \(\boldsymbol{c_2}\). Then, when considering \textit{both} \(\boldsymbol{m_1}\) and \(\boldsymbol{c_2}\), the read unit realizes it should look for ``the sphere in front" (\(\boldsymbol{c_2}\)) of the blue block (stored in \(\boldsymbol{m_1}\)), thus finding the cyan sphere and updating \(\boldsymbol{m_2}\). Finally, a similar process repeats in the next iteration, allowing the model to traverse from the cyan ball to the final objective -- \textcolor{purple}{\textit{\textbf{the purple cylinder}}}, and answer the question correctly.

\newpage
\subsubsection{The Write Unit}
\label{sec:WU}
\begin{figure}[t]
\centering
\begin{minipage}{0.54\textwidth}
\noindent
\vspace*{-4pt}
\hspace*{-16pt}
\includegraphics[width=1.0\linewidth]{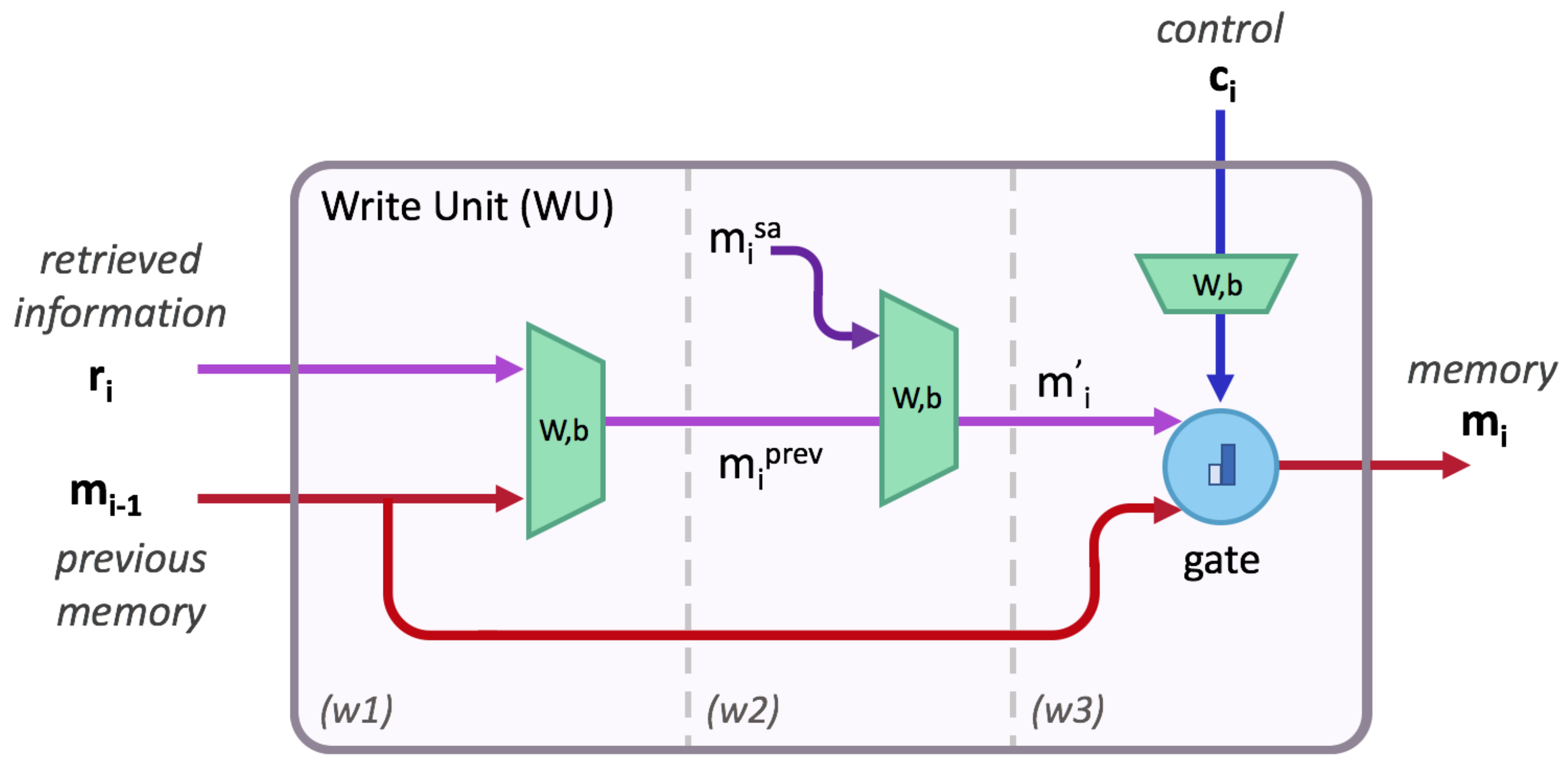}
\end{minipage}
\hspace*{-42pt}
\begin{minipage}{0.55\textwidth}
\noindent
\footnotesize
\begin{subequations}
\begin{gather}
m_i^{info} = {W}^{d\times 2d}[\boldsymbol{r_{i}},\boldsymbol{m_{i-1}}] + {b}^d\tag{w1} \\[1pt]
{\mathit{sa}_{ij}} = \textrm{softmax}\left({W}^{1\times d}{\left( \boldsymbol{c_i} \odot \boldsymbol{c_j} \right)} + {b^1} \right)\tag{w2.1} \\[-3pt]
{m_i^{sa}} = \sum\limits_{j = 1}^{i - 1} {\mathit{sa}_{ij}} \cdot \boldsymbol{{m}_{j}}\tag{w2.2} \\[1pt]
{m'_i} = {W_s^{d\times d}}{m_i^{sa}} + {W_p^{d\times d}} m_i^{info} + b^d\tag{w2.3} \\[4pt]
{c'_i} = W^{1\times d}\boldsymbol{c_i} + b^1\tag{w3.1} \\[2pt]
\boldsymbol{m_i} = \sigma \left( {{c'_i}} \right)  \boldsymbol{m_{i - 1}} + \left( {1 - \sigma \left( {{c'_i}} \right)} \right){m'_i}\tag{w3.2} 
\end{gather}
\end{subequations}
\end{minipage}

\caption{\textbf{The Write Unit (WU) architecture}. The write unit integrates the information retrieved from the knowledge base into the recurrent memory state, producing a new intermediate result \(\boldsymbol{m_i}\) that corresponds to the reasoning operation \(\boldsymbol{c_i}\). See \secref{WU} for details. }\label{fig:write}
\end{figure}

The write unit (see \figref{write}) is responsible for computing the \(i^{th}\) intermediate result of the reasoning process and storing it in the memory state \(\boldsymbol{m_i}\). Specifically, it integrates the information retrieved from the read unit \(\boldsymbol{r_i}\) with the preceding intermediate result \(\boldsymbol{m_{i-1}}\), guided by the \(i^{th}\) reasoning operation \(\boldsymbol{c_i}\). The integration proceeds in three steps, the first mandatory while the others are optional\footnote{Both self-attention connections as well as the memory gate serve to reduce long-term dependencies. However, note that for the CLEVR dataset we were able to maintain almost the same performance with the first step only, and so we propose the second and third ones as optional extensions of the basic write unit, and explore their impact on the model's performance in \secref{ablations}.}:

\begin{enumerate}
\item First, we combine the new information \(\boldsymbol{r_i}\) with the prior intermediate result \(\boldsymbol{m_{i-1}}\) by a linear transformation, resulting in \({m_{i}^{info}}\).

\item \textbf{Self-Attention} (Optional). To support non-sequential reasoning processes, such as trees or graphs, we allow each cell to consider all previous intermediate results, rather than just the preceding one \(\boldsymbol{m_{i-1}}\): We compute the similarity between the \(i^{th}\) operation \(\boldsymbol{c_i}\) and the previous ones \({\boldsymbol{c}_1,\ldots,\boldsymbol{c}_{i-1}}\) and use it to derive an attention distribution over the prior reasoning steps \({\mathit{sa}_{i,j}}\) for \(j = 0,\ldots,{i-1}\). The distribution represents the relevance of each previous step \(j\) to the current one \(i\), and is used to compute a weighted average of the memory states, yielding \({m_{i}^{sa}}\), which is then combined with \({m_{i}^{info}}\) to produce \({m'_{i}}\). Note that while we compute the attention based on the \textit{control states}, we use it to average over the \textit{memory states}, in a way that resembles Key-Value Memory Networks \citep{kvmn}.

\item \textbf{Memory Gate} (Optional). Not all questions are equally complex -- some are simpler while others are more difficult. To allow the model to dynamically adjust the reasoning process length to the given question, we add a sigmoidal gate over the memory state that interpolates between the previous memory state \(\boldsymbol{m_{i-1}}\) and the new candidate \({m'_{i}}\), \textit{conditioned on the reasoning operation \(\boldsymbol{c_{i}}\)}. The gate allows the cell to skip a reasoning step if necessary, passing the previous memory value further along the network, dynamically reducing the effective length of the reasoning process as demanded by the question.

\end{enumerate}

\subsection{The Output Unit}
\label{sec:output}

\begin{wrapfigure}[8]{r}{0.36\textwidth}
\centering
\vspace{-5mm}
\centering
\includegraphics[width=1.0\linewidth]{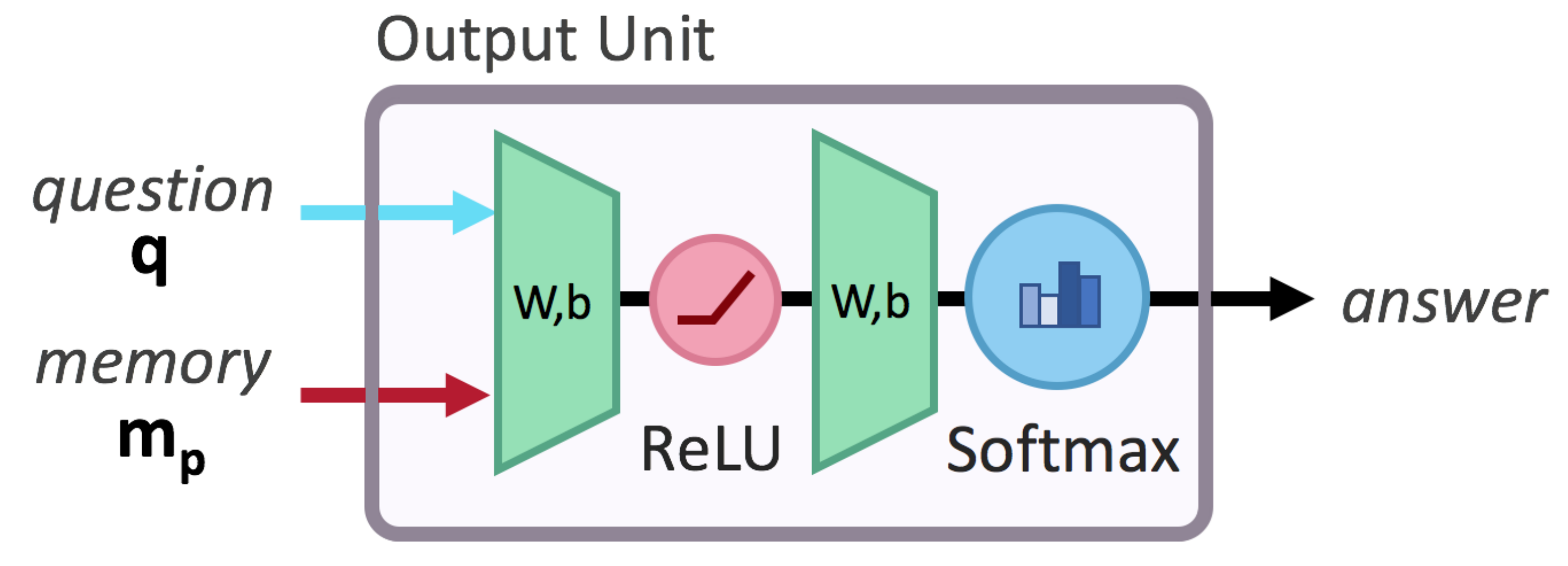}

\vspace{-0.5mm}

\caption{\textbf{The output unit.} A classifier that predicts an answer based on the question and the final memory state.}

\end{wrapfigure}

The output unit predicts the final answer to the question based on the question representation \(q\) and the final memory \(\boldsymbol{m_p}\), which represents the final intermediate result of the reasoning process, holding relevant information from the knowledge base.\footnote{Note that some questions refer to important aspects that do not have counterpart information in the knowledge base, and thus considering both the question and the memory is critical to answer them.} For CLEVR, where there is a fixed set of possible answers, the unit processes the concatenation of \(q\) and \(\boldsymbol{m_p}\) through a 2-layer fully-connected softmax classifier that produces a distribution over the candidate answers.



\section{Related Work}
\label{sec:related}

There have been several prominent models that address the CLEVR task. By and large they can be partitioned into two groups: module networks, which in practice have all used the strong supervision provided in the form of structured functional programs that accompany each data instance, and large, relatively unstructured end-to-end differentiable networks that complement a fairly standard stack of CNNs with components that aid them in performing reasoning tasks. In contrast to modular approaches \citep{nmn,nmn2,nmn3,pgee}, our model is fully differentiable and does not require additional supervision, making use of a single computational cell chained in sequence rather than a collection of custom modules deployed in a rigid tree structure. In contrast to augmented CNN approaches \citep{rn,film}, we suggest that our approach provides an ability for relational reasoning with better generalization capacity, higher computational efficiency and enhanced transparency. These approaches and other related work are discussed and contrasted in
more detail in the supplementary material in \appref{related}.

\section{Experiments}
\label{experiments}
\label{sec:experiments}

\begin{table}
\caption{CLEVR and CLEVR-Humans Accuracy by baseline methods, previous methods, and our method (MAC). For CLEVR-Humans, we show results before and after fine-tuning. (*) denotes use of extra supervisory information through program labels. ($^{\dag}$) denotes use of data augmentation. ($^{\ddag}$) denotes training from raw pixels. }
\label{tab:table1}
\centering
\scriptsize 
\begin{tabular}{l@{}cccccccc}
\rowcolor{Blue1}
\scriptsize Model & CLEVR & Count & Exist & Compare & Query & Compare & Humans & Humans \\ 
\rowcolor{Blue1}
 & Overall            &           &          & Numbers & Attribute & Attribute & before FT & after FT \\
\tiny Human \citep{pgee} & 92.6 & 86.7 & 96.6 & 86.5 & 95.0 & 96.0 & - & - \\
\tiny Q-type baseline \citep{pgee} & 41.8 & 34.6 & 50.2 & 51.0 & 36.0 & 51.3 & - & - \\
\tiny LSTM \citep{pgee} & 46.8 & 41.7 & 61.1 & 69.8 & 36.8 & 51.8 & 27.5 & 36.5 \\
\tiny CNN+LSTM \citep{pgee} & 52.3 & 43.7 & 65.2 & 67.1 & 49.3 & 53.0 & 37.7 & 43.2 \\
\tiny CNN+LSTM+SA+MLP \citep{clevr} & 73.2 & 59.7 & 77.9 & 75.1 & 80.9 & 70.8 & 50.4 & 57.6 \\
\rowcolor{Blue2}
\tiny N2NMN* \citep{nmn3} & 83.7 & 68.5 & 85.7 & 84.9 & 90.0 & 88.7  & - & - \\
\rowcolor{Blue2}
\tiny PG+EE (9K prog.)* \citep{pgee} & 88.6 & 79.7 & 89.7 & 79.1 & 92.6 & 96.0 & - & - \\
\rowcolor{Blue2}
\tiny PG+EE (18K prog.)* \citep{pgee} & 95.4 & 90.1 & 97.3 & 96.5 & 97.4 & 98.0 & 54.0 & 66.6 \\
\rowcolor{Blue2}
\tiny PG+EE (700K prog.)* \citep{pgee} & 96.9 & 92.7 & 97.1 & 98.7 & 98.1 & 98.9 & - & - \\
\tiny CNN+LSTM+RN$^{\dag\ddag}$ \citep{rn} & 95.5 & 90.1 & 97.8 & 93.6 & 97.9 & 97.1 & - & - \\
\tiny CNN+GRU+FiLM \citep{film} & 97.7 & 94.3 & 99.1 & 96.8 & 99.1 & 99.1 & 56.6 & 75.9\\
\tiny CNN+GRU+FiLM$^{\ddag}$ \citep{film} & 97.6 & 94.3 & 99.3 & 93.4 & 99.3 & 99.3 & - & - \\[1ex]
\rowcolor{Blue1}
\textbf{MAC} & \textbf{98.9} & \textbf{97.1} & \textbf{99.5} & \textbf{99.1} & \textbf{99.5} & \textbf{99.5} & \textbf{57.4} & \textbf{81.5}\\
\end{tabular}
\end{table}
 
We evaluate our model on the recent CLEVR task for visual reasoning \citep{clevr}. The dataset consists of rendered images featuring 3D-objects of various shapes, materials, colors and sizes, coupled with machine-generated compositional multi-step questions that measure performance on an array of challenging reasoning skills such as following transitive relations, counting objects and comparing their properties. Each question is also associated with a tree-structured functional program that was used to generate it, specifying the reasoning operations that should be performed to compute the answer. 


In the following experiments, our model's training is cast as a supervised classification problem to minimize the cross-entropy loss of the predicted candidate answer out of the 28 possibilities. The model uses a hidden state size of \(d=512\) and, unless otherwise stated, length of \(p=12\) MAC cells.\footnote{We initialize the word embeddings of our model to random vectors using a uniform distribution. In an earlier version of this work, we used pretrained GloVe vectors, but found that they did not improve the performance for CLEVR and led to only a marginal improvement for CLEVR-Humans.} While some prior work uses the functional programs associated with each question as additional supervisory information at training time (see \tabref{table1}), we intentionally do not use these structured representations to train our model, aiming to infer coherent reasoning strategies directly from the question and answer pairs in an end-to-end approach.

We first perform experiments on the primary 700k dataset. As shown in \tabref{table1}, our model outperforms all prior work both in overall accuracy, as well as in each of the categories of specific reasoning skills. In particular, for the overall performance, we achieve 98.94\% accuracy, more than halving the error rate of the best prior model, FiLM \citep{film}.

\textbf{Counting and Numerical Comparison.} In particular, our performance on questions about counting and numerical comparisons is significantly higher than existing models, which consistently struggle on these question types. Again, we nearly halve the corresponding error rate. These are significant results, as counting and aggregations are known to be particularly challenging in the area of VQA \citep{counting}. In contrast to CNNs, using attention enhances our model's ability to perform reasoning operations such as counting that pertain to the global aggregation of information across different regions of the image. 

\subsection{CLEVR Humans and Error Analysis}

\begin{figure}
\centering
\subfloat{\includegraphics[width=0.25\linewidth]{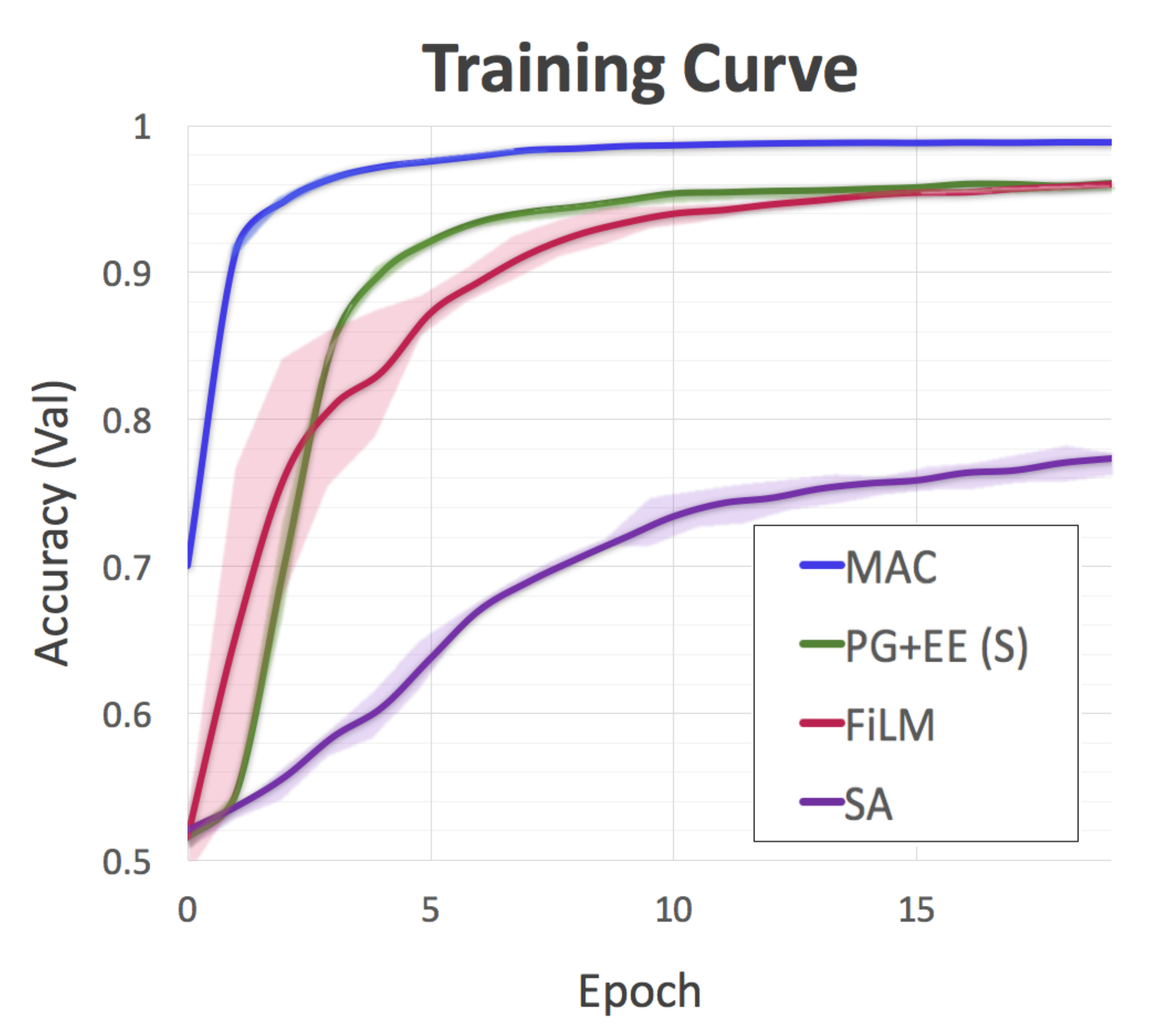}}
\hfill
\subfloat{\includegraphics[width=0.25\linewidth]{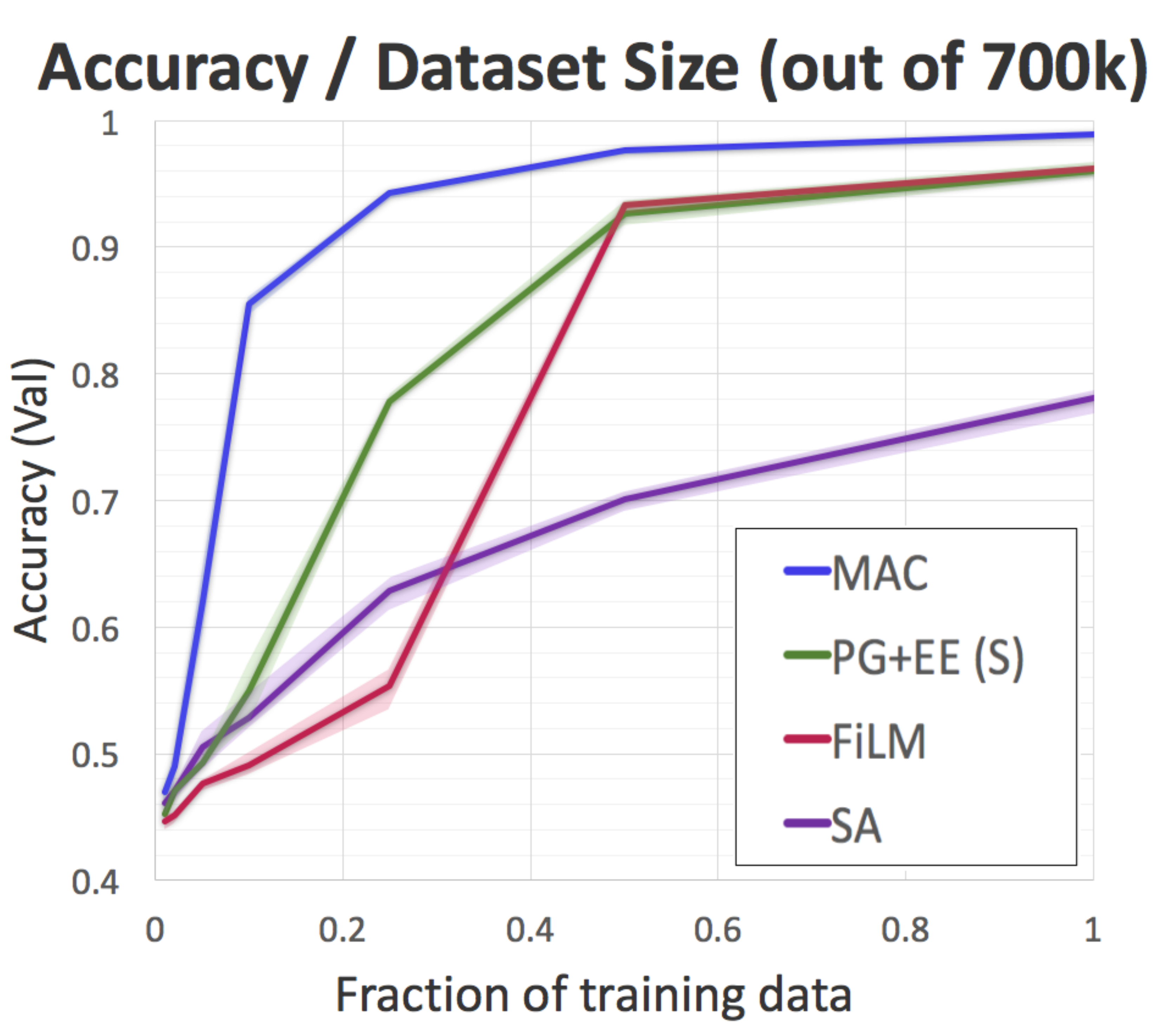}}
\hfill
\subfloat{\includegraphics[width=0.25\linewidth]{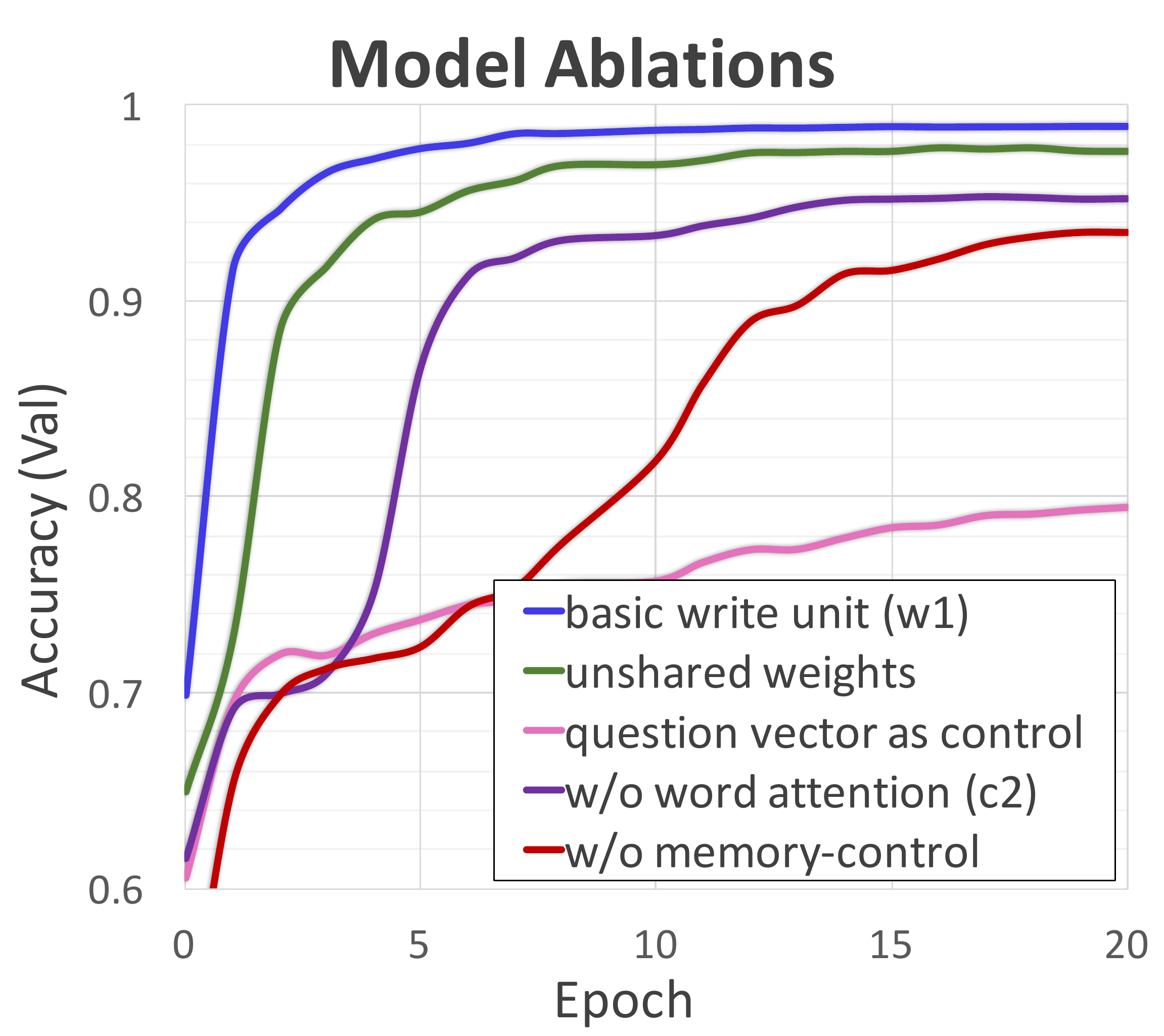}}
\hfill
\subfloat{\includegraphics[width=0.25\linewidth]{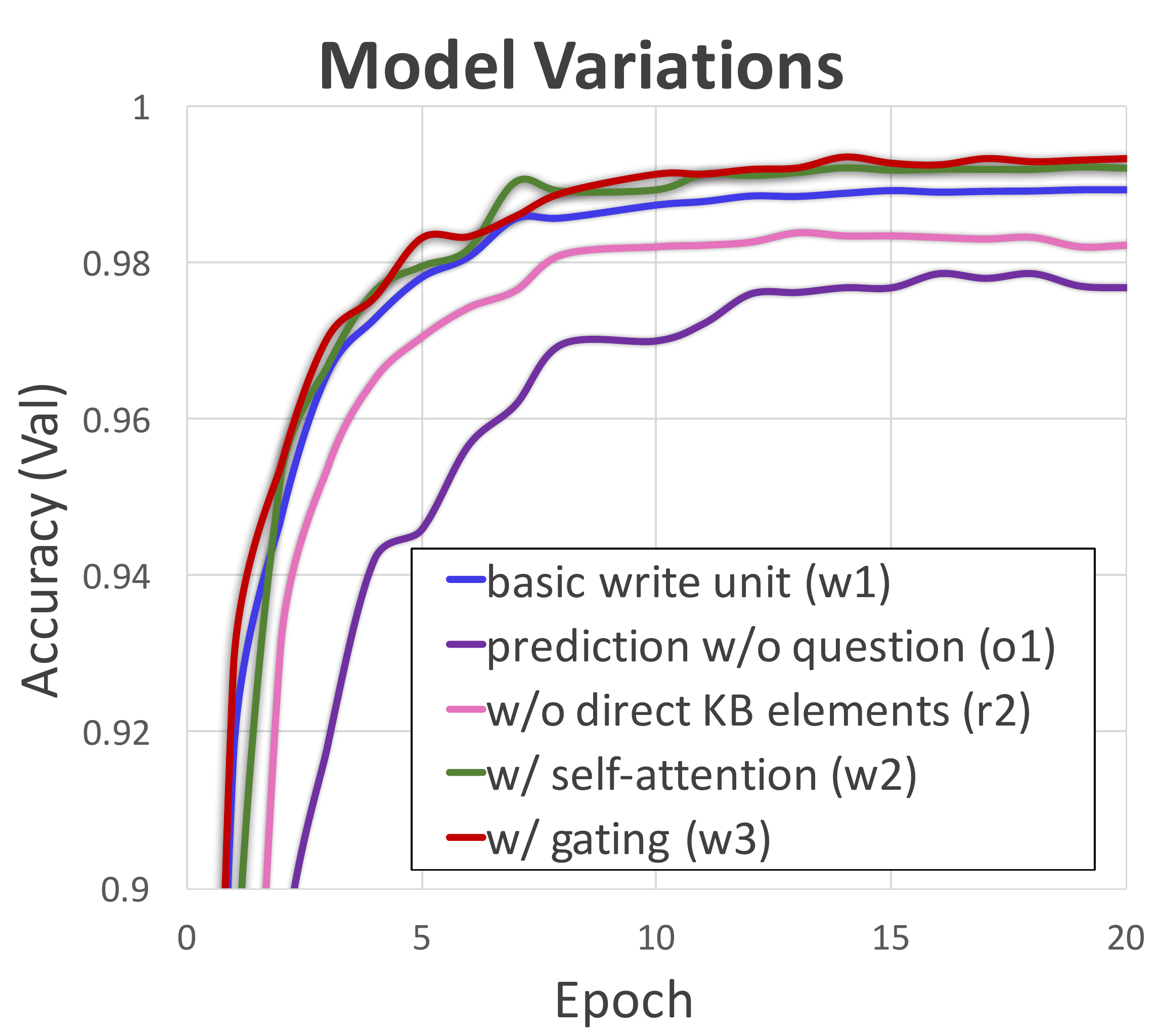}}
\caption{From left to right: (1) Learning curve of MAC and alternative approaches (accuracy / epoch). (2) Models' performance as a function of the CLEVR subset size used for training, ranging from 1\% to 100\%. (3),(4) Learning curves for ablated MAC variants. See \secref{ablations} for details. }
\label{fig:plots}
\end{figure}

\begin{wrapfigure}[19]{r}{0.245\textwidth}
\centering
\vspace{-8mm}
\centering
\subfloat{\includegraphics[width=1.0\linewidth]{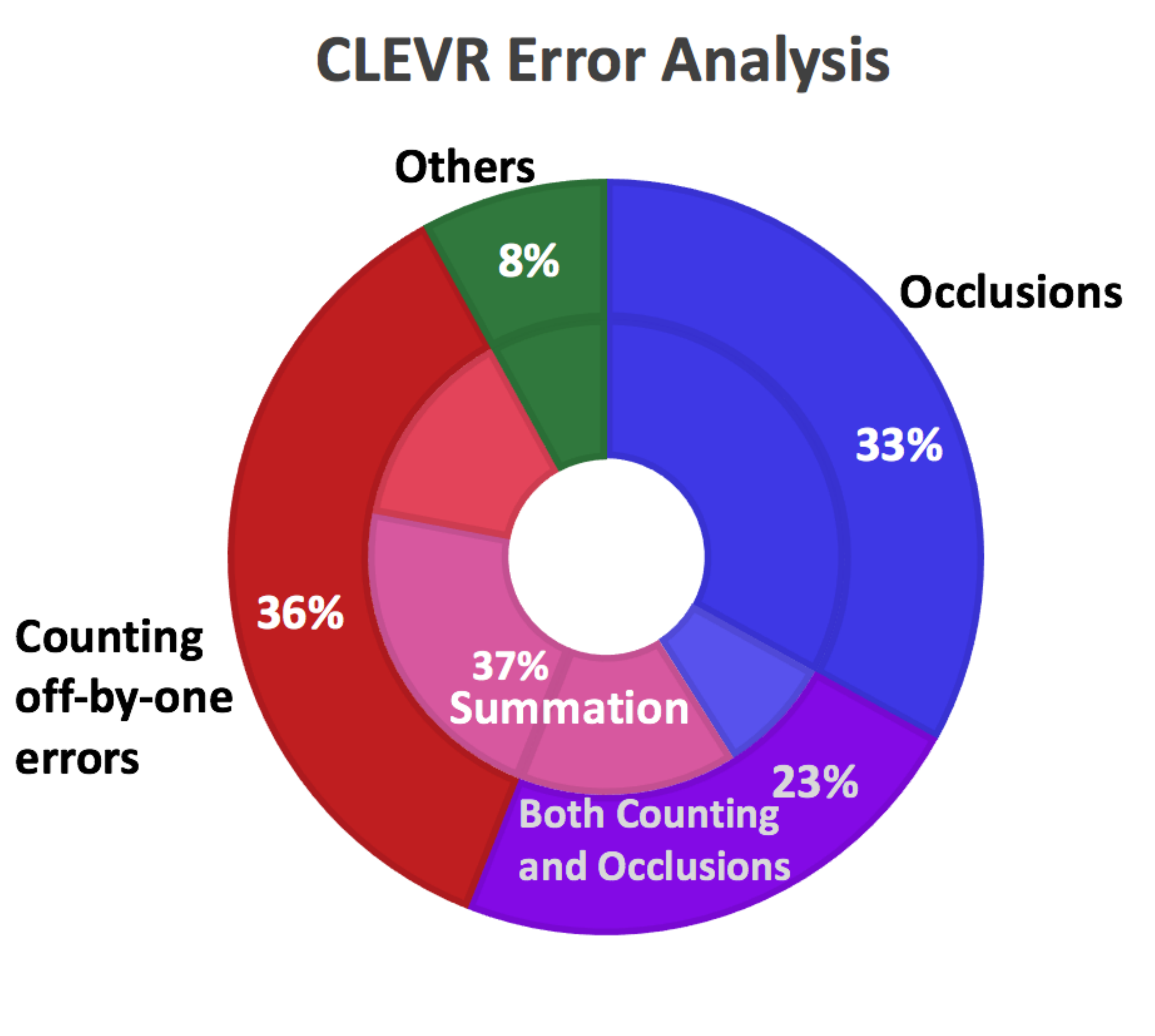}}
\vspace*{-3mm}
\subfloat{\includegraphics[width=1.0\linewidth]{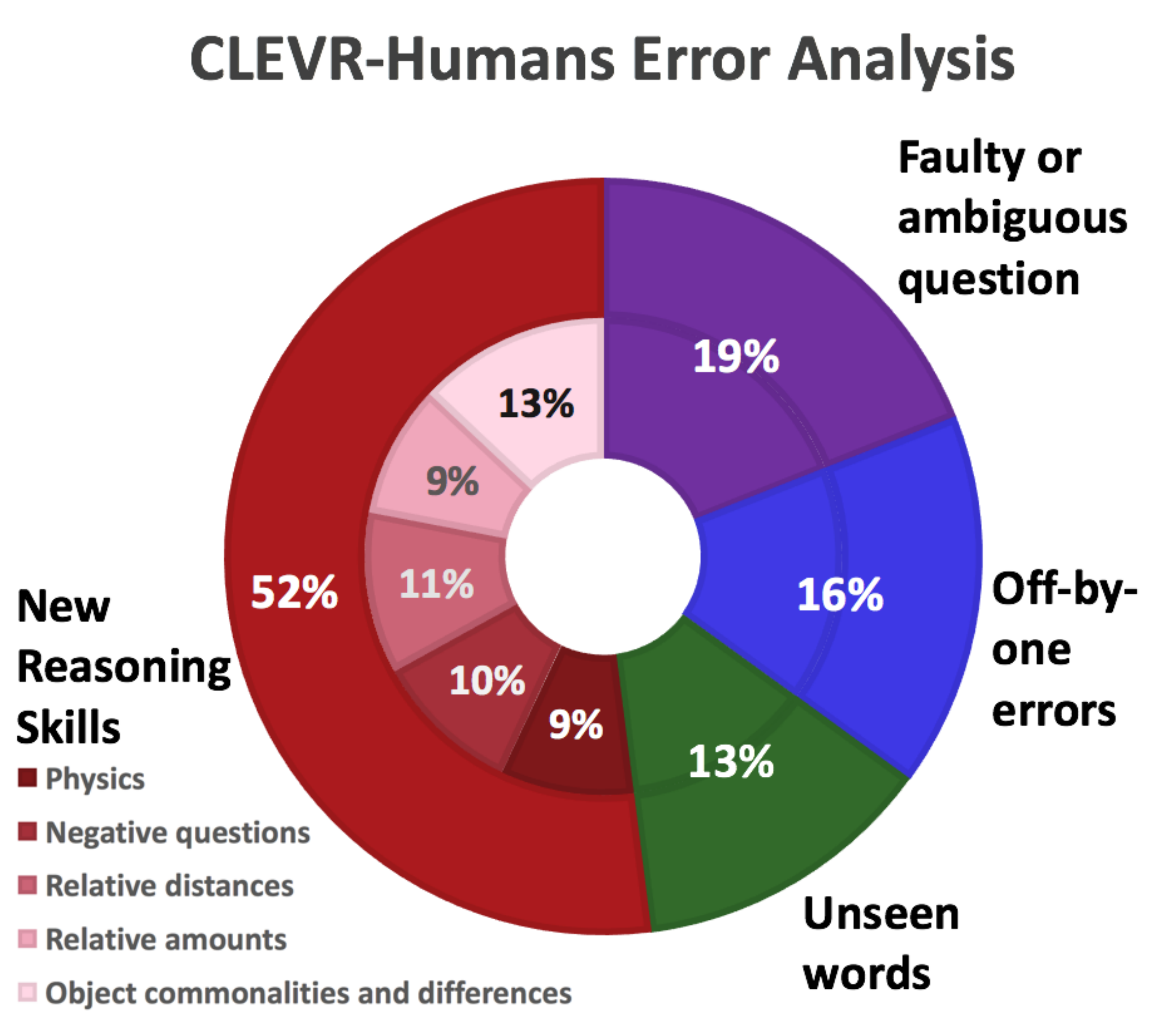}}

\caption{Error distribution for CLEVR and CLEVR-Humans.}
\label{fig:error_dist}

\end{wrapfigure}
We analyze our model's performance on the CLEVR-Humans dataset \citep{pgee}, consisting of natural language questions collected through crowdsourcing. As such, the dataset has a diverse vocabulary and linguistic variations, and it also demands more varied reasoning skills. Since the training set is relatively small, comprising 18k samples, we use it to finetune a model pre-trained on the primary CLEVR dataset, following prior work. 

As shown in \tabref{table1}, our model achieves state-of-the-art performance on CLEVR-Humans both before and after fine-tuning. It surpasses the next-best model by 5.6\% percent, achieving 81.5\%. The results substantiate the model's robustness against linguistic variations and noise as well as its ability to adapt to new and more diverse vocabulary and reasoning skills. The soft attention performed over the question allows the model to focus on the words that are most critical to answer the question while paying less attention to irrelevant linguistic variations. See figure \ref{hexample}, and figures \ref{hexample_supp} and \ref{vizz1} in the appendix for examples. 

In order to gain insight into the nature of the mistakes our model makes, we perform an error analysis for the CLEVR and CLEVR-Humans datasets (See \figref{error_dist}). Overall, we see that most of the errors in the CLEVR dataset are either off-by-one counting mistakes or result from heavy object occlusions. For CLEVR-Humans, we observe many errors that involve new reasoning skills that the model has not been trained for, such as ones that relate to physical properties (stability and reflections), relative distances and amounts, commonalities and uniqueness of objects, or negative questions. See \appref{erroranalysis} for further details. Nevertheless, the model does respond correctly to many of the questions that fall under these reasoning skills, as illustrated in figures \ref{hexample} and \ref{hexample_supp}, and so we speculate that the errors the model makes stem in part from the small size of the CLEVR-Human dataset.

\subsection{Computational and Data Efficiency}

We examine the learning curves of MAC and compare them to previous models\footnote{For previous models, we use the author's original publicly available implementations. All the models were trained with an equal batch size of 64 (as in the original implementations) and using the same hardware -- a single Maxwell Titan X GPU per model. To make sure the results are statistically significant, we run each model multiple (10) times, and plot the averages and confidence intervals.}: specifically, FiLM \citep{film}, the strongly-supervised PG+EE \citep{pgee}, and stacked-attention networks (SA) \citep{pgee, saAtt}. As shown in \figref{plots}, our model learns significantly faster than other approaches. While we do not have learning curves for the recent Relation Network model, \citet{rn} report 1.4 million iterations (equivalent to 125 epochs) to achieve 95.5\% accuracy, whereas our model achieves a comparable accuracy after only 3 epochs, yielding a 40x reduction in the length of the training process. Likewise, \citet{film} report a training time of 4 days, equivalent to 80 epochs, to reach accuracy of 97.7\%. In contrast, we achieve higher accuracy in 6 epochs, 9.5 hours overall, leading to a 10x reduction in training time.

In order to study the ability of MAC to generalize from a smaller amount of data, we explore its performance on subsets of CLEVR, sampled at random from the original 700k dataset. As shown in \figref{plots}, MAC outperforms the other models by a wide margin: For 50\% of the data, equivalent to 350k samples, other models obtain accuracies ranging between 70\% and 93\%, while our model achieves 97.6\%. The gap becomes larger as the dataset size reduces: for 25\% of the data, equivalent to 175k samples, the performance of other models is between 50\% and 77\%, while MAC maintains a high 94.3\% accuracy. 

Finally, for just 10\% of the data, amounting to 70k samples, our model is the only one to generalize well, with performance of 85.5\% on average, whereas the other leading models fail, achieving 49.0\%-54.9\%. Note that, as pointed out by \citet{clevr}, a simple baseline that predicts the most frequent answer for each question type already achieves 42.1\%, suggesting that answering only half of the questions correctly means that the other models barely learn to generalize from this smaller subset. These results demonstrate the robustness and generalization capacity of our architecture and its key role as a structural prior guiding MAC to learn the intended reasoning skills.

\subsection{Ablations}
\label{sec:ablations}

\begin{figure}[t]
\captionsetup[subfloat]{justification=raggedright, font=scriptsize,labelformat=empty,textfont=it}
\centering
\subfloat[{\normalfont{\textbf{Q}:}} What is the shape of the large item, \textcolor{purple}{\textbf{mostly occluded}} by the metallic cube? {\normalfont\textbf{A:}} sphere \textcolor{lime}{\ding{51}}]{\includegraphics[width=0.19\linewidth]{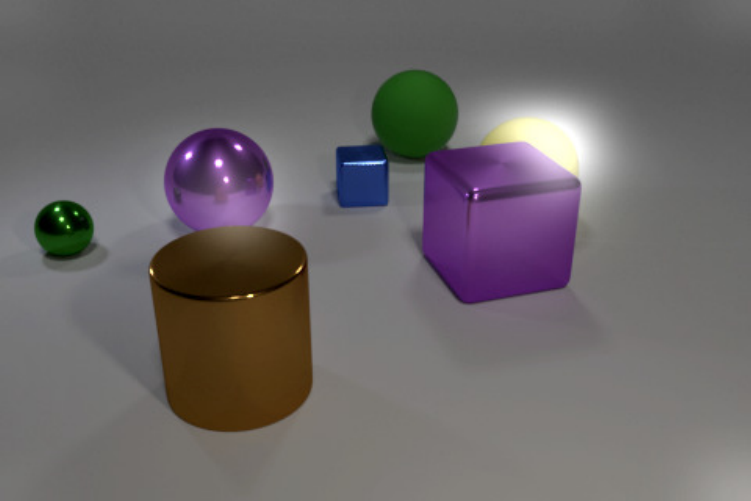}}
\hfill
\subfloat[{\normalfont\textbf{Q:}} What color is the object that is a \textcolor{blue}{\textbf{different}} size? {\normalfont\textbf{A:}} purple \textcolor{lime}{\ding{51}}]{\includegraphics[width=0.19\linewidth]{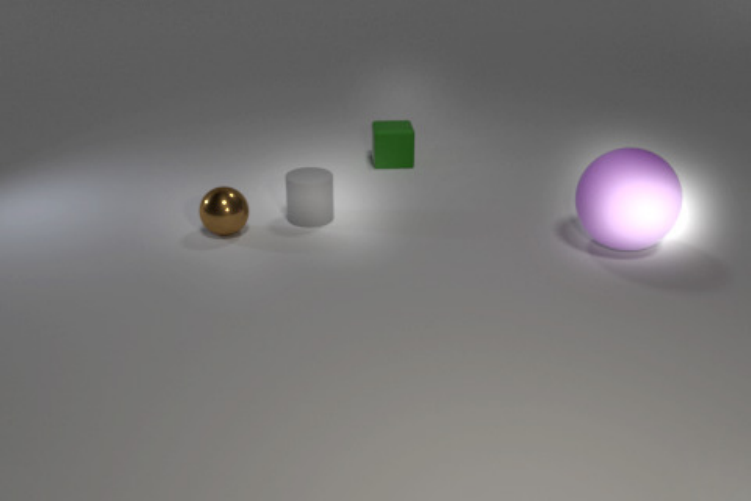}}
\hfill
\subfloat[{\normalfont\textbf{Q:}} What color ball is \textcolor{green}{\textbf{close to}} the small purple cylinder? {\normalfont\textbf{A:}} gray \textcolor{lime}{\ding{51}}]{\includegraphics[width=0.19\linewidth]{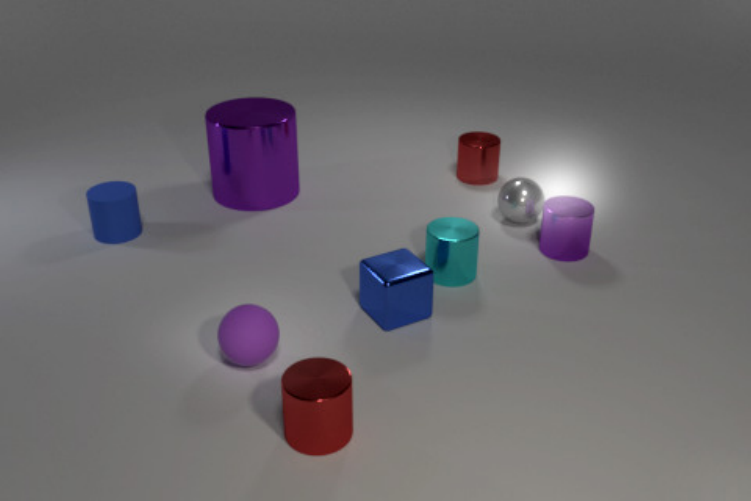}}
\hfill
\subfloat[{\normalfont\textbf{Q:}} What color block is \textcolor{red}{\textbf{farthest front}}? {\normalfont\textbf{A:}} purple \textcolor{lime}{\ding{51}}]{\includegraphics[width=0.19\linewidth]{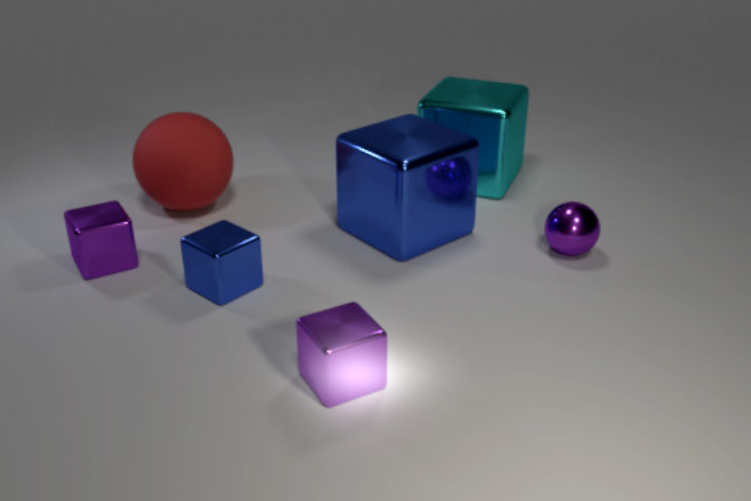}}
\hfill
\subfloat[{\normalfont\textbf{Q:}} Are any objects \textcolor{yellow}{\textbf{gold}}? {\normalfont\textbf{A}}: yes \textcolor{lime}{\ding{51}}]{\includegraphics[width=0.19\linewidth]{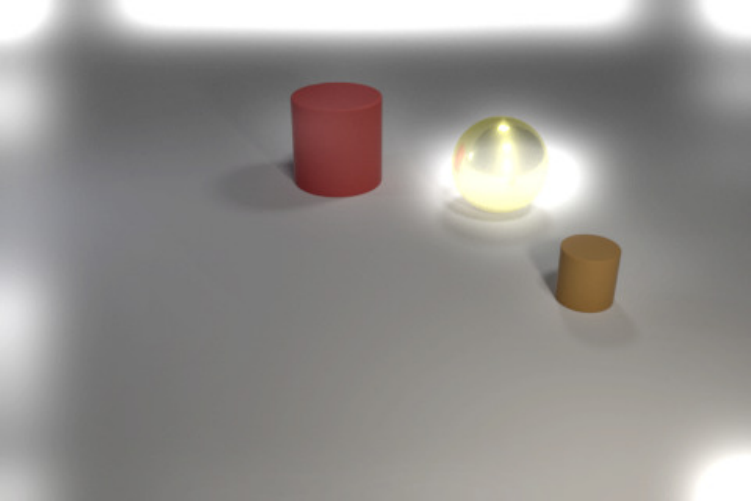}}
\caption{CLEVR-Humans examples showing the model performs novel reasoning skills that do not appear in CLEVR, including: \textcolor{purple}{\textbf{obstructions}}, \textcolor{blue}{\textbf{object uniqueness}}, \textcolor{green}{\textbf{relative distances}}, \textcolor{red}{\textbf{superlatives}} and \textcolor{yellow}{\textbf{new concepts}}. }
\label{hexample}
\label{fig:hexample}
\end{figure}

\begin{wrapfigure}[11]{r}{0.24\textwidth}
\centering
\vspace{-8mm}

\centering
\subfloat{\includegraphics[width=1.0\linewidth]{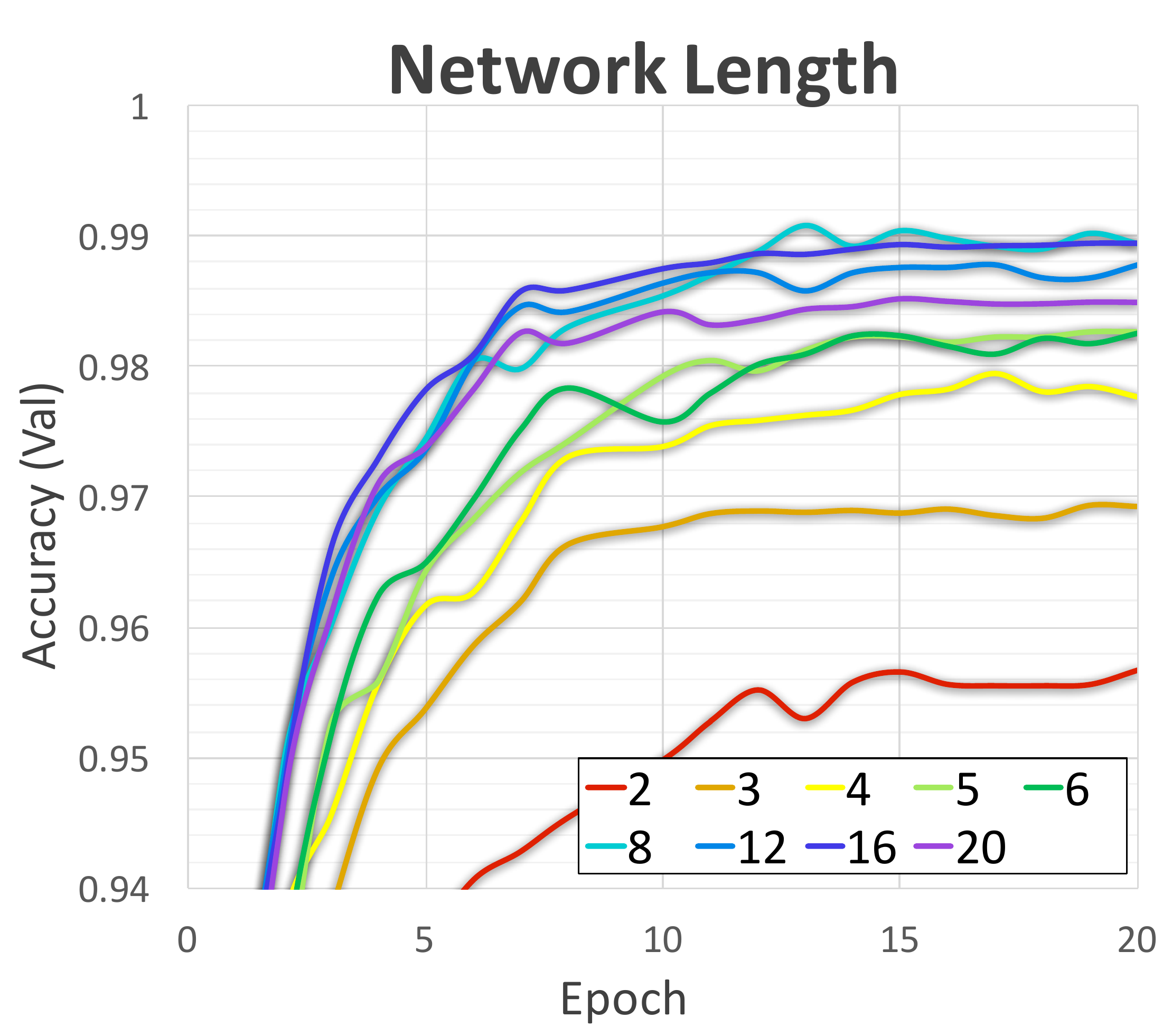}}

\vspace{-2mm}
\caption{Model performance as a function of the network length. }
\end{wrapfigure}

To gain better insight into the relative contribution of the design choices we made, we perform extensive ablation studies. See \figref{plots} and \appref{ablationsSupp} for accuracies and learning curves. The experiments demonstrate the robustness of the model to hyperparameter variations such as network dimension and length, and shed light on the significance of various aspects and components of the model, as discussed below:

\textbf{Question Attention.} The ablations show that using attention over the question words (see \secref{CU}) is highly effective in accelerating learning and enhancing generalization capacity. Using the complete question \(q\) instead of the attention-based control state leads to a significant drop of 18.5\% in the overall accuracy. Likewise, using unconstrained recurrent control states, without casting them back onto the question words space (step (3) in \secref{CU}) leads to a 6x slowdown in the model convergence rate. These results illustrate the importance and usefulness of decomposing the question into a series of simple operations, such that a single cell is faced with learning the semantics of one or a few words at a time, rather than grasping all of the question at once. They provide evidence for the efficacy of \textit{using attention as a regularization mechanism}, by restricting the input and output spaces of each MAC cell.

\textbf{Control and Memory.} Maintaining separation between control and memory proves to be another key property that contributes significantly to the model's accuracy, learning rate and data efficiency. We perform experiments for a variant of the MAC cell in which we maintain one hidden state that plays both the roles of the control and memory, iteratively attending and integrating information from both the question and the image. While this approach achieves a final accuracy of 93.75\%, it leads to a sharp drop in the convergence rate, as shown in \figref{plots}, and a 20.2\% reduction in the final accuracy for a smaller 10\% subset of CLEVR. The results make a strong case for our model's main design choice, namely, splitting the computation into two dual paths: one that decomposes the linguistic information and another that reconstructs the corresponding visual information.

The design choices discussed above were found to be the most significant to the model's overall accuracy, convergence rate and generalization. Other design choices that were found  beneficial include (1) predicting the final answer based on both the final memory state and the question (see \secref{output}), and (2) considering knowledge base elements directly (step (2) in \secref{RU}), resulting in 19.8\%  and 11.1\% improvement for a 10\% subset of CLEVR, respectively. Please refer to \appref{ablationsSupp} for further discussion and results.
\label{sec:ablations}

\begin{wrapfigure}[20]{r}{0.3\textwidth}
\vspace*{-10mm}
\begin{minipage}{0.12\textwidth}
\centering
\vspace*{4mm}
\subfloat{\includegraphics[width=1.0\linewidth]{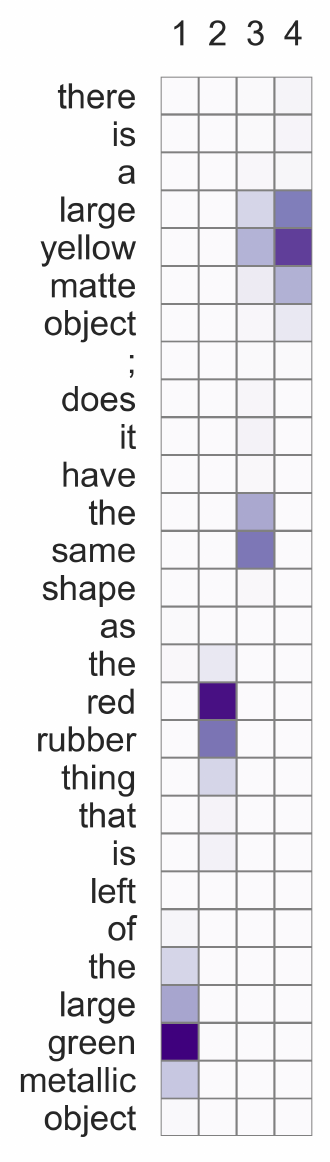}}
\end{minipage}
\begin{minipage}{0.13\textwidth}
\noindent
\centering
\subfloat{\includegraphics[width=1.0\linewidth]{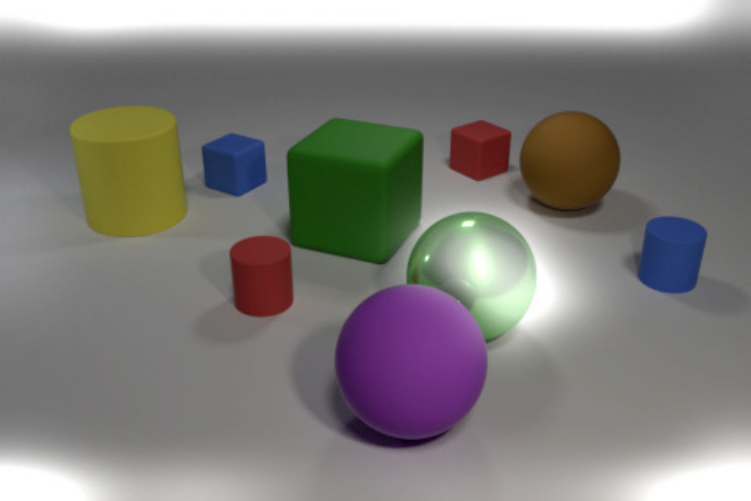}}
\vspace{-2mm}
\subfloat{\includegraphics[width=1.0\linewidth]{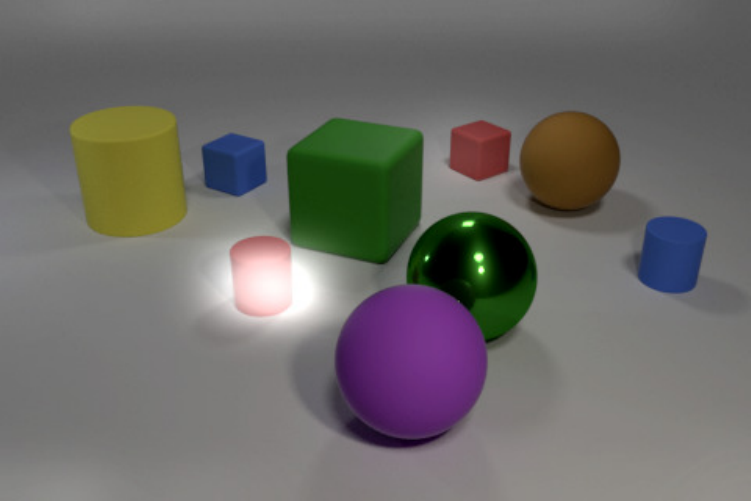}}
\vspace{-2mm}
\subfloat{\includegraphics[width=1.0\linewidth]{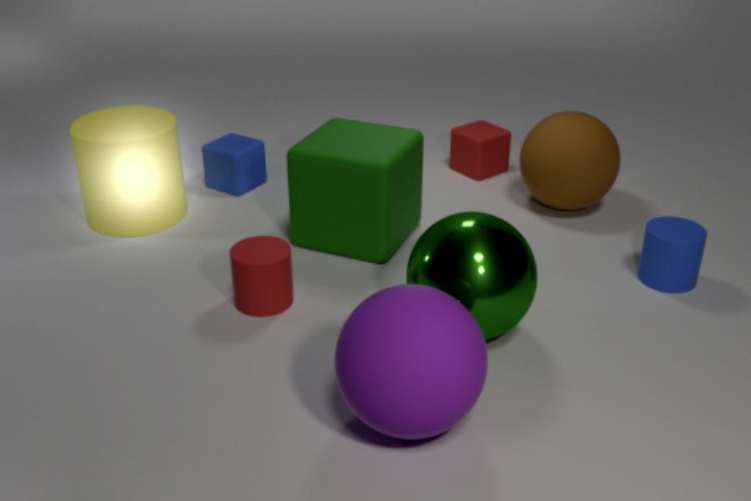}}
\vspace{-2mm}
\subfloat{\includegraphics[width=1.0\linewidth]{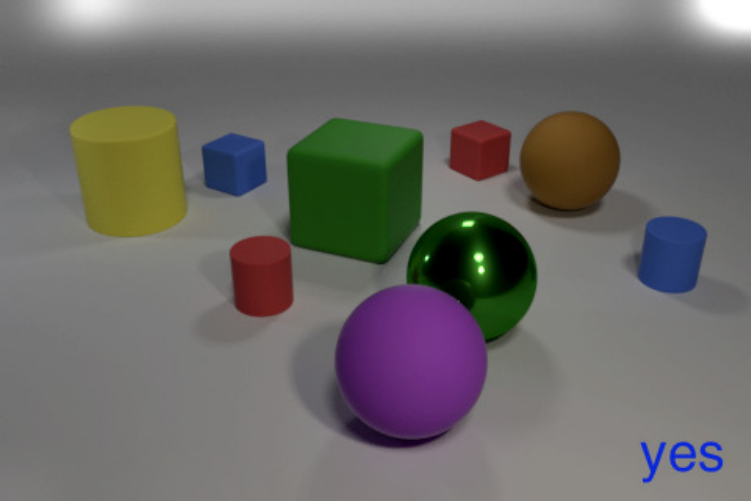}}
\end{minipage}

\scriptsize 

\caption{Attention maps produced by MAC, showing how it tracks transitive relations between objects.}
\label{int_example2}

\end{wrapfigure} 

\subsection{Interpretability}

\begin{figure}[t]
\vspace*{-15mm}
\centering
\begin{minipage}{0.18\textwidth}
\vspace*{4mm}
\subfloat{\includegraphics[width=1.15\linewidth]{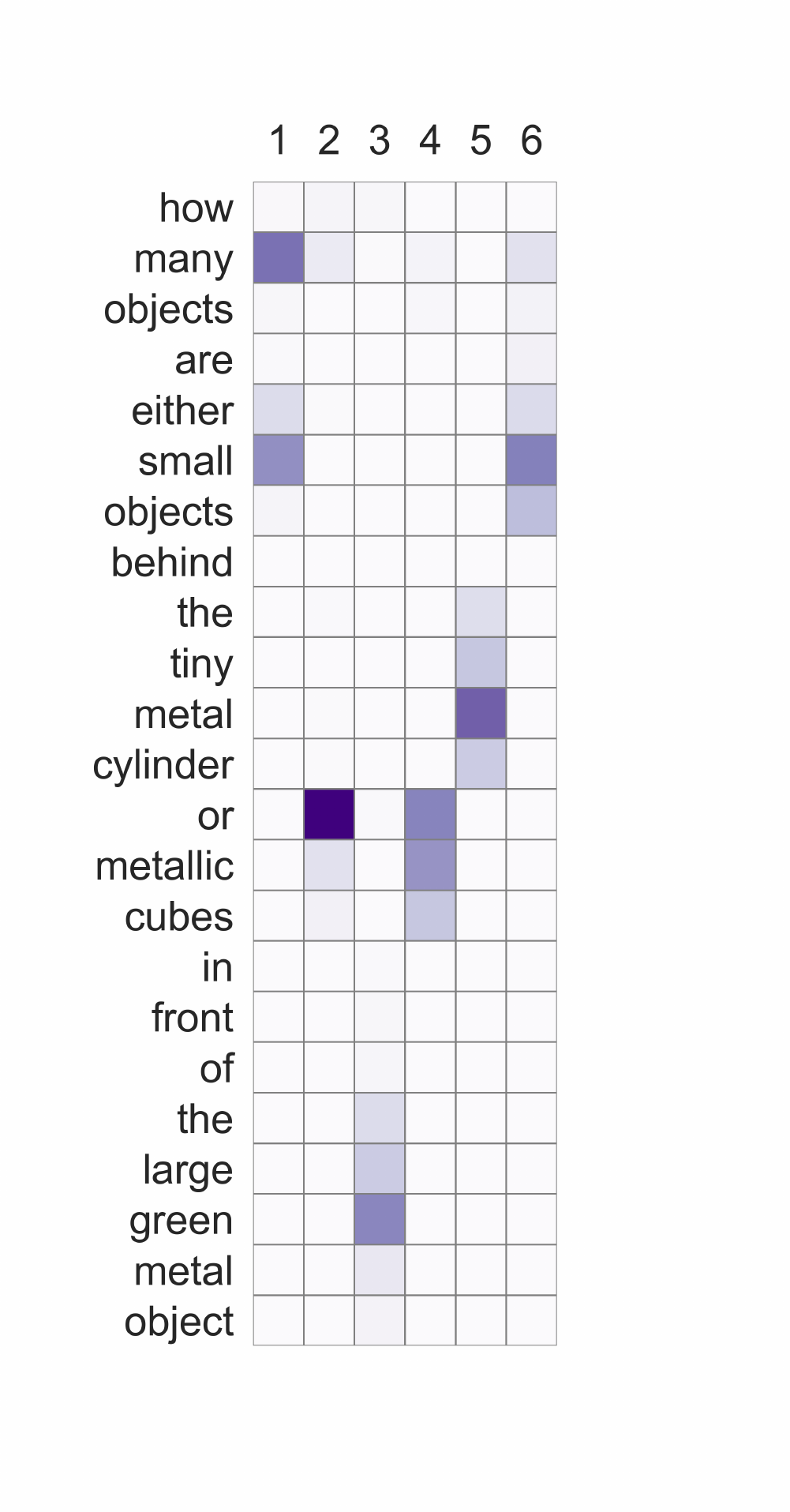}}
\end{minipage}
\begin{minipage}{0.7\textwidth}
\noindent
\centering
\subfloat{\includegraphics[width=0.325\linewidth]{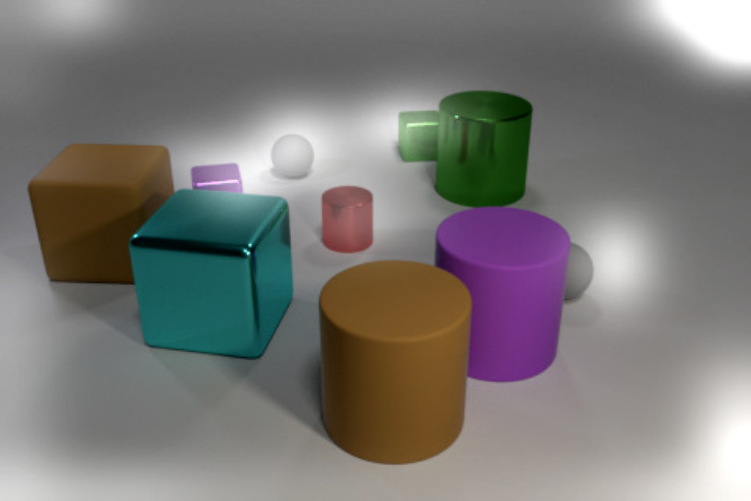}}
\hfill
\subfloat{\includegraphics[width=0.325\linewidth]{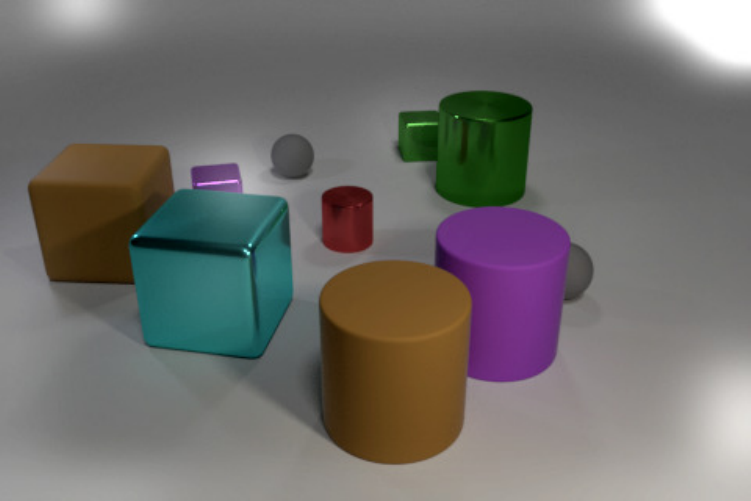}}
\hfill
\subfloat{\includegraphics[width=0.325\linewidth]{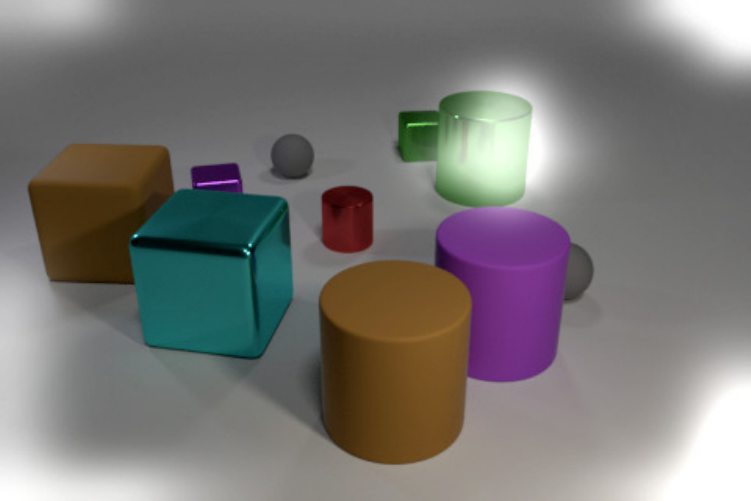}}
\vspace*{-2.5mm}
\subfloat{\includegraphics[width=0.325\linewidth]{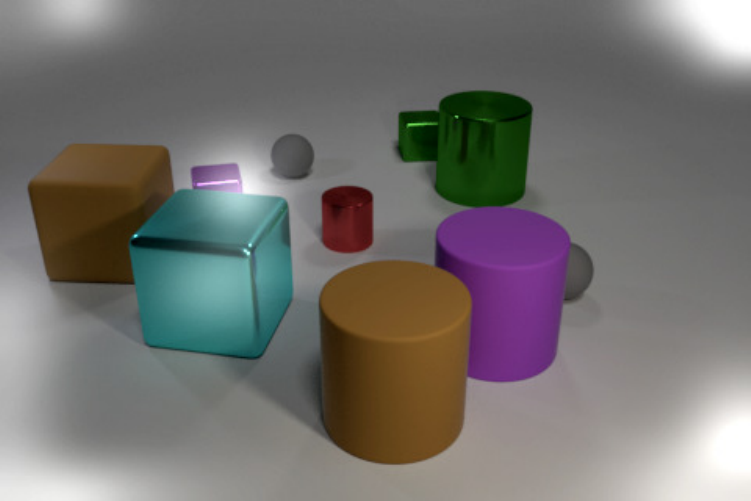}}
\hfill
\subfloat{\includegraphics[width=0.325\linewidth]{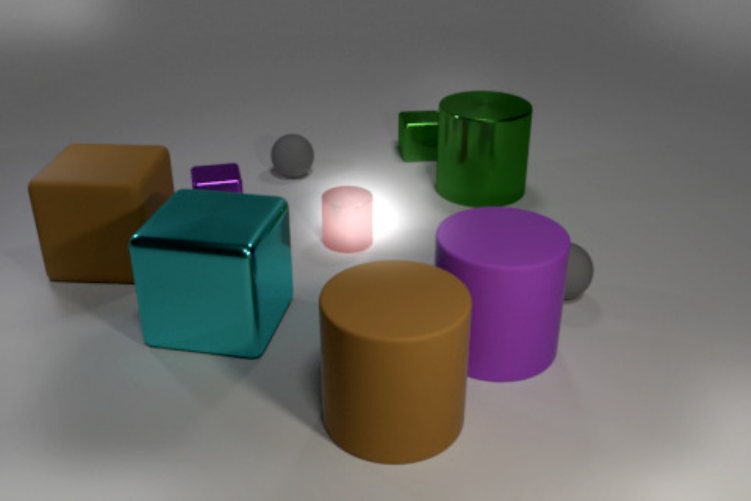}}
\hfill
\subfloat{\includegraphics[width=0.325\linewidth]{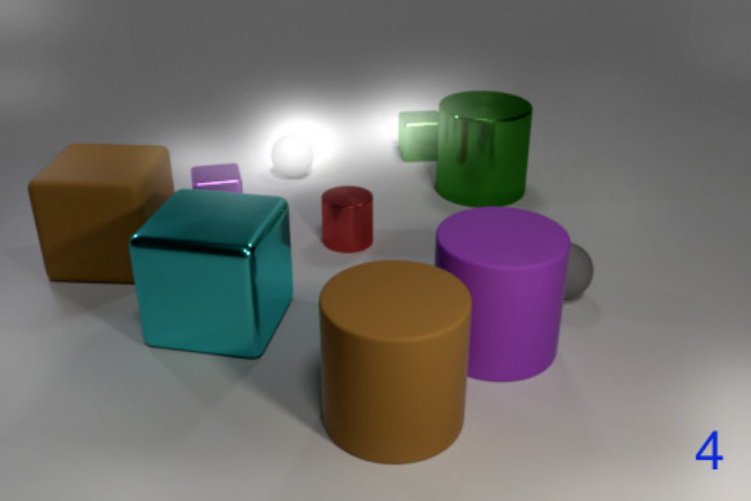}}
\end{minipage}

\vspace*{-4mm}

\scriptsize 

\caption{Attention maps produced by MAC which provide some evidence for the ability of the model to perform counting and summation of small numbers. Note how the first iterations focus on the key structural question words ``many" and ``or" that inform the model of the required reasoning operation it has to perform.}
\label{intr}

\end{figure}

To obtain better insight into the underlying reasoning processes MAC learns to perform, we study visualizations of the attention distributions produced by the model during its iterative computation, and provide examples in figures \ref{int_example2}, \ref{intr}, \ref{vizz1}, and \ref{vizz2}. Examining the sequence of attention maps over the image and the question reveals several qualitative patterns and properties that characterize MAC's mode of operation. 

First, we observe that both the linguistic and visual attentions of the model are very focused on specific terms or regions in the image, and commonly refer to concrete objects (``the shiny red cube" or the ``metallic cylinder") or question structural keywords (``or", ``and" or ``how many"). More importantly, the attention maps give evidence of the ability of the model to capture the  underlying semantic structure of the question, traversing the correct transitive relations between the objects it refers to. For instance, we see in figure \ref{int_example2} how the model explicitly decomposes the question into the correct reasoning steps: first identifying the \textcolor{green}{\textit{\textbf{green ball}}}, then focusing on the \textcolor{red}{\textit{\textbf{red cylinder}}} that is located left of the ball, and finally attending to the \textcolor{yellow}{\textit{\textbf{yellow cylinder}}}. In the second step, note how the model attends only to the relevant red cylinder and not to other \textit{red rubber things}, correctly resolving the indirect reference in the question. This shows strong evidence of the ability of the model to perform transitive reasoning, integrating information from prior steps that allows it to focus only on the relevant objects, even when they are not mentioned explicitly.

In figure \ref{intr}, we further see how the model interprets a multi-step counting question, apparently summing up the amounts of two referenced object groups to produce the correct overall count. These observations suggest that the model infers and effectively performs complex reasoning processes in a transparent manner. 

\section{Conclusion}
We have introduced the Memory, Attention and Composition (MAC) network, an end-to-end differentiable architecture for machine reasoning. The model solves problems by decomposing them into a series of inferred reasoning steps that are performed successively to accomplish the task at hand. It uses a novel recurrent MAC cell that aims to formulate the inner workings of a single universal reasoning operation by maintaining a separation between memory and control. These MAC cells are chained together to produce explicit and structured multi-step reasoning processes. We demonstrate the versatility, robustness and transparency of the model through quantitative and qualitative studies, achieving state-of-the-art results on the CLEVR task for visual reasoning, and generalizing well even from a 10\% subset of the data. The experimental results further show that the model can adapt to novel situations and diverse language, and generate interpretable attention-based rationales that reveal the underlying reasoning it performs. While CLEVR provides a natural testbed for our approach, we believe that the architecture will prove beneficial for other multi-step reasoning and inference tasks, including reading comprehension, textual question answering, and real-world VQA.

\section{Acknowledgments}
We wish to thank Justin Johnson, Aaron Courville, Ethan Perez, Harm de Vries, Mateusz Malinowski, Jacob Andreas, and the anonymous reviewers for the helpful suggestions, comments and discussions. Stanford University gratefully acknowledges the support of the Defense Advanced
Research Projects Agency (DARPA) Communicating with Computers (CwC) program under ARO prime contract no. W911NF15-1-0462 for supporting this work.

\newpage
\bibliography{iclr2018_conference}
\bibliographystyle{iclr2018_conference}

\newpage
\appendix

\section*{Supplementary Material}

\section{Implementation and Training details}
\label{sec:expDetails} 
We train our model using Adam \citep{adam}, with a learning rate of \({10}^{-4}\) and a batch size of \(64\). We use gradient clipping, and employ early stopping based on the validation accuracy, resulting in a training process of 10--20 epochs, equivalent to roughly 15--30 hours on a single Maxwell Titan X GPU. Word vectors have dimension 300 and were initialized randomly using a standard uniform distribution. The exponential moving averages of the model weights are maintained during training, with a decay rate of 0.999, and used at test time instead of the raw weights. We use variational dropout of 0.15 across the network, along with ELU as non-linearity, which, in our experience, accelerates training and performs favorably compared to the more standard ReLU. 

\section{Error Analysis}
\label{sec:erroranalysis} 
\begin{figure}[t]
\captionsetup[subfloat]{justification=raggedright, font=scriptsize,labelformat=empty,textfont=it}
\centering
\subfloat[{\normalfont{\textbf{Q}:}} What color is the metallic cylinder in front of the \textcolor{yellow}{\textbf{silver}} cylinder? {\normalfont\textbf{A:}} cyan \textcolor{lime}{\ding{51}}]{\includegraphics[width=0.24\linewidth]{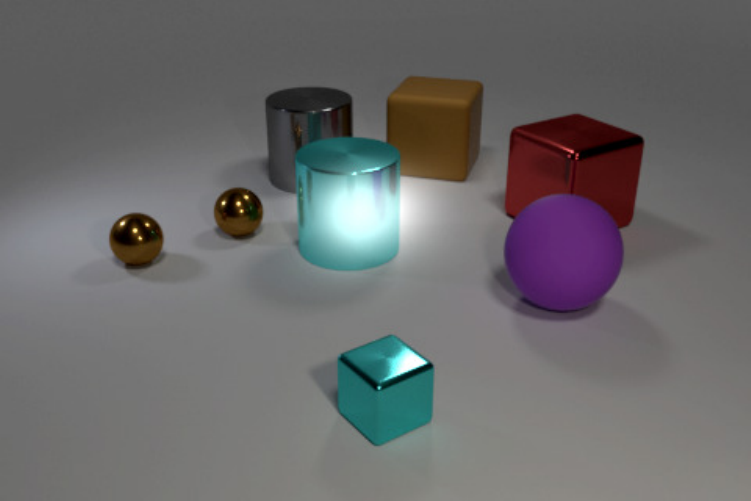}}
\hfill
\subfloat[{\normalfont{\textbf{Q}:}} What is the object made of \textcolor{purple}{\textbf{hiding behind}} the green cube? {\normalfont\textbf{A:}} rubber \textcolor{lime}{\ding{51}}]{\includegraphics[width=0.24\linewidth]{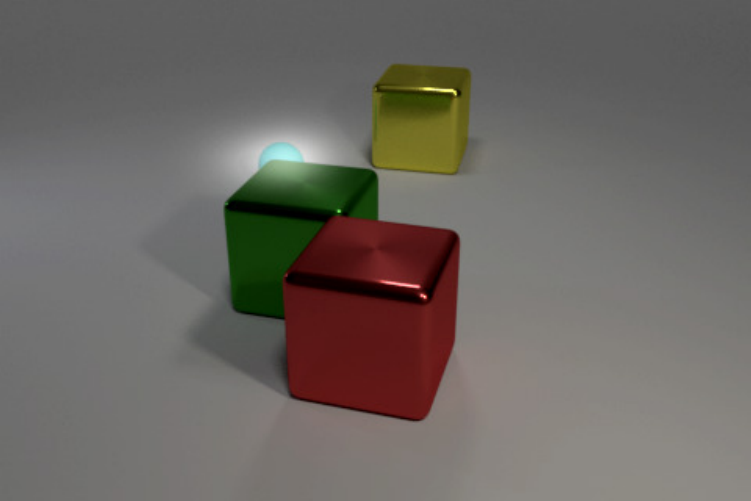}}
\hfill
\subfloat[{\normalfont{\textbf{Q}:}} What is the color of the ball that is \textcolor{red}{\textbf{farthest away}}? {\normalfont\textbf{A:}} blue \textcolor{lime}{\ding{51}}]{\includegraphics[width=0.24\linewidth]{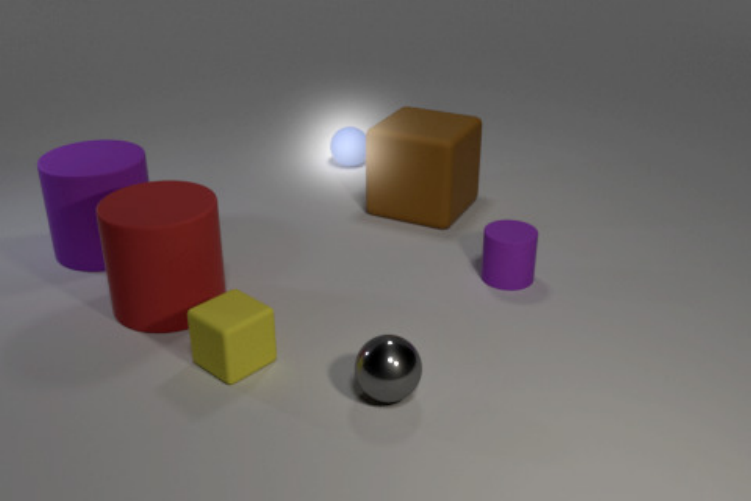}}
\hfill
\subfloat[{\normalfont{\textbf{Q}:}} How many matte cubes are there? {\normalfont\textbf{A:}} 2 \textcolor{lime}{\ding{51}}]{\includegraphics[width=0.24\linewidth]{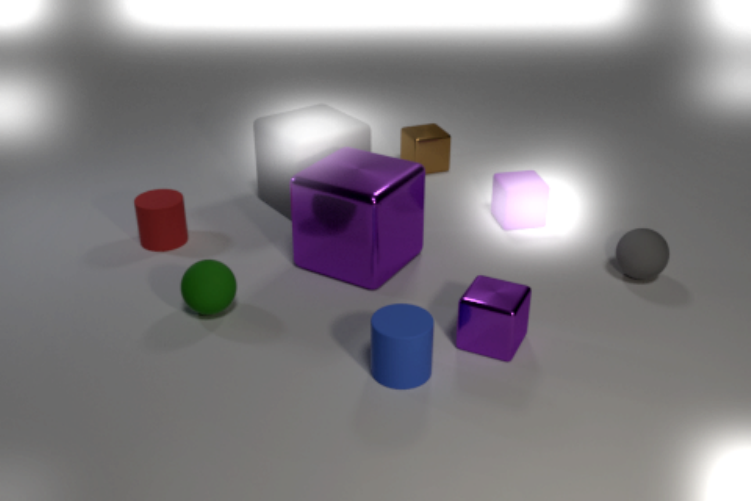}}

\subfloat[{\normalfont{\textbf{Q}:}} How many spheres are \textcolor{yellow}{\textbf{pictured}}? {\normalfont\textbf{A:}} 4 \textcolor{lime}{\ding{51}}]{\includegraphics[width=0.24\linewidth]{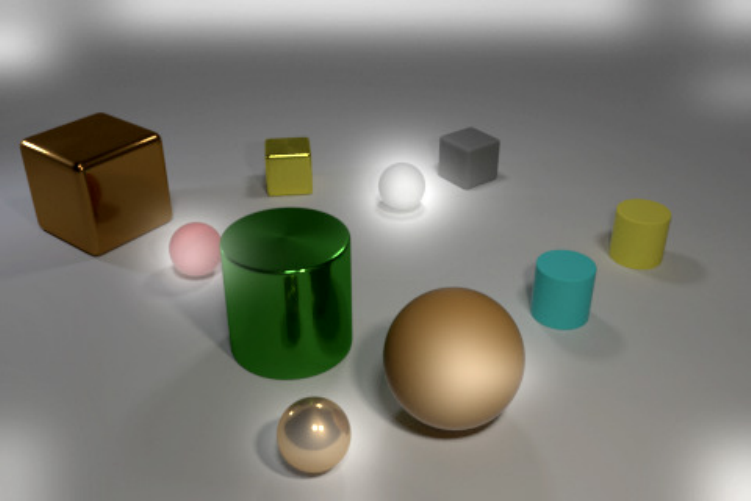}}
\hfill
\subfloat[{\normalfont{\textbf{Q}:}} How many \textcolor{yellow}{\textbf{square}} objects are in the \textcolor{yellow}{\textbf{picture}}? {\normalfont\textbf{A:}} 4 \textcolor{lime}{\ding{51}}]{\includegraphics[width=0.24\linewidth]{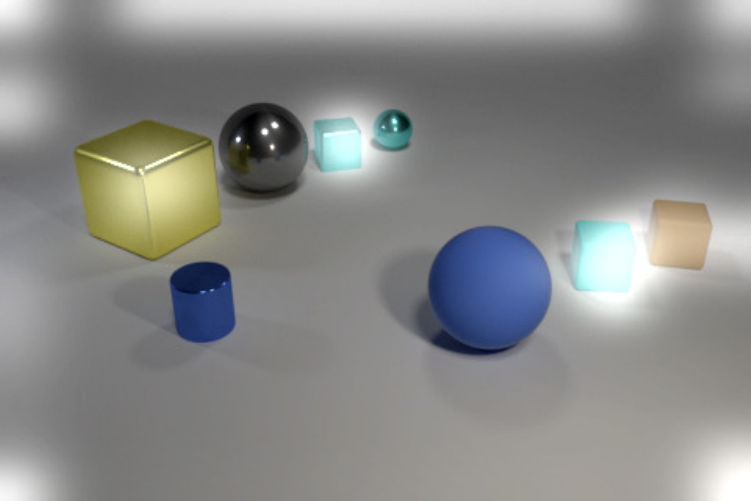}}
\hfill
\subfloat[{\normalfont{\textbf{Q}:}} What object is to the \textcolor{red}{\textbf{far right}}? {\normalfont\textbf{A:}} cube \textcolor{lime}{\ding{51}}]{\includegraphics[width=0.24\linewidth]{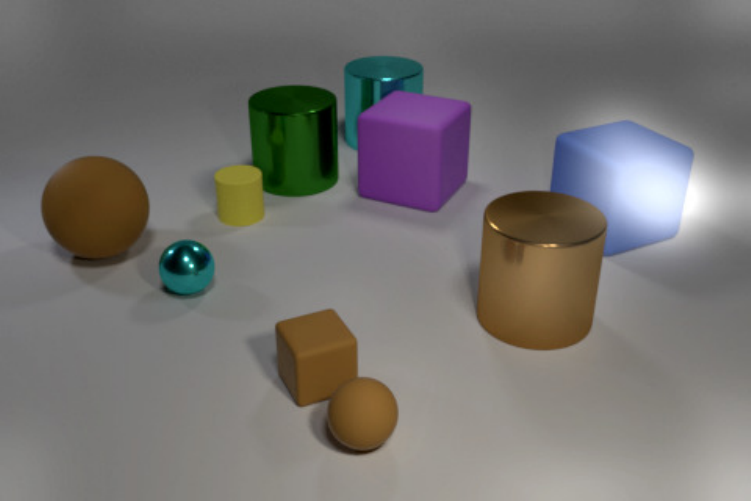}}
\hfill
\subfloat[{\normalfont{\textbf{Q}:}} Are the yellow blocks \textcolor{blue}{\textbf{the same}}? {\normalfont\textbf{A:}} no \textcolor{lime}{\ding{51}}]{\includegraphics[width=0.24\linewidth]{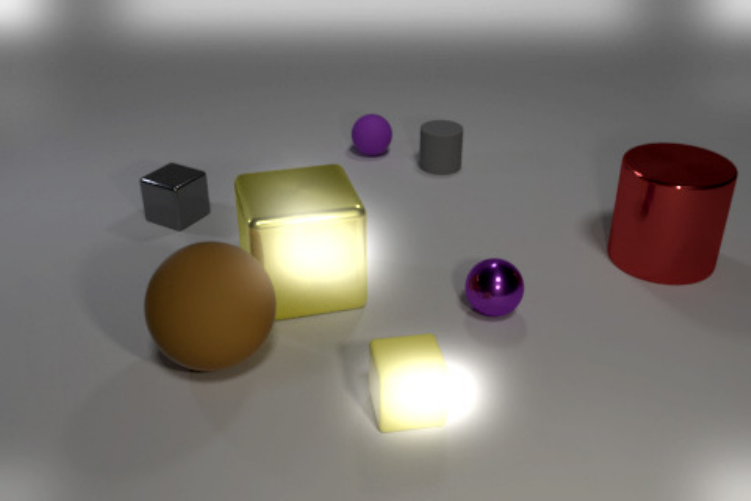}}

\subfloat[{\normalfont{\textbf{Q}:}} What shape is the \textcolor{red}{\textbf{smallestt}} object in this image? {\normalfont\textbf{A:}} sphere \textcolor{lime}{\ding{51}}]{\includegraphics[width=0.24\linewidth]{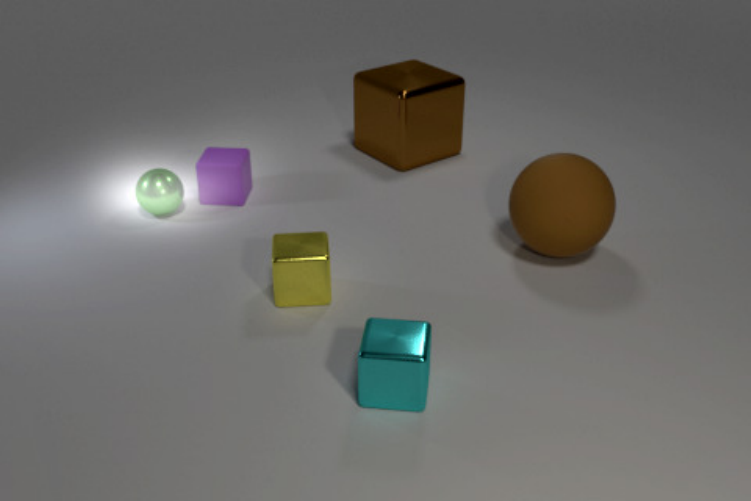}}
\hfill
\subfloat[{\normalfont{\textbf{Q}:}} What object looks like a \textcolor{yellow}{\textbf{caramel}}? {\normalfont\textbf{A:}} cube \textcolor{lime}{\ding{51}}]{\includegraphics[width=0.24\linewidth]{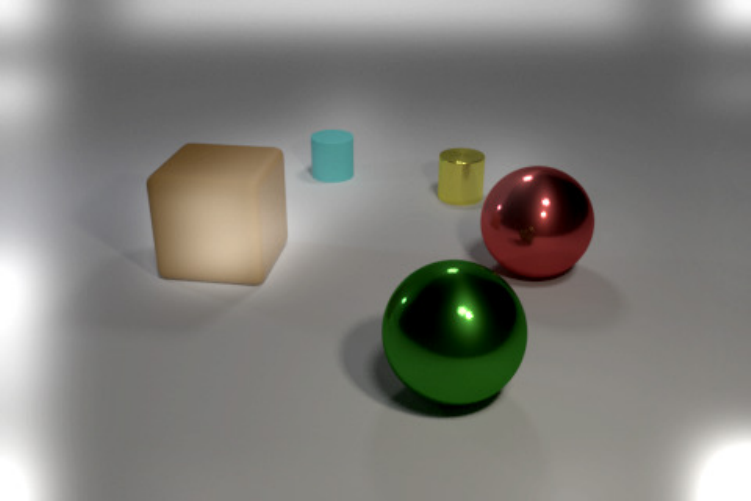}}
\hfill
\subfloat[{\normalfont{\textbf{Q}:}} Can a ball \textcolor{purple}{\textbf{stay still on top}} of one another? {\normalfont\textbf{A:}} yes (no) \textcolor{red}{\ding{55}}]{\includegraphics[width=0.24\linewidth]{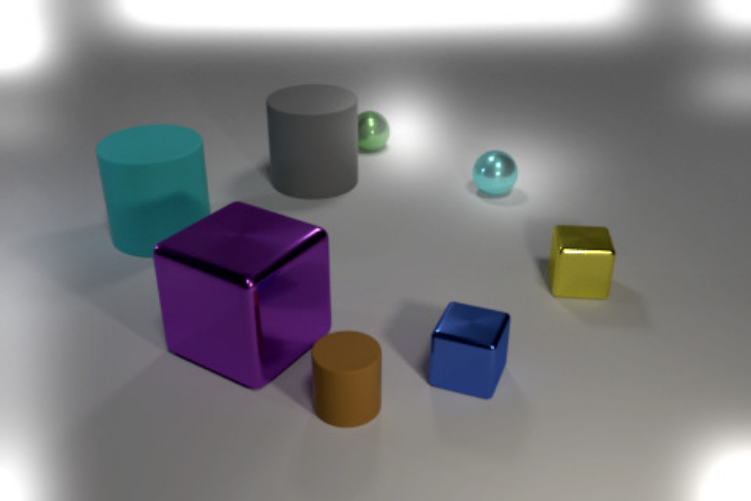}}
\hfill
\subfloat[{\normalfont{\textbf{Q}:}} What color is the \textcolor{green}{\textbf{center}} object? {\normalfont\textbf{A:}} blue \textcolor{lime}{\ding{51}}]{\includegraphics[width=0.24\linewidth]{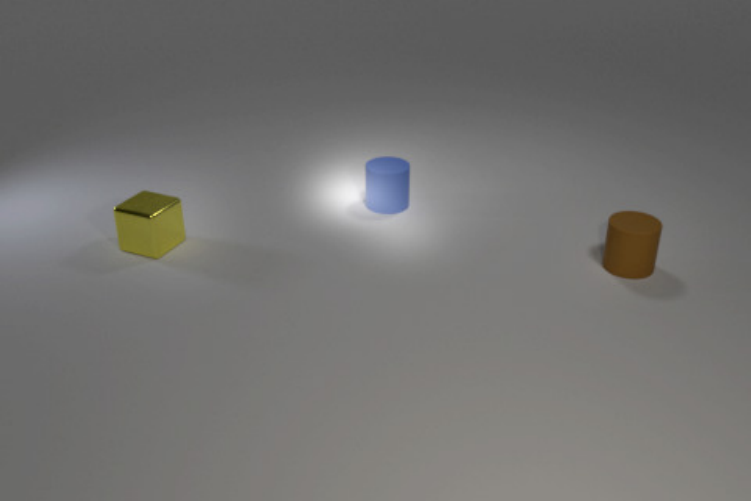}}

\subfloat[{\normalfont{\textbf{Q}:}} How many gray objects are shown? {\normalfont\textbf{A:}} 3 \textcolor{lime}{\ding{51}}]{\includegraphics[width=0.24\linewidth]{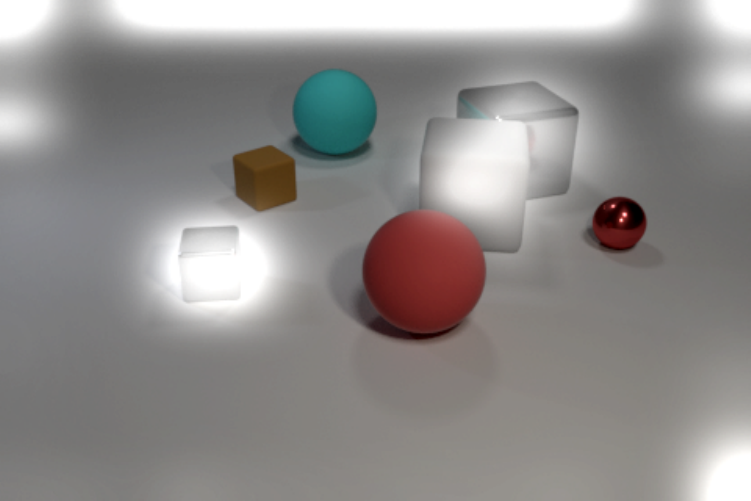}}
\hfill
\subfloat[{\normalfont{\textbf{Q}:}} How many small objects are rubber? {\normalfont\textbf{A:}} 2 \textcolor{lime}{\ding{51}}]{\includegraphics[width=0.24\linewidth]{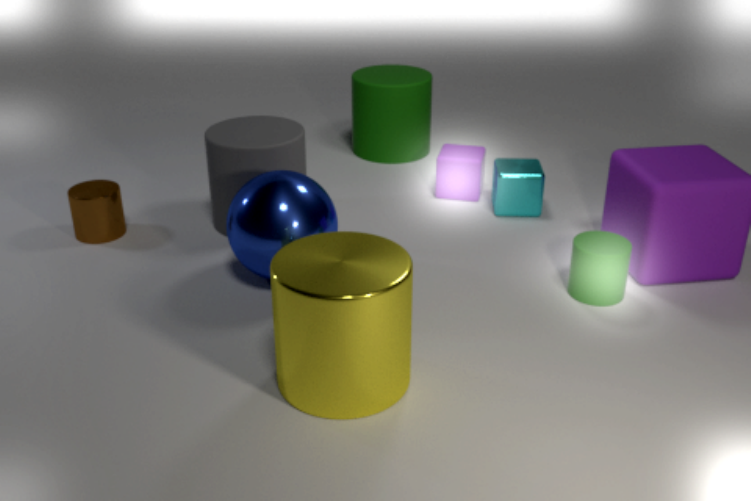}}
\hfill
\subfloat[{\normalfont{\textbf{Q}:}} What color is the \textcolor{red}{\textbf{largest}} cube? {\normalfont\textbf{A:}} yellow \textcolor{lime}{\ding{51}}]{\includegraphics[width=0.24\linewidth]{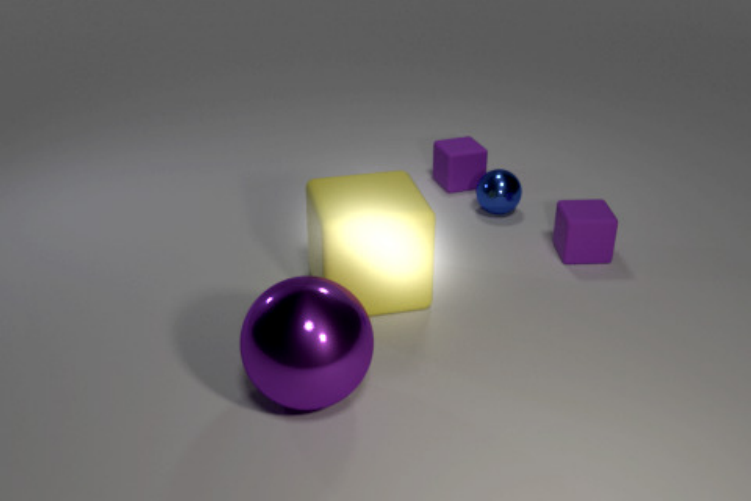}}
\hfill
\subfloat[{\normalfont{\textbf{Q}:}} What shape are \textcolor{blue}{\textbf{most}} of the shiny items? {\normalfont\textbf{A:}} sphere \textcolor{lime}{\ding{51}}]{\includegraphics[width=0.24\linewidth]{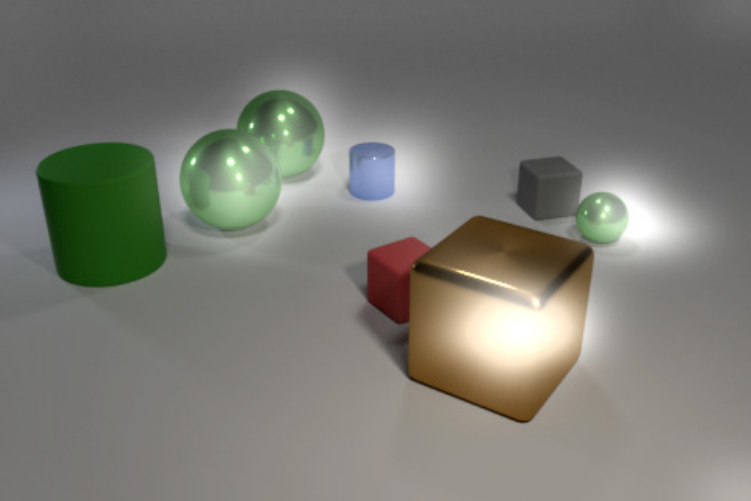}}

\subfloat[{\normalfont{\textbf{Q}:}} What is the \textcolor{yellow}{\textbf{tan}} object made of? {\normalfont\textbf{A:}} rubber \textcolor{lime}{\ding{51}}]{\includegraphics[width=0.24\linewidth]{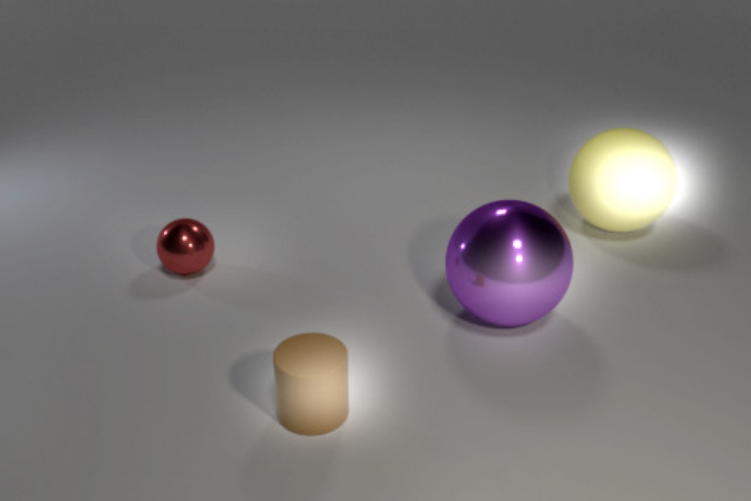}}
\hfill
\subfloat[{\normalfont{\textbf{Q}:}} Are \textcolor{blue}{\textbf{half}} the items shown green? {\normalfont\textbf{A:}} yes (no) \textcolor{red}{\ding{55}}]{\includegraphics[width=0.24\linewidth]{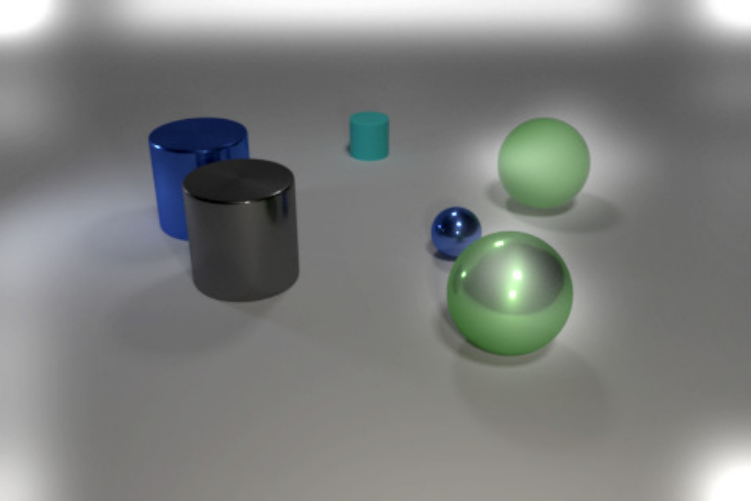}}
\hfill
\subfloat[{\normalfont{\textbf{Q}:}} What color object is \textcolor{red}{\textbf{biggest}}? {\normalfont\textbf{A:}} blue \textcolor{lime}{\ding{51}}]{\includegraphics[width=0.24\linewidth]{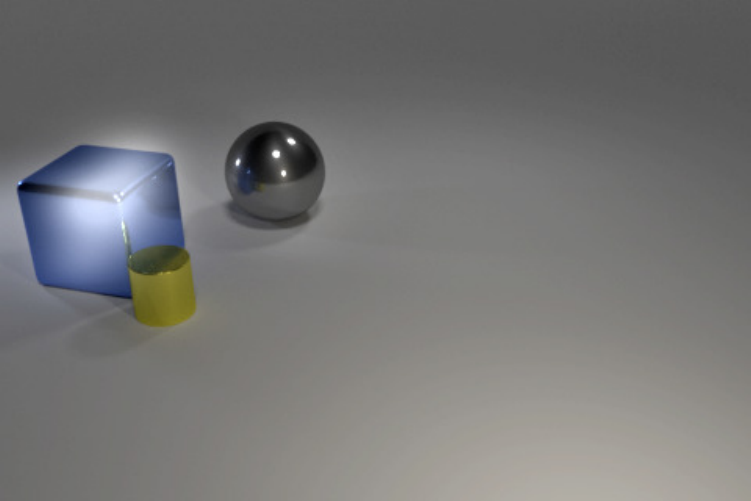}}
\hfill
\subfloat[{\normalfont{\textbf{Q}:}} Which shape is a \textcolor{blue}{\textbf{different}} color from the others? {\normalfont\textbf{A:}} cylinder \textcolor{lime}{\ding{51}}]{\includegraphics[width=0.24\linewidth]{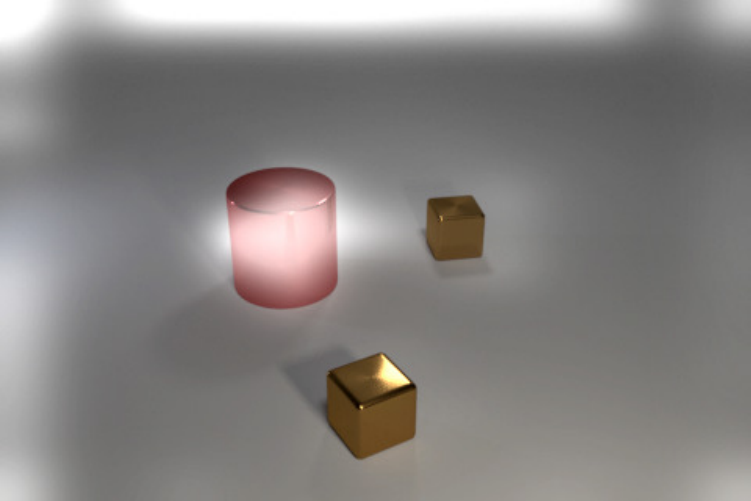}}

\subfloat[{\normalfont{\textbf{Q}:}} How many other objects are \textcolor{blue}{\textbf{the same}} size as the blue ball? {\normalfont\textbf{A:}} 7 \textcolor{lime}{\ding{51}}]{\includegraphics[width=0.24\linewidth]{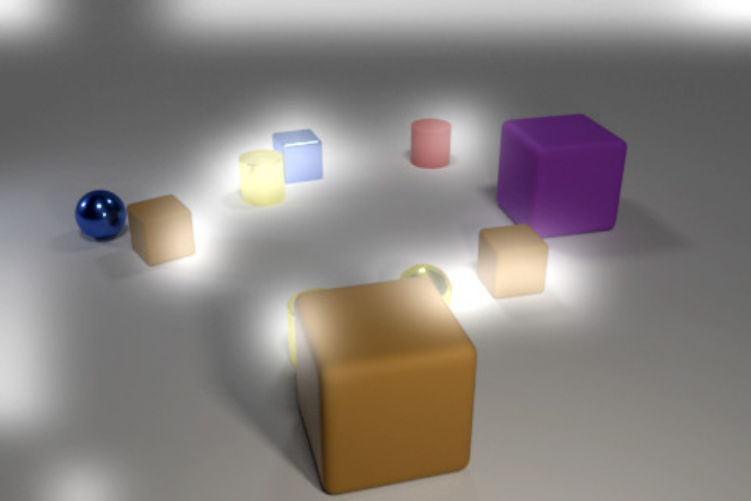}}
\hfill
\subfloat[{\normalfont{\textbf{Q}:}} What is the shape of the object that is to the left of the red rubber cube and behind the metallic cylinder? {\normalfont\textbf{A:}} sphere (cylinder) \textcolor{red}{\ding{55}}]{\includegraphics[width=0.24\linewidth]{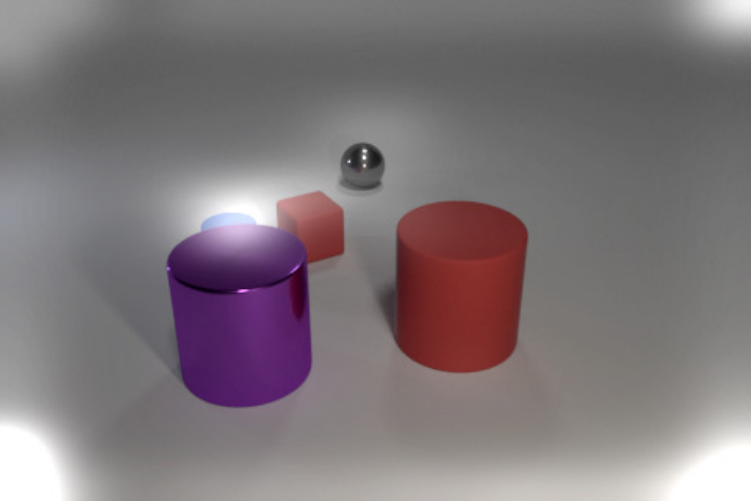}}
\hfill
\begin{minipage}{0.49\textwidth}
\vspace*{-20mm}
\subfloat{\includegraphics[width=0.49\linewidth]{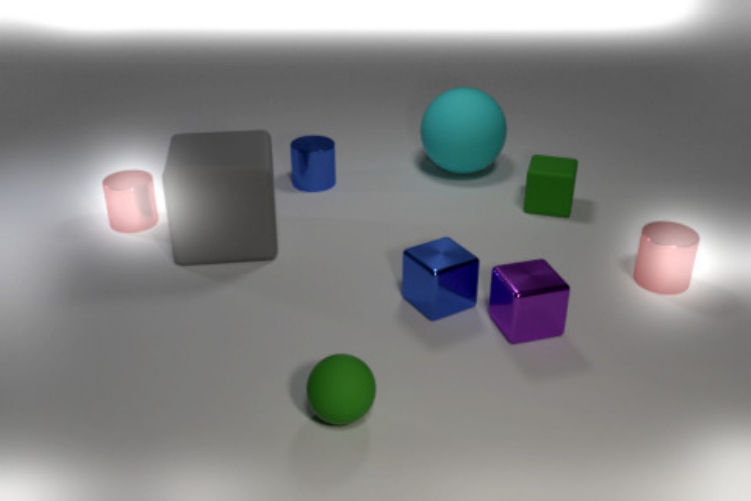}}
\hfill
\subfloat{\includegraphics[width=0.49\linewidth]{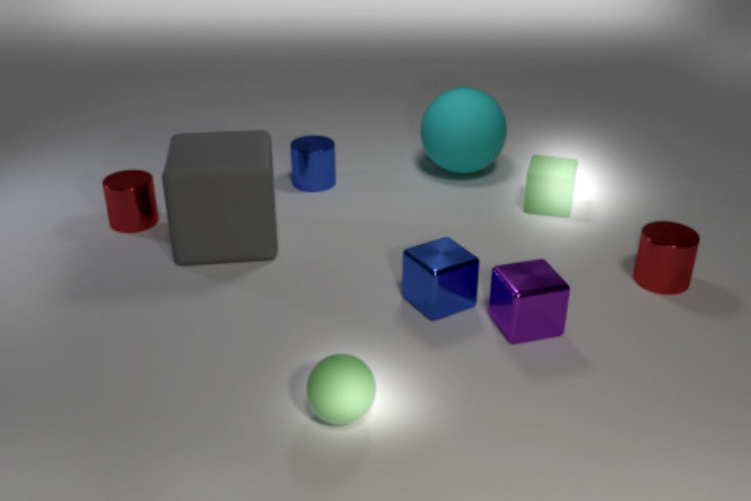}}

\scriptsize

{\normalfont{\textbf{Q}:}} How many tiny objects are green things or red objects? {\normalfont\textbf{A:}} 4 \textcolor{lime}{\ding{51}}
\end{minipage}

\caption{The first five rows show examples of the final attention map produced by the model for CLEVR-Human questions, demonstrating the ability of the model to perform novel reasoning skills and cope with new concepts that have not been introduced in CLEVR. These include in particular: \textcolor{purple}{\textbf{obstructions}}, \textcolor{blue}{\textbf{object uniqueness}}, \textcolor{green}{\textbf{relative distances}}, \textcolor{red}{\textbf{superlatives}} and \textcolor{yellow}{\textbf{new terms}}. The final row shows examples from CLEVR with object occlusions and summation.}
\label{fig:hexample_supp}
\label{hexample_supp}
\end{figure}

To gain insight into the model's failure cases, we perform error analysis for the CLEVR and CLEVR-Humans datasets. For CLEVR, we see that many of the errors arise from object occlusions, which may make it harder for the model to recognize the objects' material or shape. Most of the other errors are off-by-one counting mistakes, oftentimes for questions that ask to sum up two groups of objects (see examples in \figref{hexample_supp}). Interestingly, we noticed that when the model is required to count objects that are heavily occluded, it may lead the model to slightly underestimate the correct number of objects, suggesting that it may perform some sort of ``continuous" counting rather than discrete. 

For CLEVR-Humans, the errors made by the model are more diverse. About half of the errors result from questions about reasoning skills that the model has not been exposed to while training on CLEVR. These include questions about physical properties: lighting and shadows, reflections and objects stability (``How many of the items are casting a shadow?", ``Can a ball stay still on top of one another?"); relative distances (``How many objects\ldots are almost touching?", ``What color is the sphere positioned closest to\ldots", ``What object is in between\ldots"); relative amounts (``Are half the items\ldots", ``Are the objects mainly\ldots"); commonalities (``What color are the identical objects?", ``What shape do the items that \ldots have in common?"); and negative questions, which refer to objects that do not maintain some property (``How many items do not\ldots"). In addition, we observed some cases where the model misinterprets unseen words, capturing plausible but incorrect semantics: for instance, in some cases it interpreted a ``caramel" object as yellow whereas the original question referred to a brown one, or considered a cylinder to be ``circle" while the question referred to the spheres only. We believe that these errors may arise from diversity in the semantics of these words across the dataset, making the model learn those potentially valid but incorrect interpretations. Similarly to CLEVR, we observed some off-by-one errors in CLEVR-Humans. Finally, in one fifth of the cases, the errors result from faulty or ambiguous questions, which may mistakenly regard a cyan object as blue or mention references that cannot be uniquely resolved to a specific object.

\section{Ablation studies}
\label{sec:ablationsSupp} 

\begin{figure}[t]
\centering
\subfloat{\includegraphics[width=0.20\linewidth]{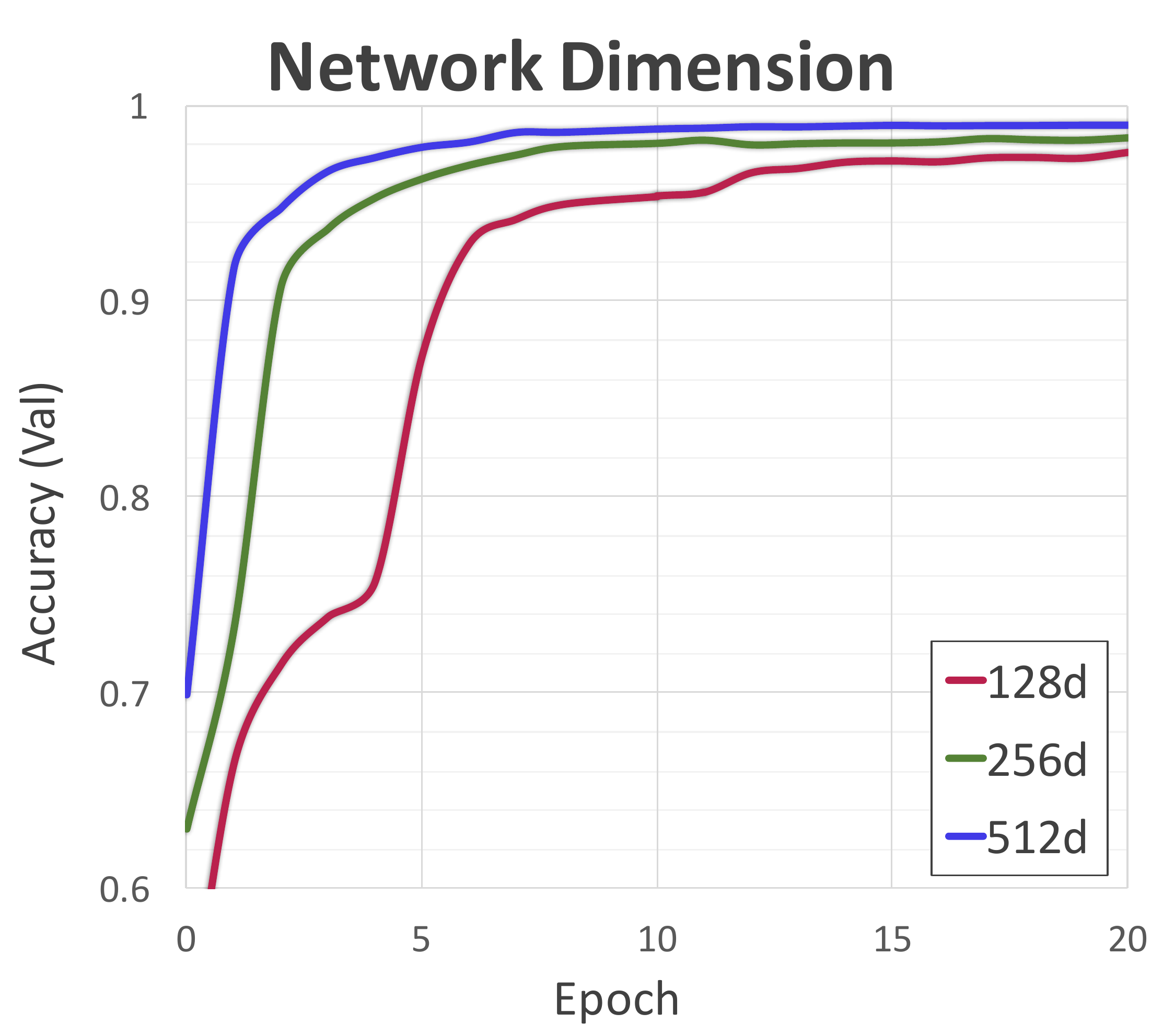}}
\hfill
\subfloat{\includegraphics[width=0.20\linewidth]{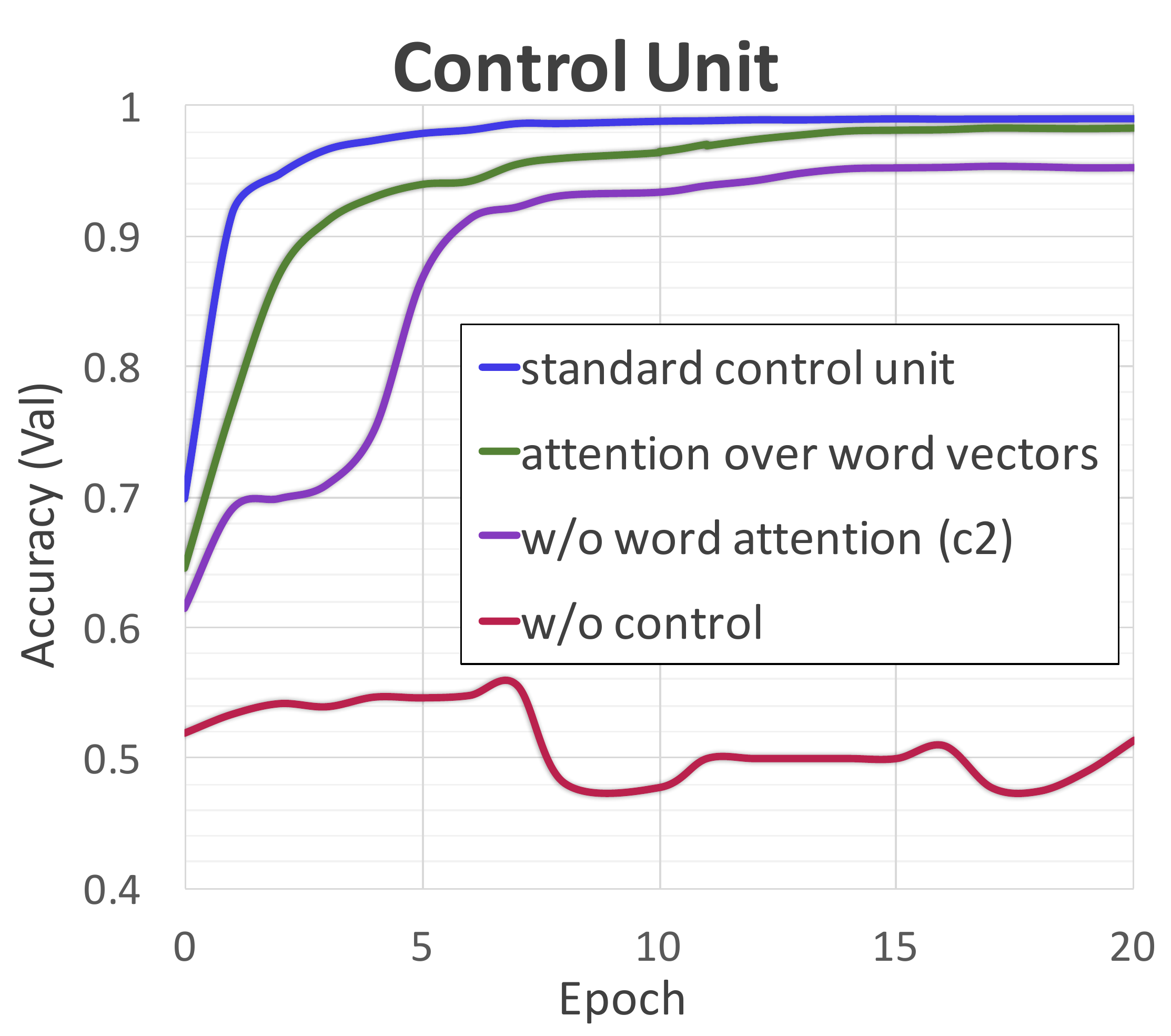}}
\hfill
\subfloat{\includegraphics[width=0.20\linewidth]{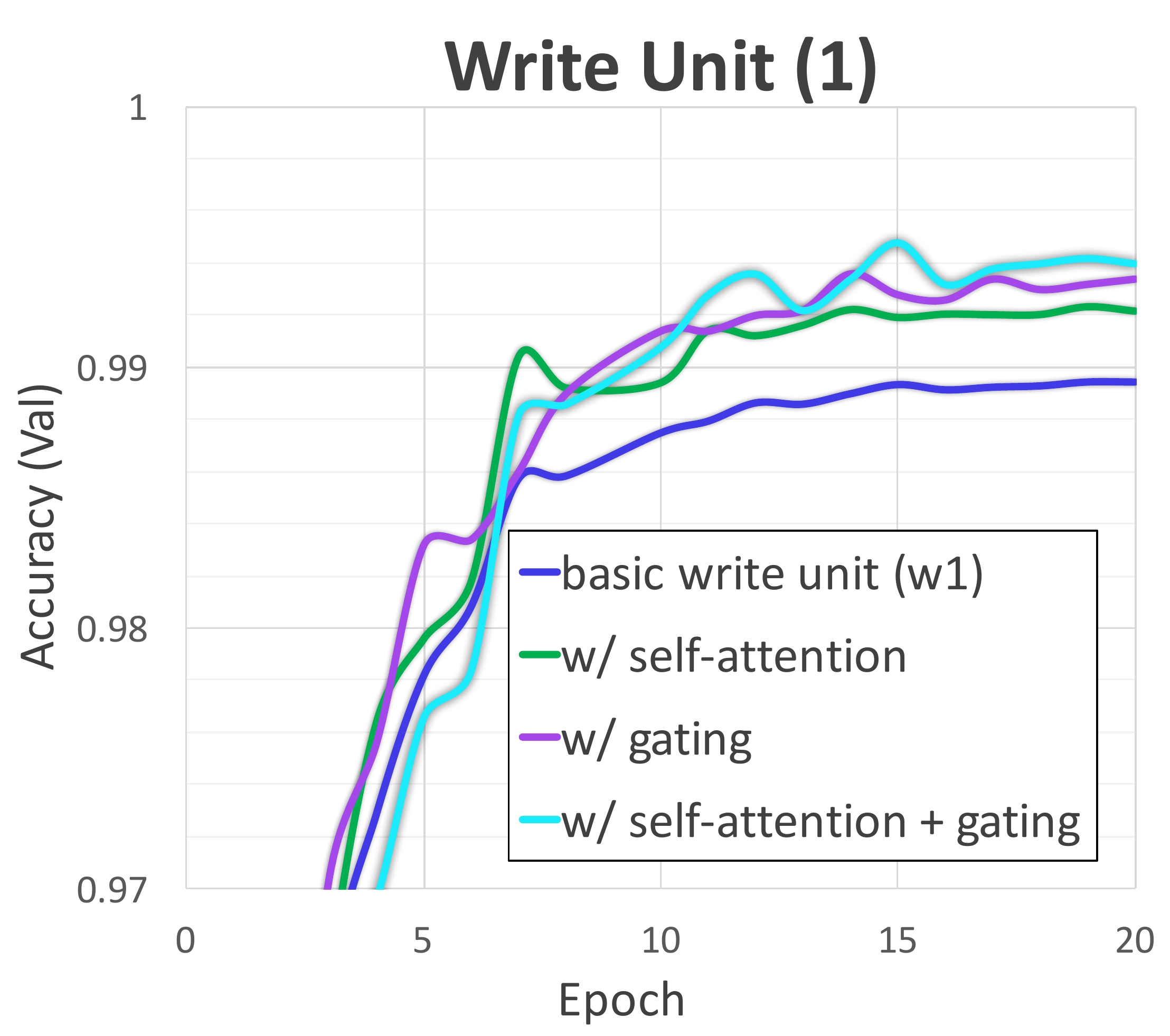}}
\hfill
\subfloat{\includegraphics[width=0.20\linewidth]{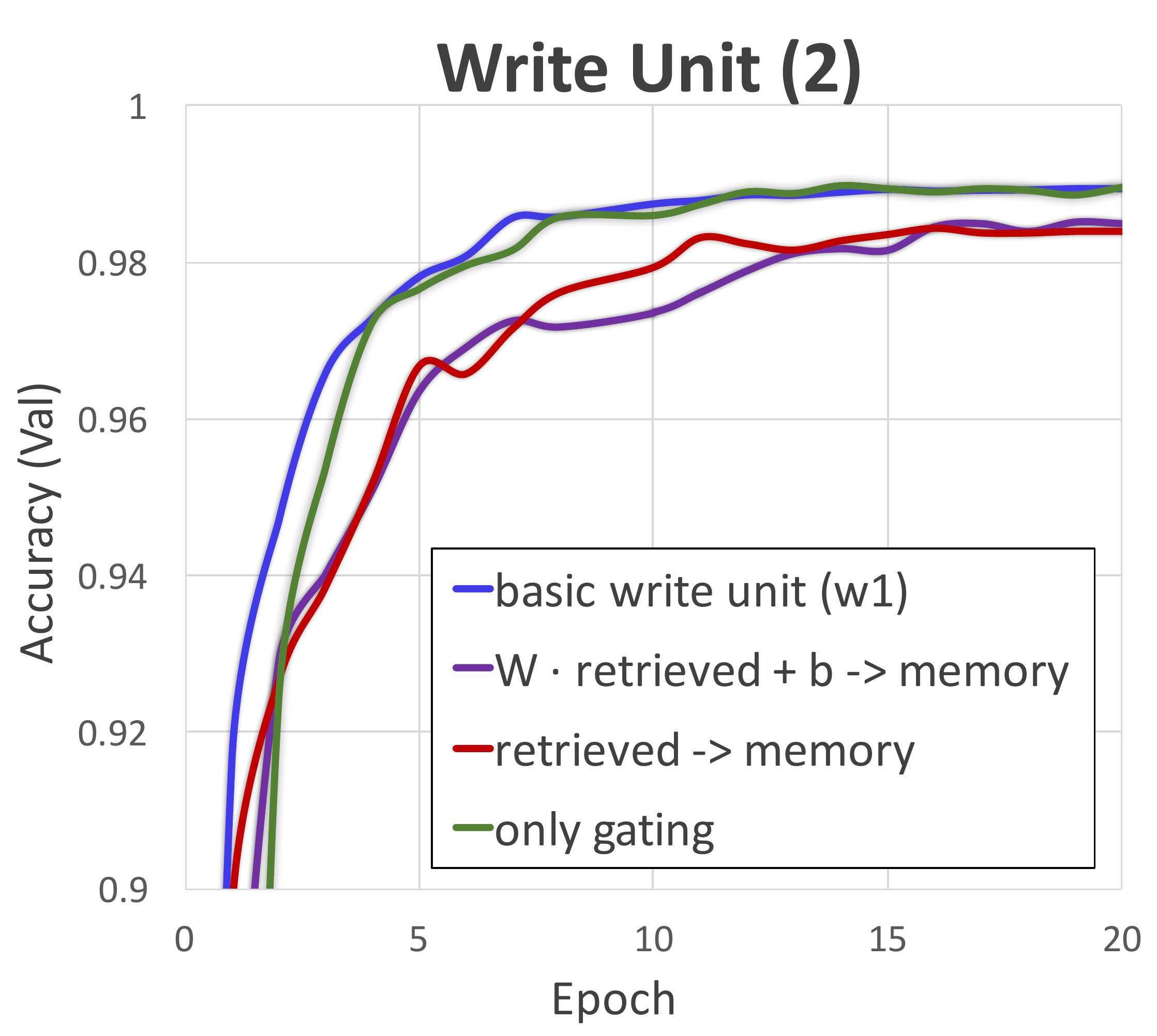}}
\hfill
\subfloat{\includegraphics[width=0.20\linewidth]{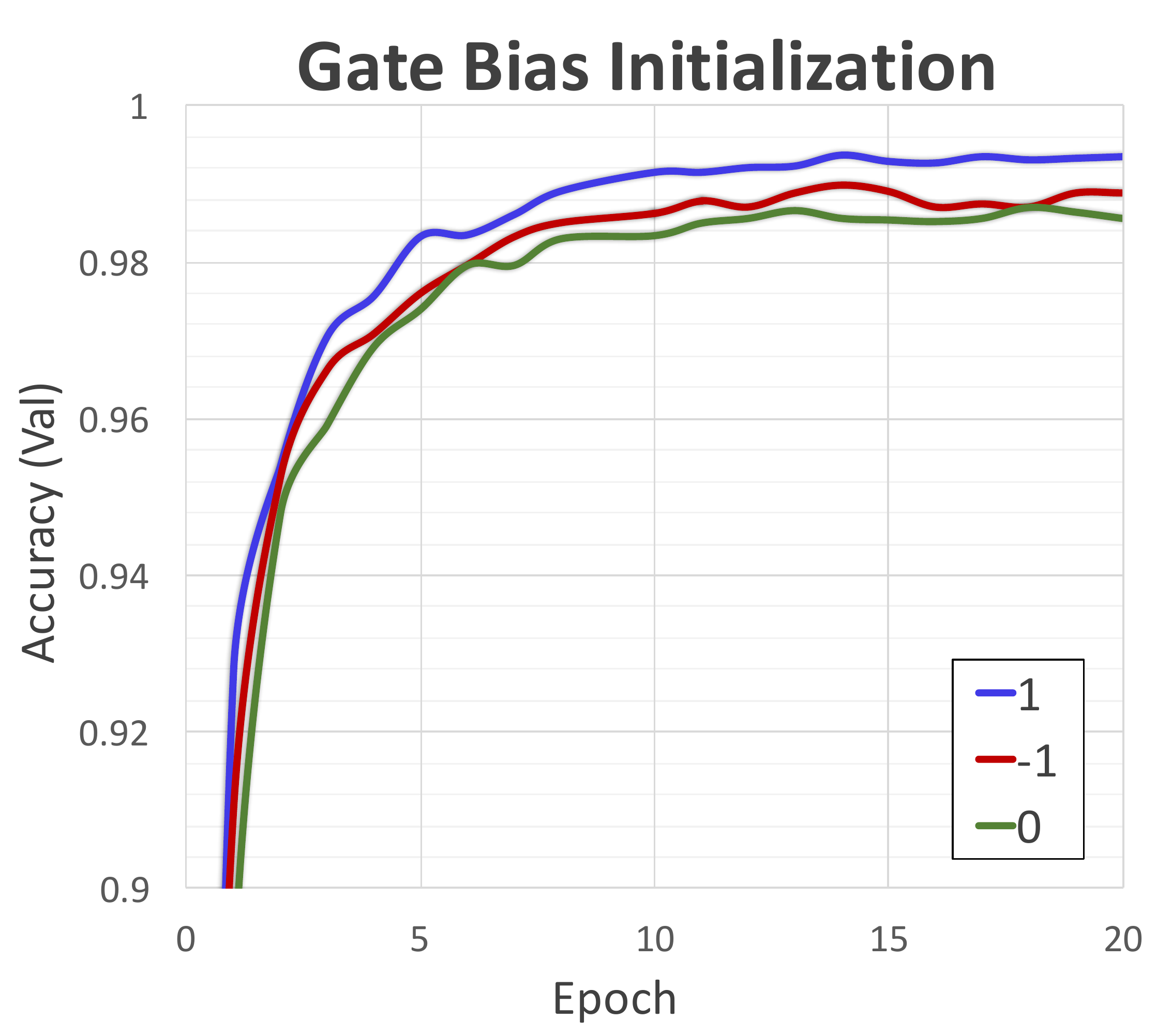}}
\caption{Learning curves for ablated MAC variants (accuracy / epoch). See \appref{ablationsSupp} for details.}
\label{fig:ablationsSupp}
\end{figure}

Based on the validation set, we have conducted an ablation study for MAC to better understand the impact of each of its components on the overall performance. We have tested each setting for the primary 700K CLEVR dataset as well as on a 10\% subset of it. See \tabref{table2}, \figref{plots} and \figref{ablationsSupp} for final accuracies and training curves. The following discussion complements the main conclusions presented in \secref{ablations}:

\textbf{Network Length.} We observe a positive correlation between the network length and its performance, with significant improvements up to length \(p=8\). These results stand out from other multi-hop architectures that tend to benefit from a lower number of iterations, commonly 2--3 only \citep{saAtt, dmn}, and suggest that MAC makes effective use of the recurrent cells to perform compositional reasoning.


\textbf{Weight Sharing.} Weight sharing across the \(p\) cell instances has also proven to be useful both for the primary CLEVR task as well as for settings with limited data. 
In contrast to alternative approaches that apply specialized modules for different parts of the reasoning, these results provide some evidence for the ability of the same MAC cell to adapt its behavior to the task at hand and demonstrate different behaviors as inferred from the context.

\textbf{Control Unit.} We have performed several ablations in the control unit architecture to identify its contribution to the model behavior and performance. First, we observe that, as would be expected, an ablated model that reasons over the image alone with no access to the question, performs poorly, achieving accuracy of 51.1\%. As discussed in \secref{ablations}, the ablations further show the importance of applying attention over the question words to decompose it into an explicit sequence of steps. Finally, we find that using the ``contextual words" -- the output states of a biLSTM processing the question -- results in better performance and faster learning than attending directly to the learned word vectors. This implies that the model benefits from interpreting the word semantics and entailed behaviors in the broader context of the question rather than as a sequence of independent entities.

\textbf{Write Unit.} The basic MAC write unit integrates new information \(\boldsymbol{r_{i}}\) with the previous memory state \(\boldsymbol{m_{i-1}}\) through a linear transformation (step (1) in \secref{WU}). In this experiment, we explore other variants of the unit. We begin by measuring the impact of the self-attention and gating mechanisms, both aiming to reduce long-range dependencies in the reasoning process. Compared to the basic MAC model, which achieves 98.94\% on the validation set, self-attention yields an accuracy of 99.23\%, memory gating -- 99.36\%, and adding both results in 99.48\%. While we can see some gain from using these components for CLEVR, we speculate that they may prove more useful for tasks that necessitate longer or more complex reasoning processes over larger knowledge bases. 

Next, we examine ablated write unit variants that assign the newly retrieved information \(\boldsymbol{r_{i}}\) (or its linear transformation) to \(\boldsymbol{m_{i}}\) directly, ignoring the prior memory content \(\boldsymbol{m_{i-1}}\). Notably, the results show that in fact such variants are only slightly worse than the default basic write unit, reducing accuracy by 0.4\% only. We perform further experiments in which we compute the new memory state \(\boldsymbol{m_{i}}\) by averaging the retrieved information \(\boldsymbol{r_{i}}\) with the previous memory state \(\boldsymbol{m_{i-1}}\) using a sigmoidal gate alone. This architecture results in equivalent performance to that of the standard basic write unit variant.

\textbf{Gate Bias Initialization.} Finally, we test the impact of the gate bias (step (3) in \secref{WU}), initializing it to either $-1$, 0 or 1. Intuitively, initialization of $-1$ amounts to biasing the model to retain previous memory states and thereby shortening  the effective reasoning process, while initialization of 1 is equivalent to using all new intermediate results derived by each of the cells. The experiments show that a bias of 1 is optimal for training on the full dataset while 0 is ideal for settings of limited data (training on 10\% of the data). These results demonstrate that when enough data is available, the MAC network benefits from utilizing its full capacity,  whereas biasing the model towards using less cells helps to mitigate overfitting when data is more scarce. 

\begin{table}
\caption{Accuracies for ablated MAC models, measured for the validation set after training on the full CLEVR dataset (left) and 10\% subset of it (right).}
\label{tab:table2}
\centering
\scriptsize
\begin{tabular}{lcc}
\rowcolor{Blue1}
\textbf{Model} & \textbf{Standard CLEVR} & \textbf{10\% CLEVR} \\
MAC & 98.9 & 84.5  \\
\rowcolor{Blue2}
state dimension 256 & 98.4 & 76.3  \\
\rowcolor{Blue2}
state dimension 128 & 97.6 & 77.0  \\
unshared weights & 97.8 & 67.5  \\
\rowcolor{Blue2}
attention over word vectors  & 98.3 & 61.4  \\
\rowcolor{Blue2}
w/o word-attention & 95.3 & 63.2  \\
\rowcolor{Blue2}
question vector as control & 80.7 & 65.0  \\
\rowcolor{Blue2}
w/o control & 55.6 & 51.5  \\
w/o memory-control separation  & 93.9 & 64.7  \\
w/o direct KB elements  & 98.4 & 73.4  \\
\rowcolor{Blue2}
retrieved $\rightarrow$ memory & 98.2 & 84.5  \\
\rowcolor{Blue2}
W $\cdot$ retrieved + b $\rightarrow$ memory &  98.5 &  83.7 \\
\rowcolor{Blue2}
only memory gate & 99.3 & 83.1  \\
w/ self-attention & 99.2 & 83.2   \\
w/ memory gate & 99.4 & 83.1  \\
w/ self-attention and memory gate & 99.5 & 85.5  \\
\rowcolor{Blue2}
gate bias 0 & 98.7 & 84.9  \\
\rowcolor{Blue2}
gate bias 1 & 99.4 & 68.5  \\
\rowcolor{Blue2}
gate bias $-1$ & 99.0 & 77.1  \\
prediction w/o question & 97.8 & 64.7
\end{tabular}
\end{table}

\section{Related Work}
\label{sec:related}

In this section we provide detailed discussion of related work. Several models have been applied
to the CLEVR task. These can be partitioned into two groups, module networks that use the strong
supervision provided as a tree-structured functional program associated with each instance, and
end-to-end, fully differentiable networks that combine a fairly standard stack of CNNs with components
that aid them in performing reasoning tasks. We also discuss the relation of MAC to other
approaches, such as memory networks and neural computers.

\subsection{Module Networks}

The modular approach \citep{nmn,nmn2,nmn3,pgee} first translates a given question into a tree-structured action plan, aiming to imitate the question underlying structural representation externally provided as strong supervision. Then, it constructs a tailor-made network that progressively executes the plan over the image. The network is composed of discrete units selected out of a fixed collection of predefined ``modules", each responsible for an elementary reasoning operation, such as identifying an object's color, filtering them for their shape, or comparing their amounts. Each module has its own set of learned parameters \citep{pgee}, or even a hand-crafted design \citep{nmn} that guides it towards its intended behavior. Overall, this approach makes discrete choices at two levels: the identity of each module -- the behavior it should learn among a fixed set of possible behavior types, and the network layout -- the way in which the modules are wired together to compute the answer. The model differentiability is thus confined to the boundaries of a single module. 

Several key differences exist between our approaches. First, MAC replaces the fixed and specialized modules inventory with one universal cell that adapts its operation to the task at hand, selected from a continuous range of reasoning behaviors. Therefore, in contrast to module networks, our cell can be applied across all the reasoning steps, sharing both its parameters and architecture. Second, we replace the dynamic recursive tree structures with a sequential topology, augmented by soft attention mechanisms, inspired by \citet{alignTrans}. This confers the network with the capacity to represent arbitrarily complex Directed Acyclic Graphs (DAGs) in a soft way, while still having efficient and readily deployed physical sequential structure. Together, these relaxations allow us to effectively train our model end-to-end by backpropagation alone, whereas module networks demand more involved training schemes that rely on the strongly-supervised programs at the first stage, and on various reinforcement learning (RL) techniques at the second. Finally, since the only source of supervision in training our model arises from the answer to each question, MAC is free to acquire more robust and adaptive reasoning strategies from the bottom up -- inferred directly form the data, rather than trying to imitate the behaviors dictated by brittle parsers or closed domain functional programs, and is thus more applicable to real-world settings.    

\subsection{Augmented Convolutional Neural Networks}
Alternative approaches for the CLEVR task that do not rely on the provided programs as a strong supervision signal are \citet{rn} and \citet{film}. Both complement standard multi-layer Convolutional Neural Networks (CNNs) with components that aid them in handling compositional and relational questions.

\textbf{Relation Networks.} \citet{rn} appends a Relation Network (RN) layer to the CNN. This layer inspects all pairs of pixels in the image, thereby enhancing the network capacity to reason over binary relations between objects. While this approach is very simple and elegant conceptually, it suffers from quadratic computational complexity, in contrast to our approach, which is linear. But beyond that, closer inspection reveals that this direct pairwise comparison might be unnecessary. Based on the analogy suggested by \citet{rn}, according to which pixels are equivalent to objects and their pairwise interactions to relations, an RN layer attempts to grasp the induced graph between objects all at once in one shallow and broad layer. Conversely, our attention-based model proceeds in steps, iteratively comparing the image to a memory state that had aggregated information from the image in prior iterations. By the same analogy, MAC traverses a narrow and deep reasoning ``path" that progressively follows transitive relations. Consequently, our model exhibits a relational capacity while circumventing the computational inefficiency.

\textbf{FiLM.} \citet{film} proposes a model for visual reasoning that interleaves standard CNN layers with linear layers, reminiscent of layer normalization techniques \citep{ln,bn}. Each of these layers, called FiLM, is conditioned on the question, which is translated into matching bias and variance terms that tilt the layer's activations to reflect the specifics of the given question, thus influencing the computation done over the image. Similarly to our model, this approach features distant modulation between the question and the image, where the former can affect the latter only through constrained means. However, since the same normalization is applied across all the activations homogeneously, agnostic to both their spatial location as well as their feature values, FiLM does not allow the question to differentiate between regions in the image based on their semantics -- the objects or concepts they represent. 

This stands in stark contrast to our attention-based model, which readily allows and actually encourages the question to inform the model about relevant regions to focus on. As supported by \secref{experiments}, this more selective interaction between the question and the image facilitates learning and increases the model's generalizability. Indeed, since attention is commonly used in models designed for standard VQA \citep{vqa, vqaSurv,coatt,saAtt}, it is reasonable to assume that it would be beneficial to incorporate such methods into visual reasoning systems for the CLEVR task as well. In fact, attention mechanisms should be especially useful for multi-step reasoning questions such as those present in CLEVR. Such questions refer to several relations between different objects in the image and feature compositional structure that may be approached one step at a time. Thus, it should be beneficial for a cogent responder to have the capacity to selectively focus on one or some objects at each step, traversing the relevant relational links one after the other, both at the image level, and at the question level.

\subsection{Memory and Attention}
Our architecture draws inspiration from recent research on mechanisms for neural memory and attention \citep{dmn,vdmn,ntm,dnc}. \citet{dmn} and \citet{vdmn} propose the Dynamic Memory Network (DMN) model that proceeds in an iterative process, attending to relevant information from a given knowledge base, which is then successively accumulated into the model's memory state. However, the DMN views the question as one atomic unit, whereas our model decomposes it into a multi-step action plan that informs each cell of its specific objective. Another key difference is the distant interaction between the question and the knowledge base that characterizes our model. Conversely, DMN fuses their representations together into the same vector space.

\citet{ntm,dnc} complement a neural network with an external memory it can interact with through the means of soft attention. Similarly to our approach, the model consists of a controller that performs read and write operations over a fixed-size memory array. However, in contrast to \citet{ntm,dnc}, we employ a recurrent memory structure, where each MAC cell is associated with its own memory state. Rather than reading and writing iteratively into multiple slots in a shared memory resource, each cell creates a new memory, building upon the contents of the prior ones. This allows us to avoid potential issues of content blurring due to multiple global write operations, while still supporting the emergence of complex reasoning processes that progressively interact with preceding memories and intermediate results to accomplish the task at hand.

\begin{figure}[t]
\captionsetup[subfloat]{labelformat=empty}
\centering
\begin{minipage}{0.11\textwidth}
\centering
\subfloat{\includegraphics[width=0.75\linewidth]{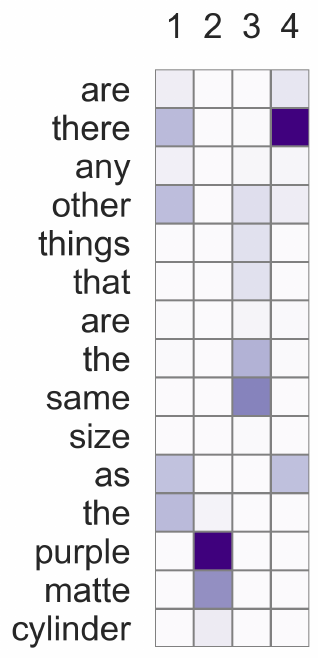}}
\end{minipage}
\begin{minipage}{0.88\textwidth}
\subfloat{\includegraphics[width=0.245\linewidth]{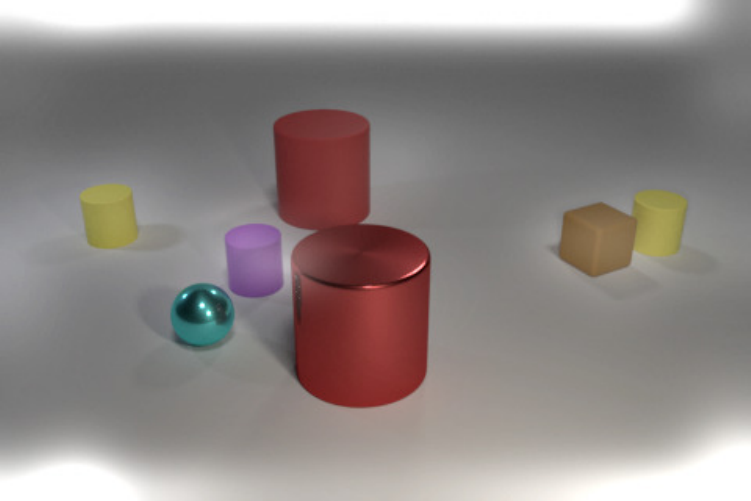}}
\hfill
\subfloat{\includegraphics[width=0.245\linewidth]{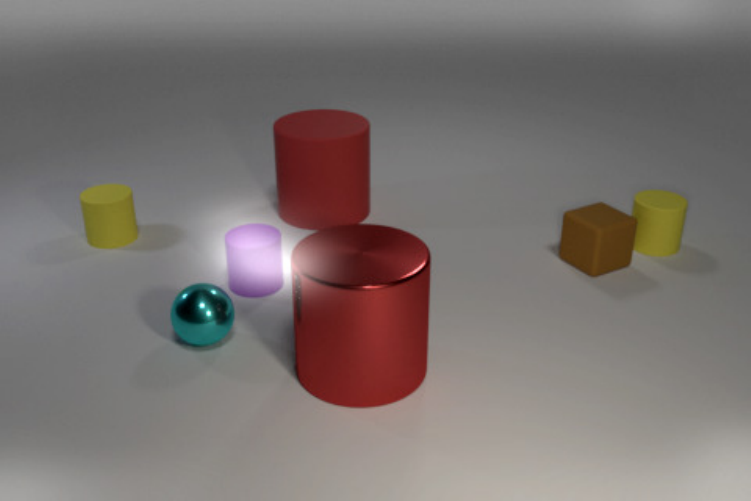}}
\hfill
\subfloat{\includegraphics[width=0.245\linewidth]{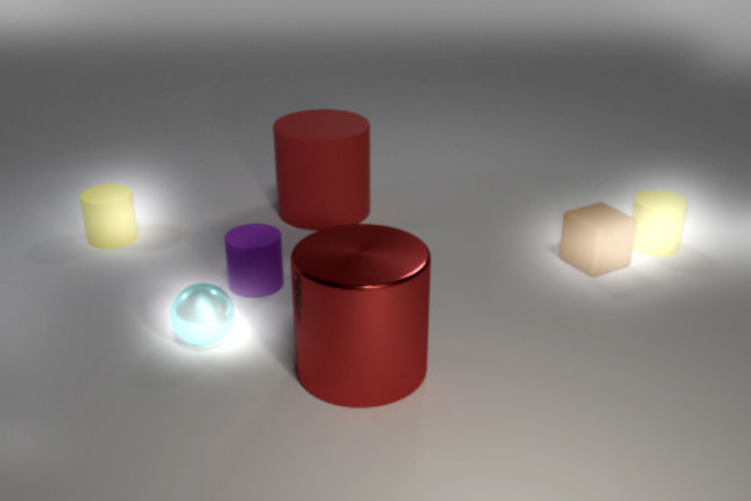}}
\hfill
\subfloat{\includegraphics[width=0.245\linewidth]{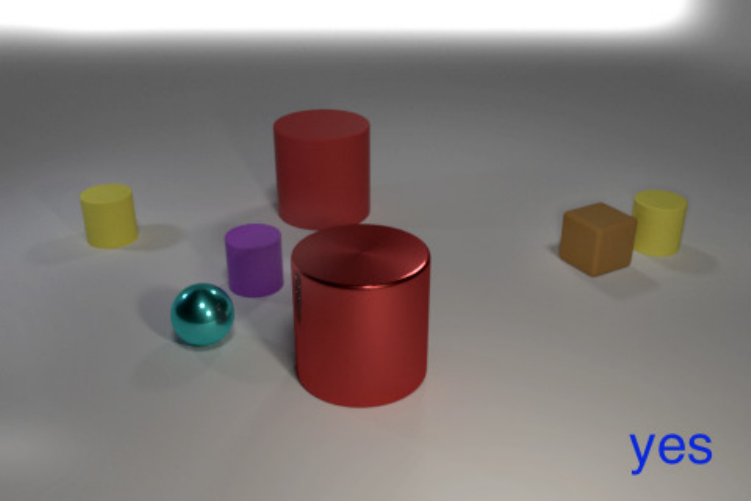}}
\end{minipage}

\begin{minipage}{0.11\textwidth}
\centering
\subfloat{\includegraphics[width=0.75\linewidth]{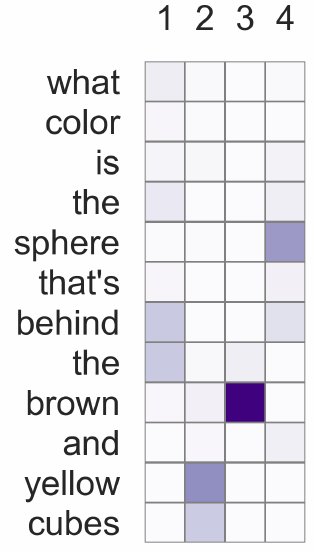}}
\end{minipage}
\begin{minipage}{0.88\textwidth}
\subfloat{\includegraphics[width=0.245\linewidth]{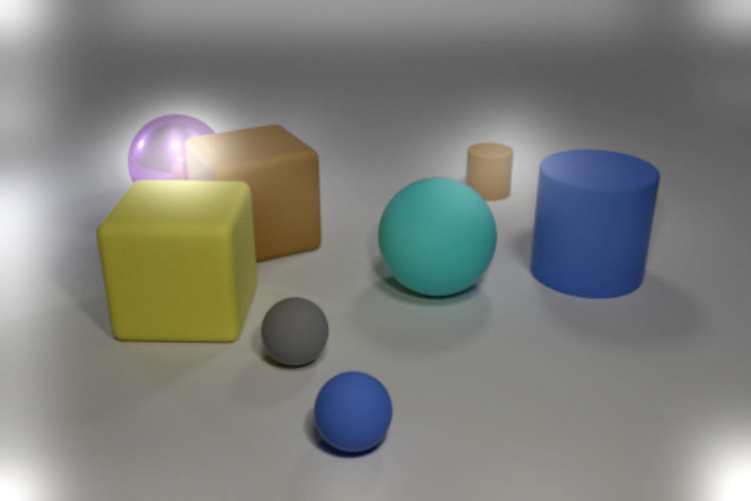}}
\hfill
\subfloat{\includegraphics[width=0.245\linewidth]{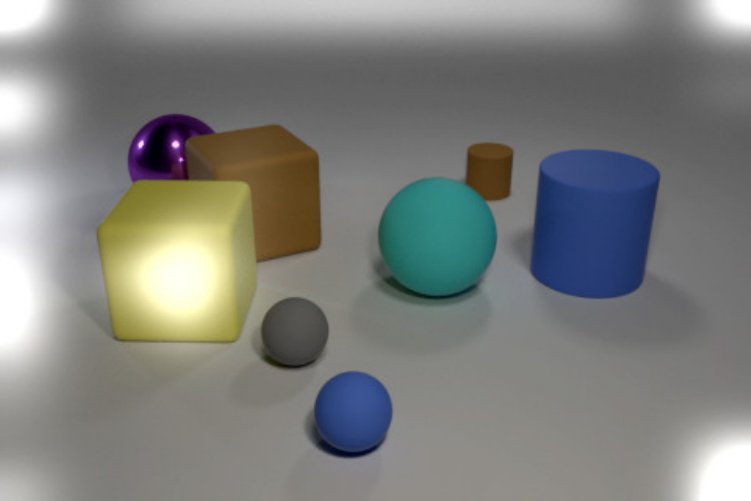}}
\hfill
\subfloat{\includegraphics[width=0.245\linewidth]{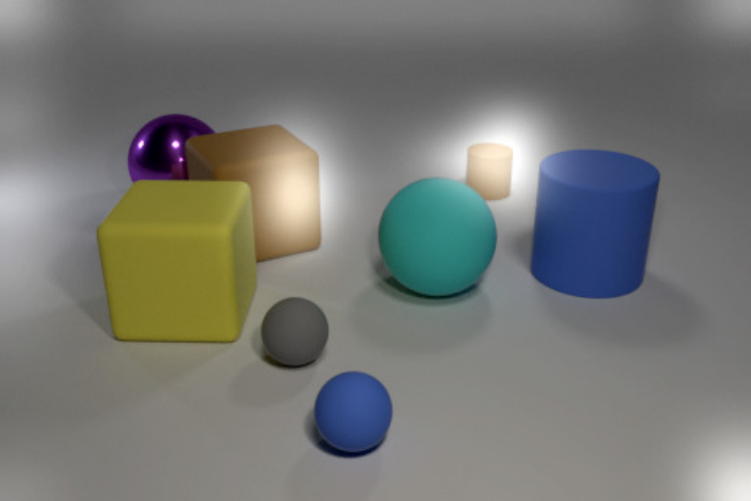}}
\hfill
\subfloat{\includegraphics[width=0.245\linewidth]{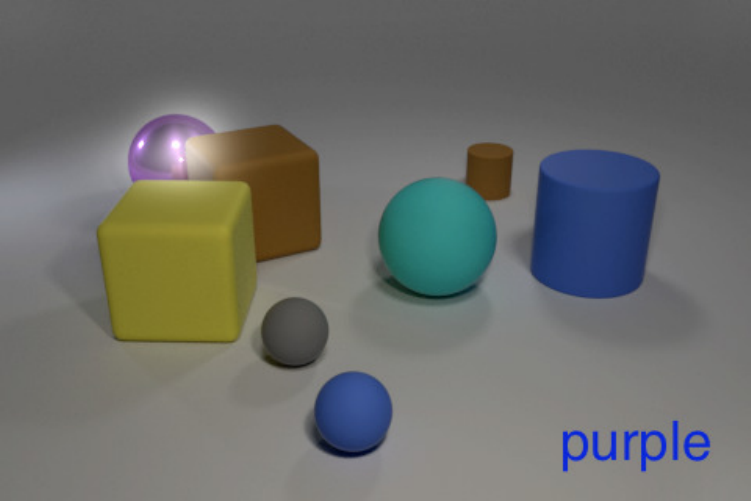}}
\end{minipage}

\begin{minipage}{0.11\textwidth}
\centering
\subfloat{\includegraphics[width=1.0\linewidth]{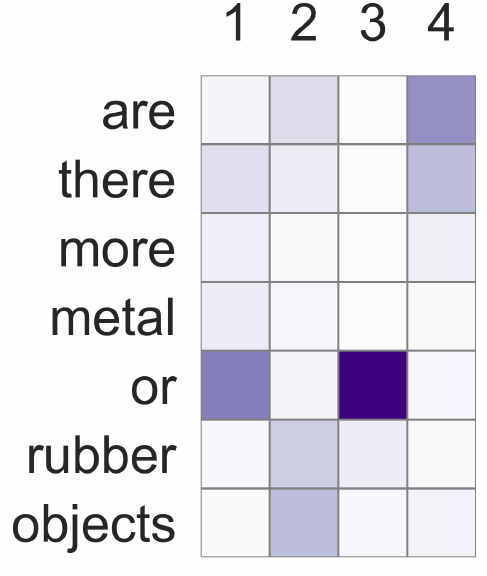}}
\end{minipage}
\begin{minipage}{0.88\textwidth}
\subfloat{\includegraphics[width=0.245\linewidth]{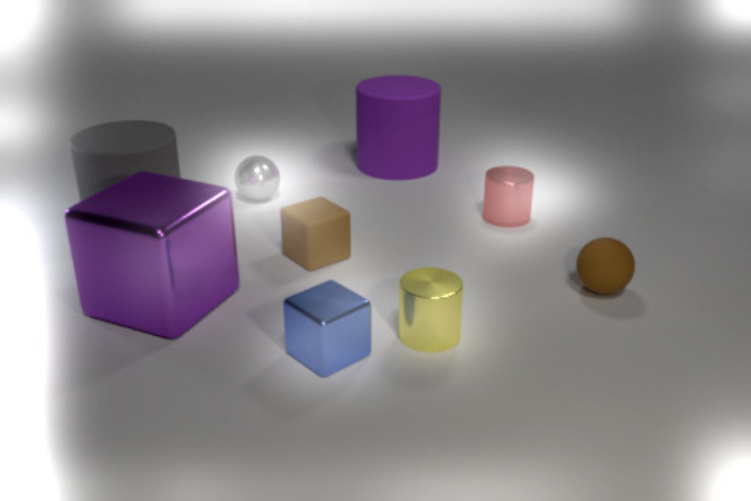}}
\hfill
\subfloat{\includegraphics[width=0.245\linewidth]{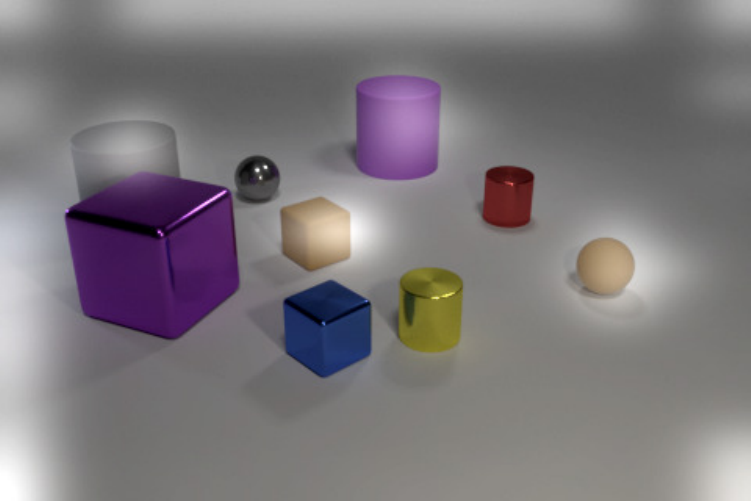}}
\hfill
\subfloat{\includegraphics[width=0.245\linewidth]{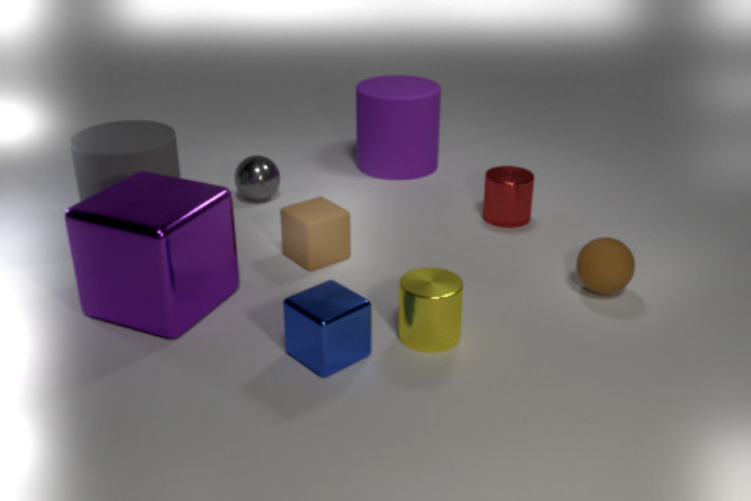}}
\hfill
\subfloat{\includegraphics[width=0.245\linewidth]{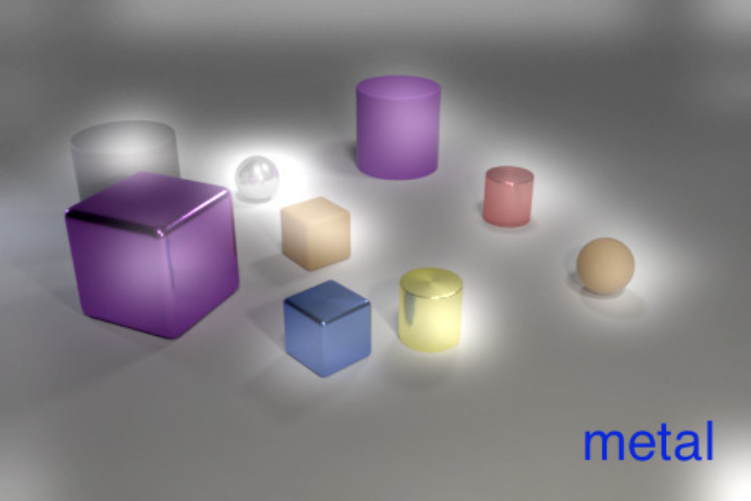}}
\end{minipage}

\begin{minipage}{0.11\textwidth}
\centering
\subfloat{\includegraphics[width=1.0\linewidth]{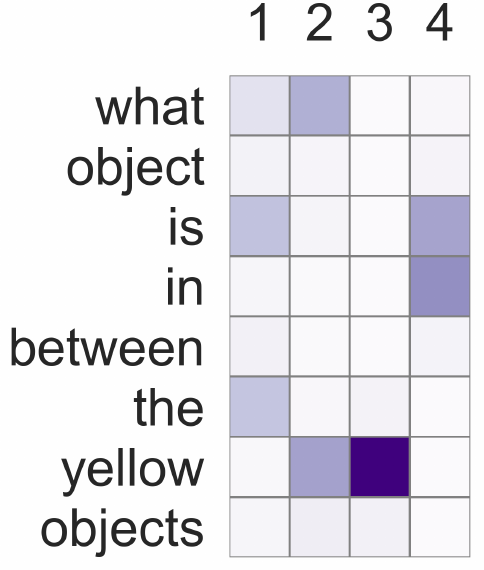}}
\end{minipage}
\begin{minipage}{0.88\textwidth}
\subfloat{\includegraphics[width=0.245\linewidth]{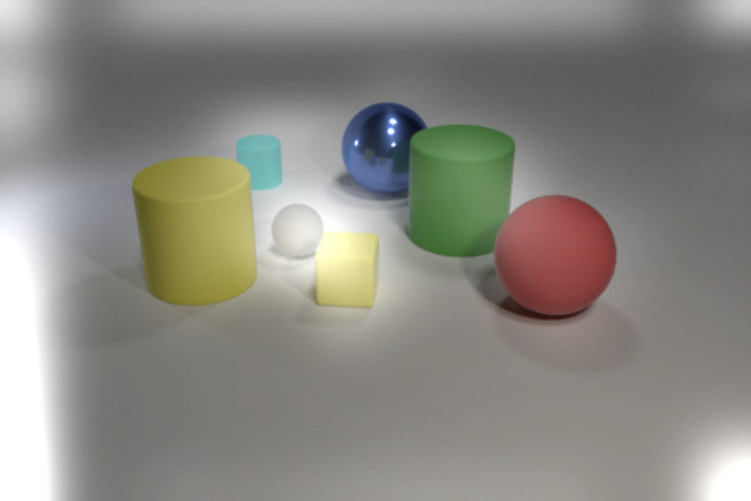}}
\hfill
\subfloat{\includegraphics[width=0.245\linewidth]{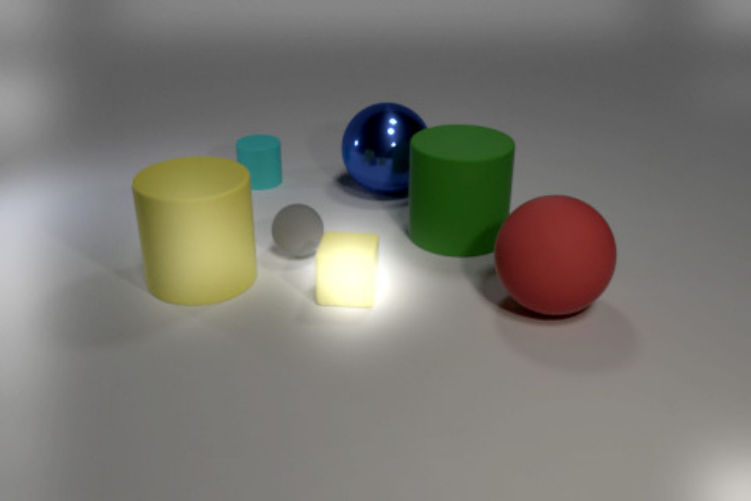}}
\hfill
\subfloat{\includegraphics[width=0.245\linewidth]{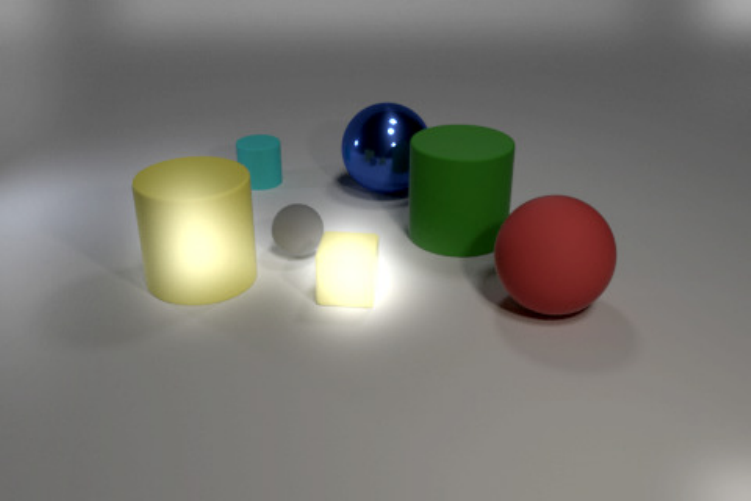}}
\hfill
\subfloat{\includegraphics[width=0.245\linewidth]{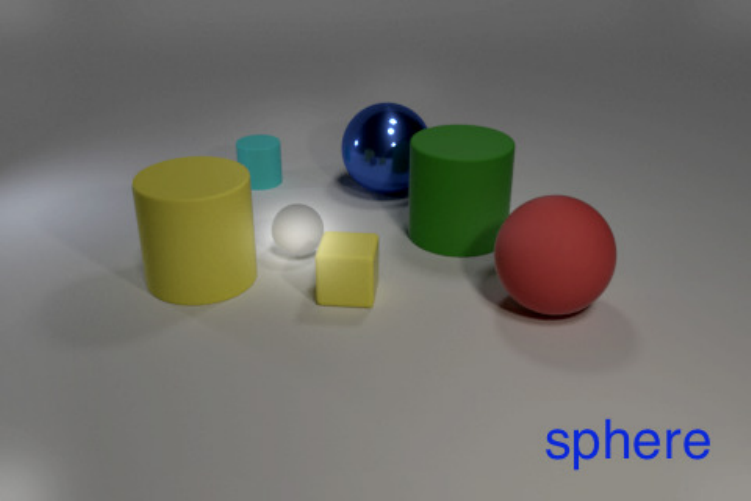}}
\end{minipage}

\begin{minipage}{0.16\textwidth}
\vspace*{2mm}
\subfloat{\includegraphics[width=1.0\linewidth]{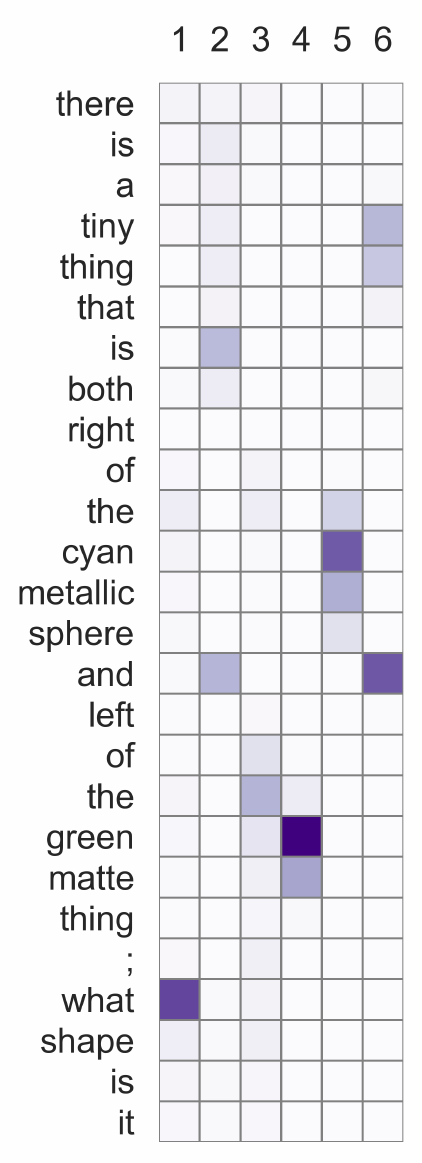}}
\end{minipage}
\begin{minipage}{0.82\textwidth}
\noindent
\centering
\subfloat{\includegraphics[width=0.325\linewidth]{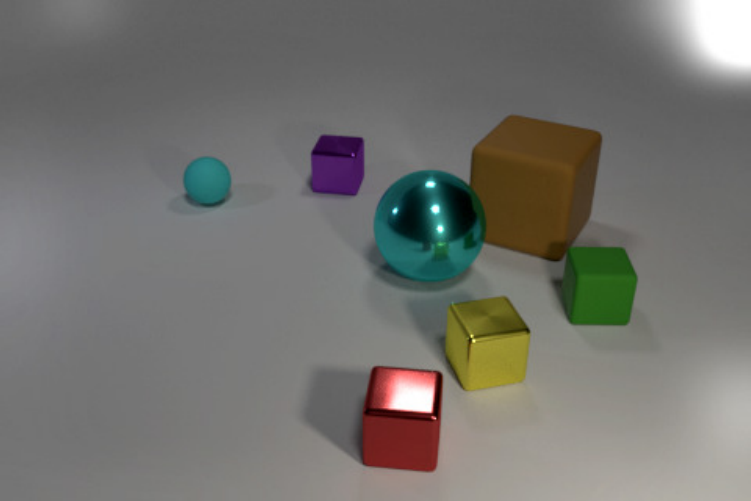}}
\hfill
\subfloat{\includegraphics[width=0.325\linewidth]{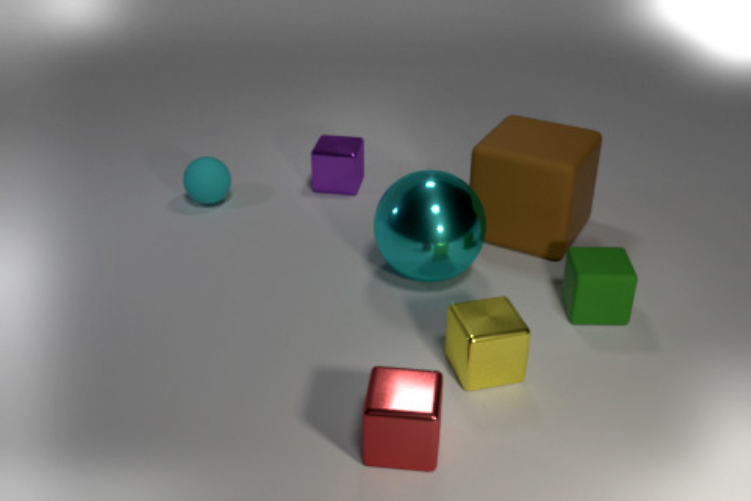}}
\hfill
\subfloat{\includegraphics[width=0.325\linewidth]{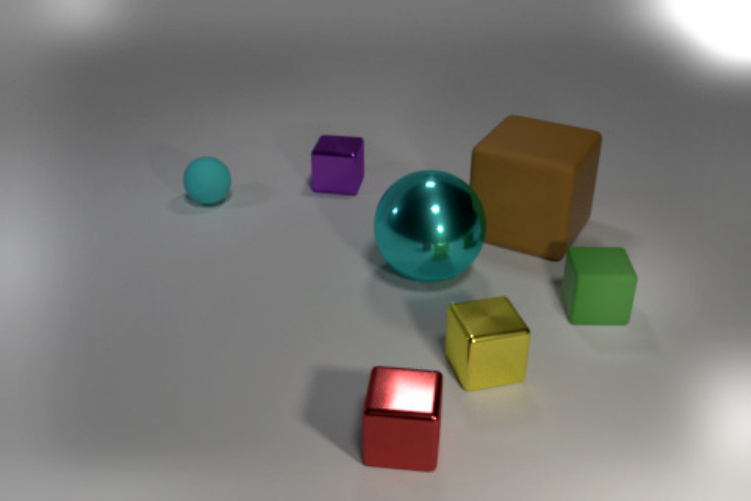}}
\hfill
\subfloat{\includegraphics[width=0.325\linewidth]{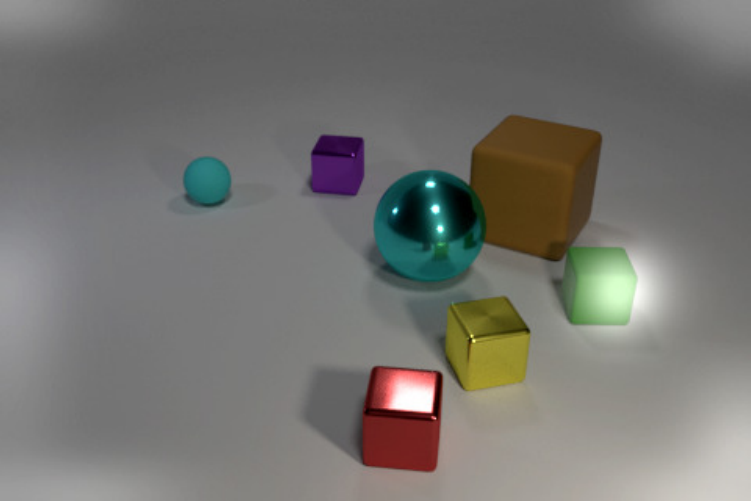}}
\hfill
\subfloat{\includegraphics[width=0.325\linewidth]{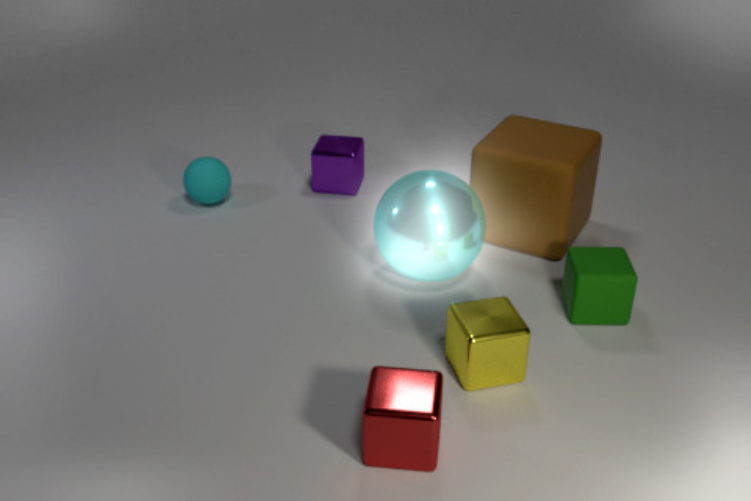}}
\hfill
\subfloat{\includegraphics[width=0.325\linewidth]{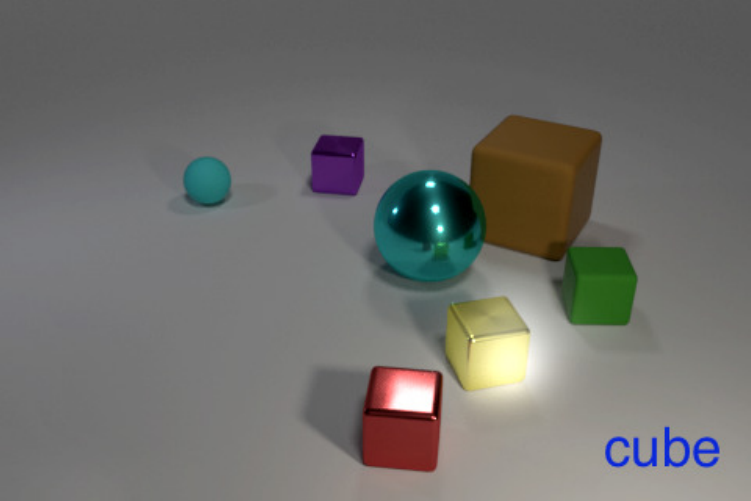}}
\end{minipage}

\begin{minipage}{0.1\textwidth}
\centering
\vspace*{4mm}
\subfloat{\includegraphics[width=0.9\linewidth]{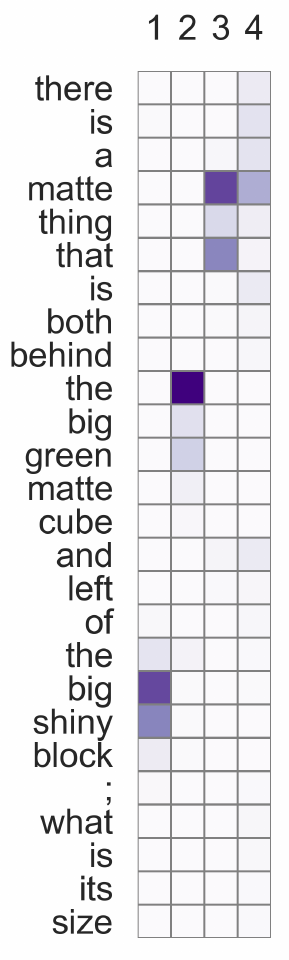}}
\end{minipage}
\begin{minipage}{0.38\textwidth}
\noindent
\centering
\subfloat{\includegraphics[width=0.49\linewidth]{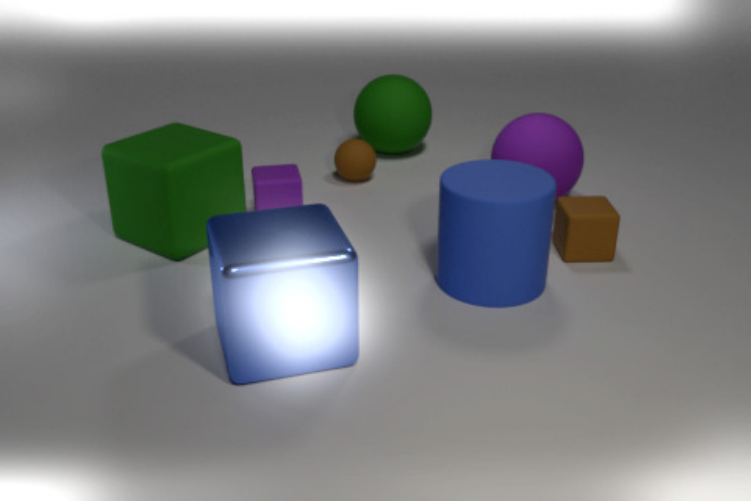}}
\hfill
\subfloat{\includegraphics[width=0.49\linewidth]{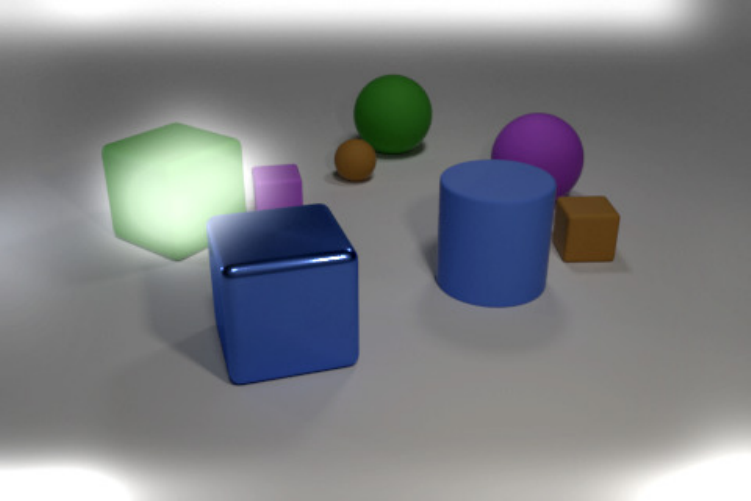}}
\hfill
\subfloat{\includegraphics[width=0.49\linewidth]{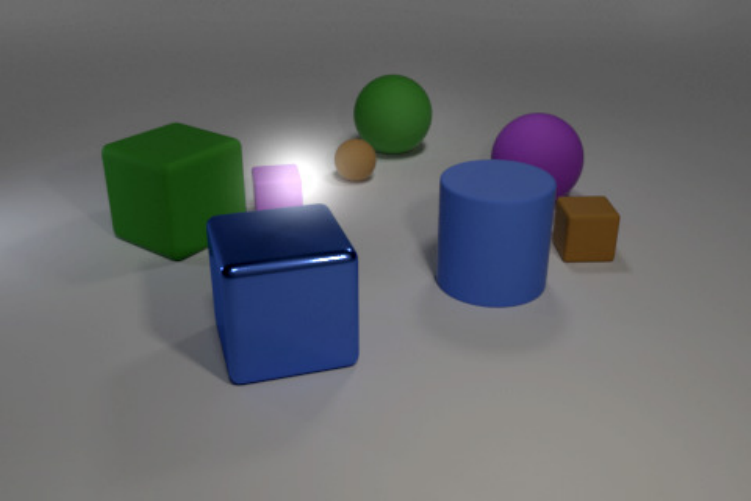}}
\hfill
\subfloat{\includegraphics[width=0.49\linewidth]{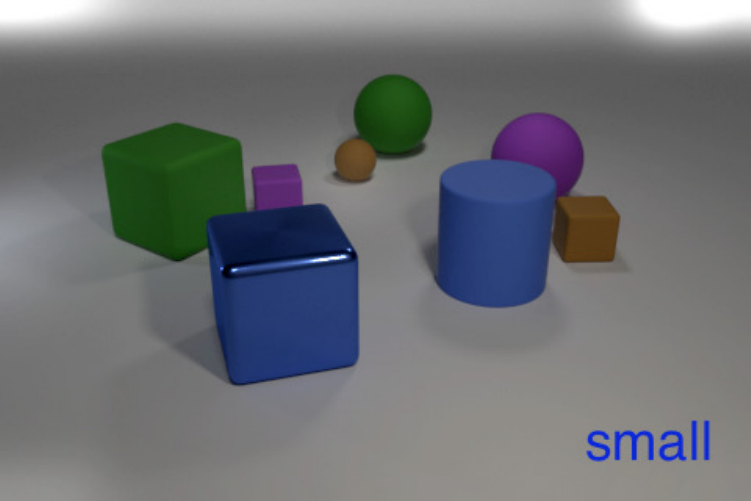}}

\end{minipage}
\begin{minipage}{0.1\textwidth}
\centering
\vspace*{4mm}
\subfloat{\includegraphics[width=1.0\linewidth]{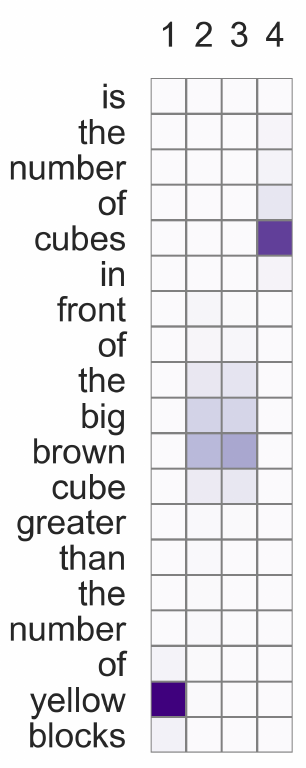}}
\end{minipage}
\begin{minipage}{0.38\textwidth}
\noindent
\centering
\subfloat{\includegraphics[width=0.49\linewidth]{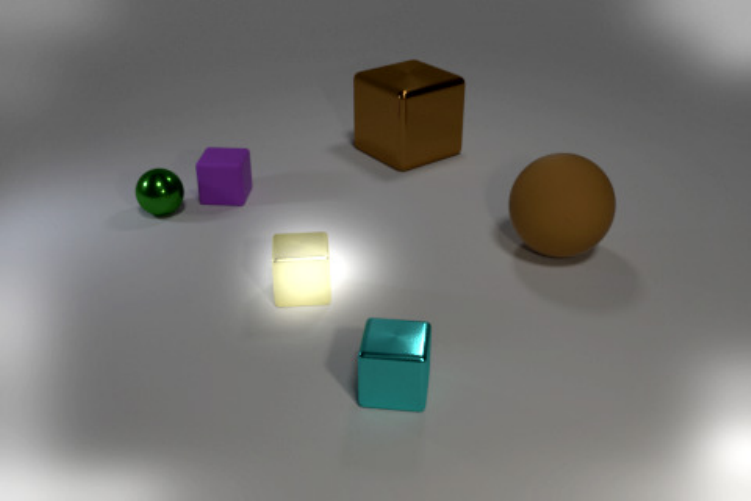}}
\hfill
\subfloat{\includegraphics[width=0.49\linewidth]{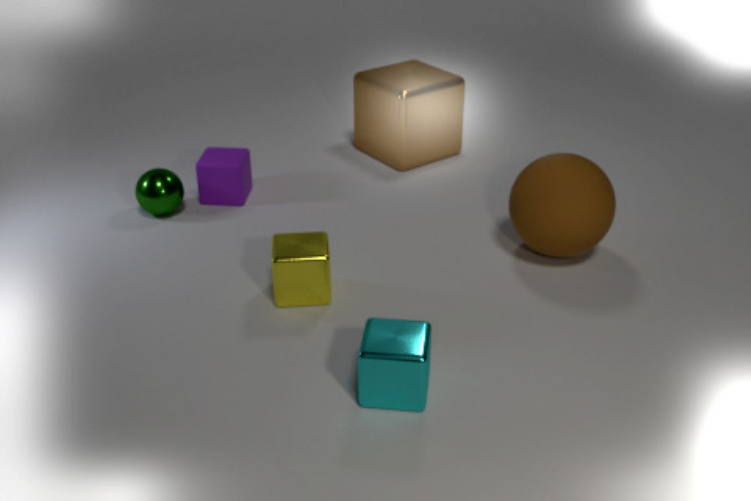}}
\hfill
\subfloat{\includegraphics[width=0.49\linewidth]{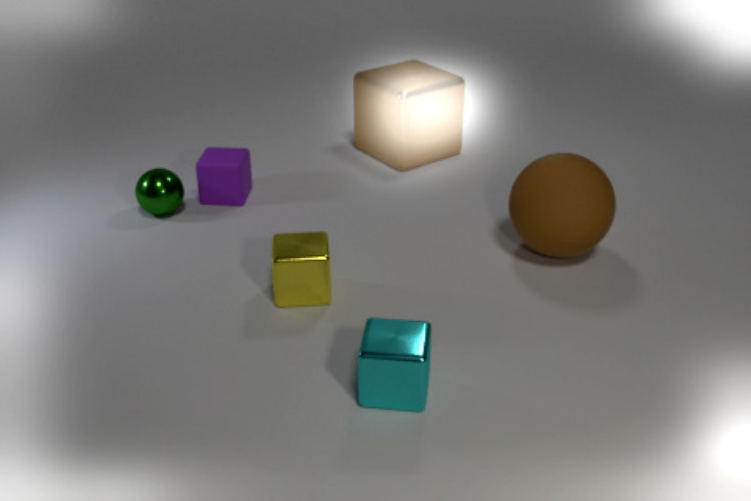}}
\hfill
\subfloat{\includegraphics[width=0.49\linewidth]{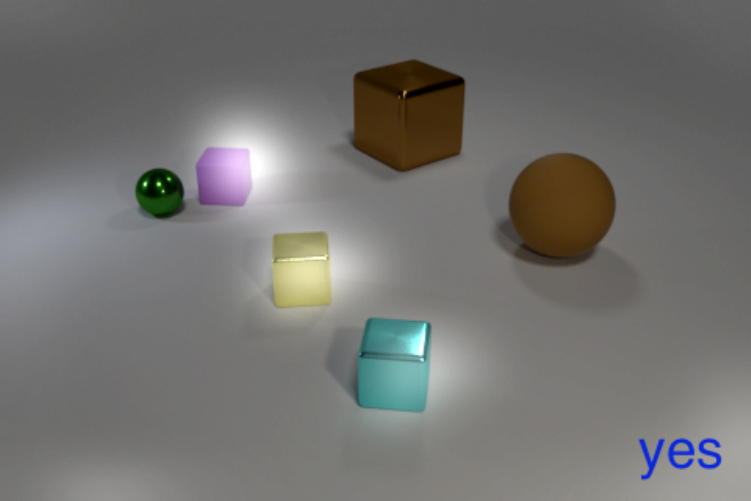}}
\end{minipage}

\caption{Attention maps produced by MAC networks of lengths 4 and 6, providing evidence for the ability of the model to track transitive relations and perform logical operations. Note how the model tends to proceed from the end of the question backwards, tracking the relevant objects iteratively.}
\label{vizz1}
\end{figure}


\begin{figure}[t]
\captionsetup[subfloat]{labelformat=empty}
\centering
\begin{minipage}{0.2\textwidth}
\centering
\vspace*{4mm}
\subfloat{\includegraphics[width=0.9\linewidth]{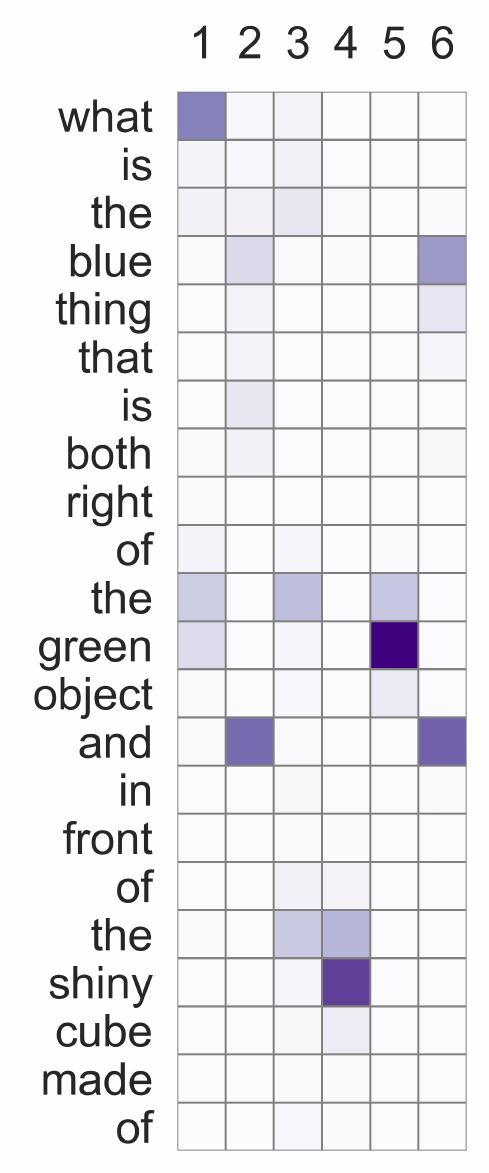}}
\end{minipage}
\begin{minipage}{0.79\textwidth}
\noindent
\centering
\subfloat{\includegraphics[width=0.325\linewidth]{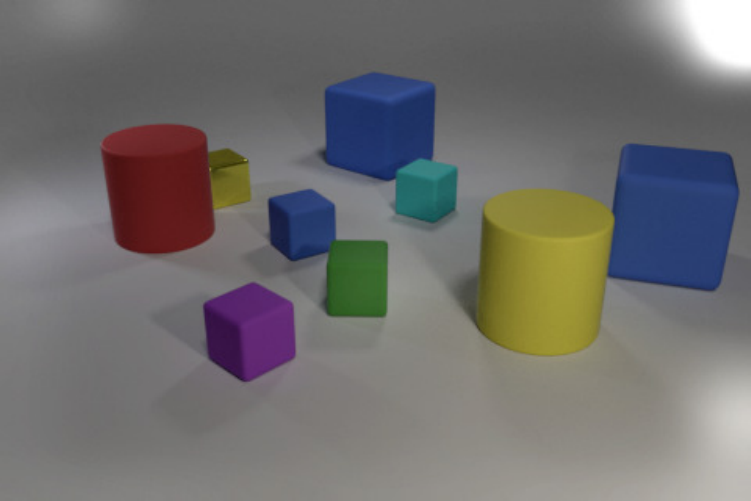}}
\hfill
\subfloat{\includegraphics[width=0.325\linewidth]{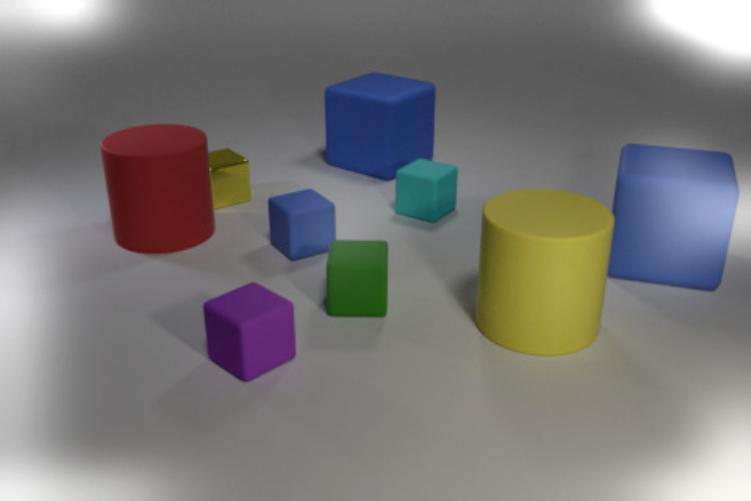}}
\hfill
\subfloat{\includegraphics[width=0.325\linewidth]{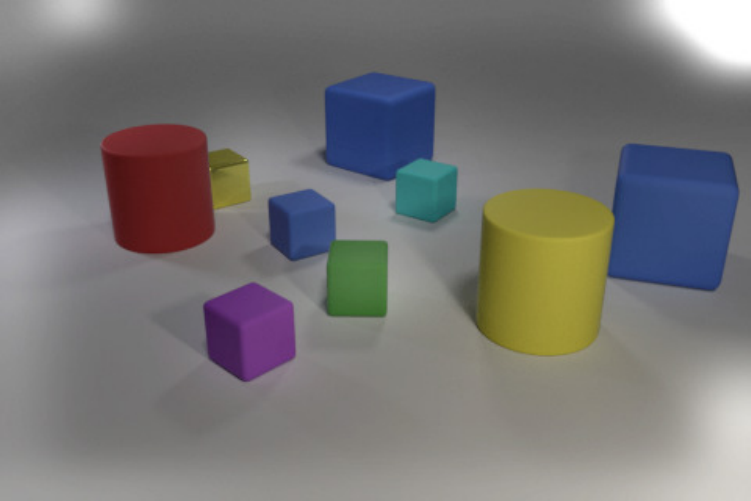}}
\hfill
\subfloat{\includegraphics[width=0.325\linewidth]{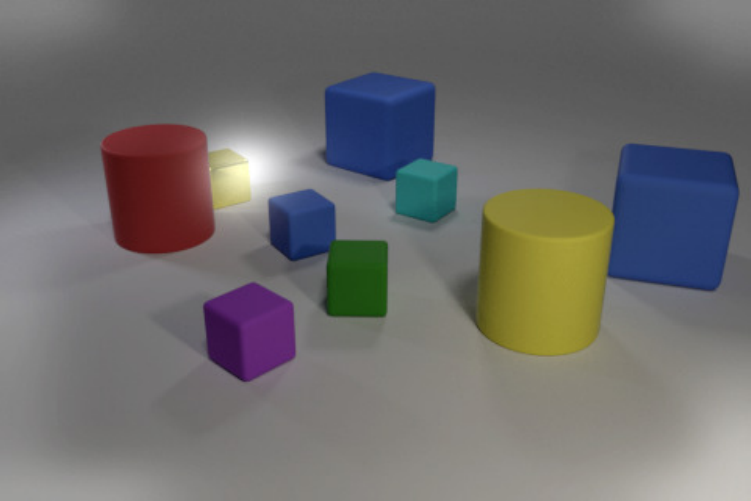}}
\hfill
\subfloat{\includegraphics[width=0.325\linewidth]{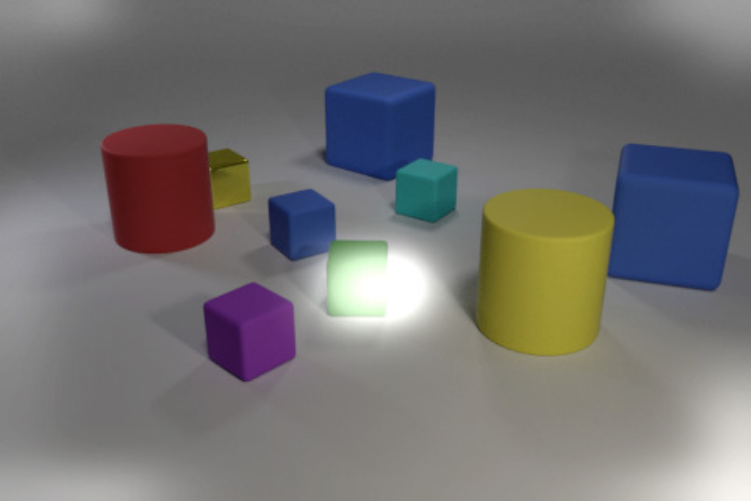}}
\hfill
\subfloat{\includegraphics[width=0.325\linewidth]{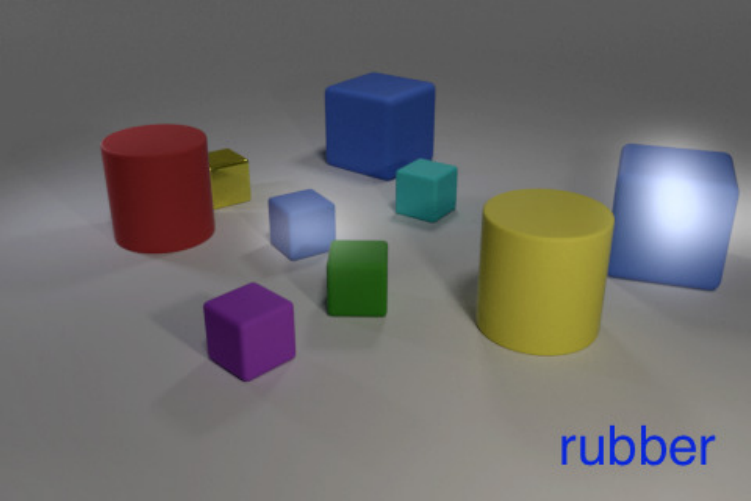}}
\end{minipage}

\begin{minipage}{0.2\textwidth}
\centering
\vspace*{4mm}
\subfloat{\includegraphics[width=1.0\linewidth]{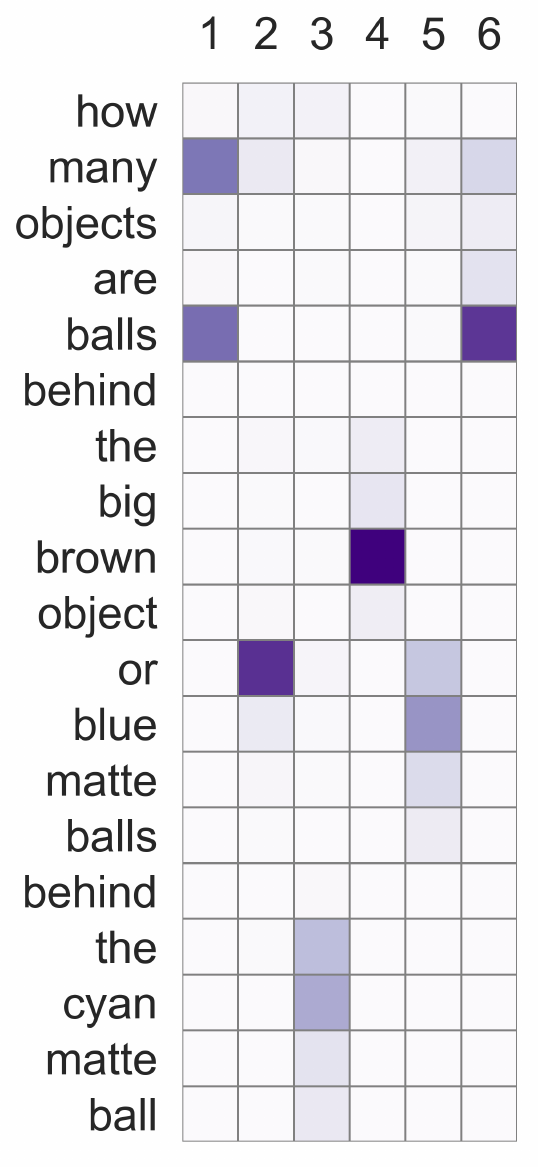}}
\end{minipage}
\begin{minipage}{0.79\textwidth}
\noindent
\centering
\subfloat{\includegraphics[width=0.325\linewidth]{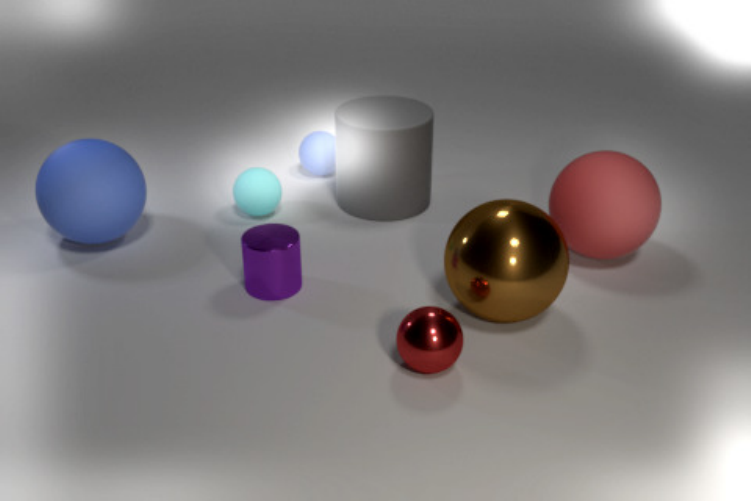}}
\hfill
\subfloat{\includegraphics[width=0.325\linewidth]{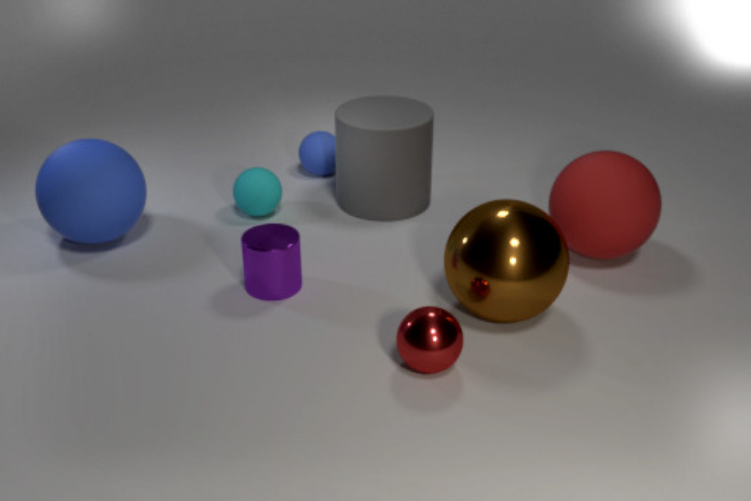}}
\hfill
\subfloat{\includegraphics[width=0.325\linewidth]{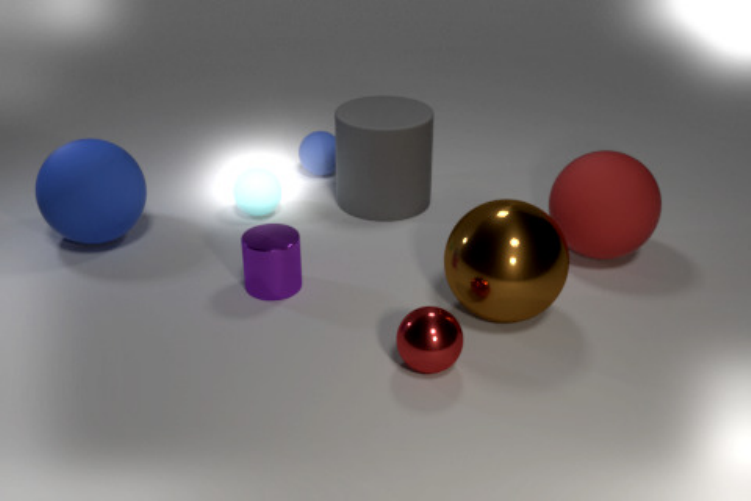}}
\hfill
\subfloat{\includegraphics[width=0.325\linewidth]{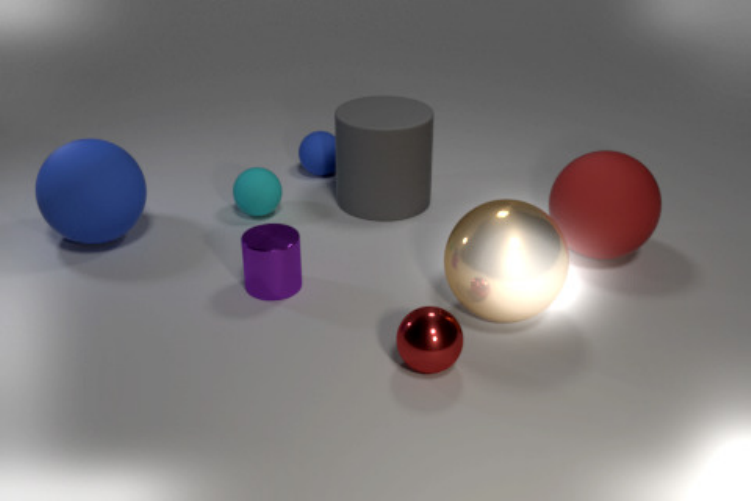}}
\hfill
\subfloat{\includegraphics[width=0.325\linewidth]{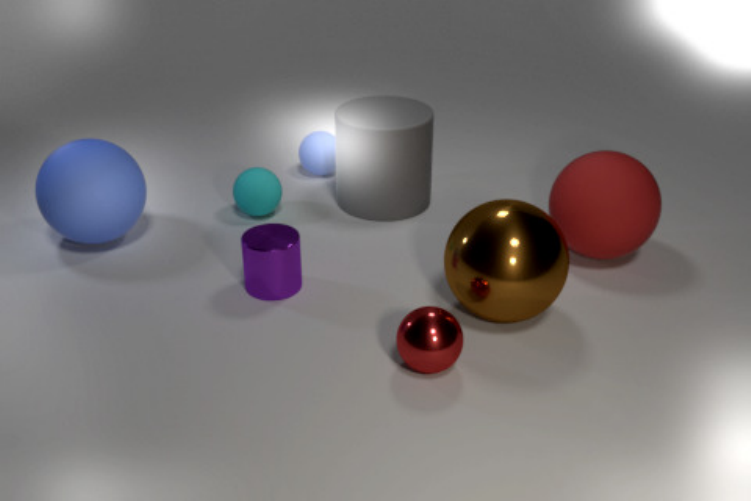}}
\hfill
\subfloat{\includegraphics[width=0.325\linewidth]{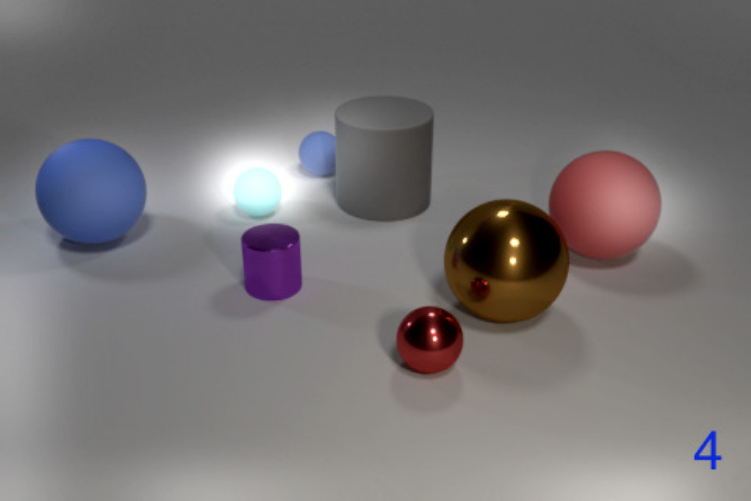}}
\end{minipage}

\begin{minipage}{0.2\textwidth}
\centering
\vspace*{4mm}
\subfloat{\includegraphics[width=0.75\linewidth]{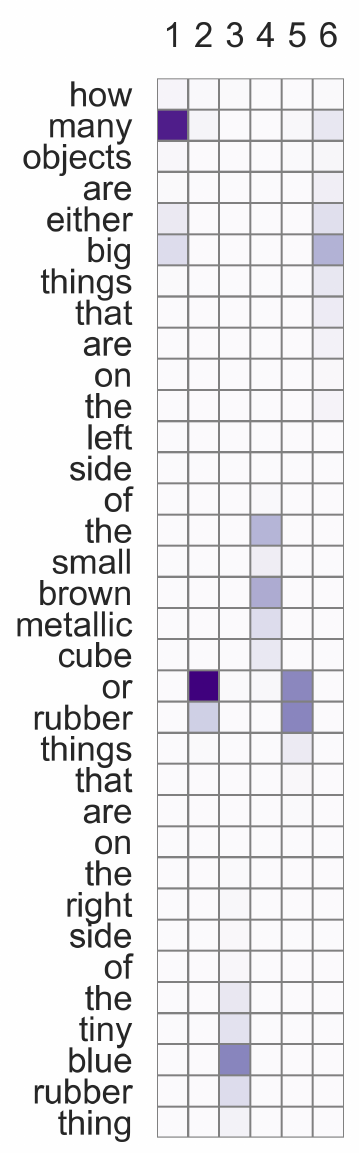}}
\end{minipage}
\begin{minipage}{0.79\textwidth}
\noindent
\centering
\subfloat{\includegraphics[width=0.325\linewidth]{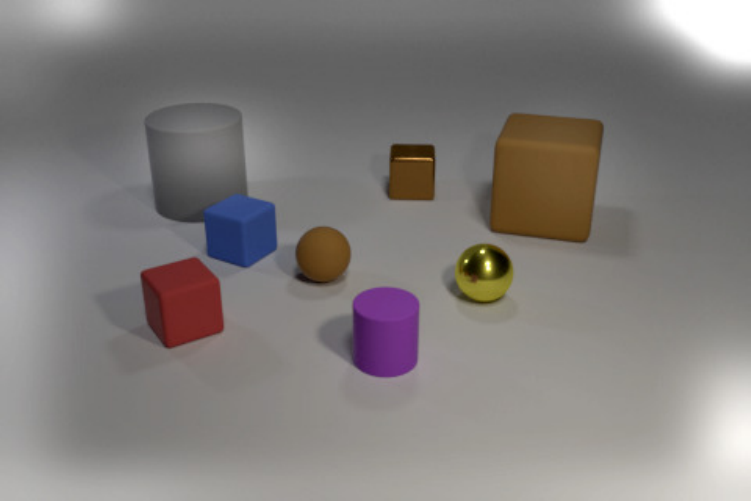}}
\hfill
\subfloat{\includegraphics[width=0.325\linewidth]{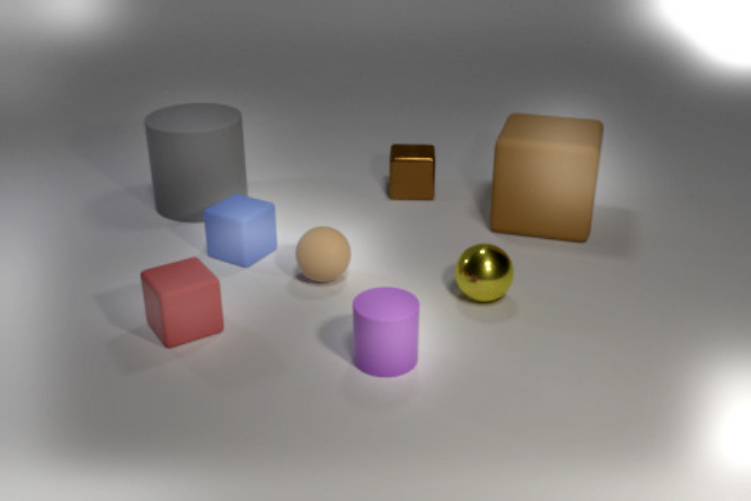}}
\hfill
\subfloat{\includegraphics[width=0.325\linewidth]{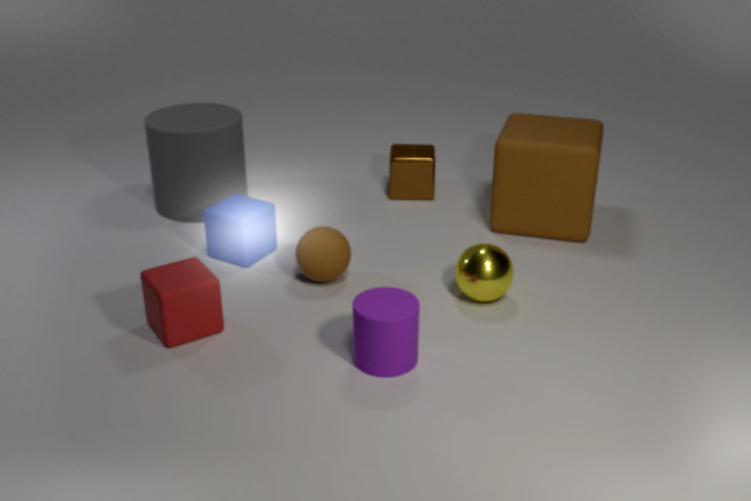}}
\hfill
\subfloat{\includegraphics[width=0.325\linewidth]{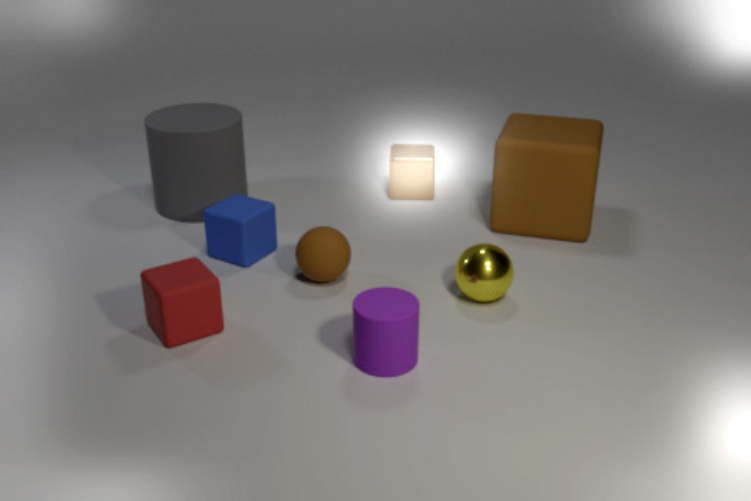}}
\hfill
\subfloat{\includegraphics[width=0.325\linewidth]{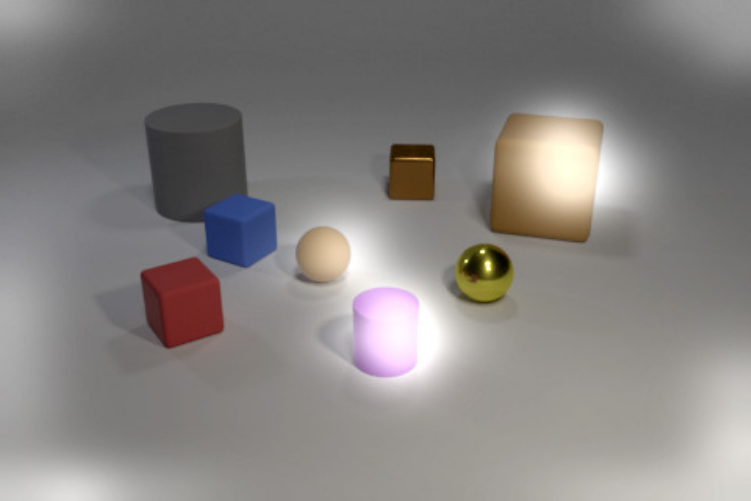}}
\hfill
\subfloat{\includegraphics[width=0.325\linewidth]{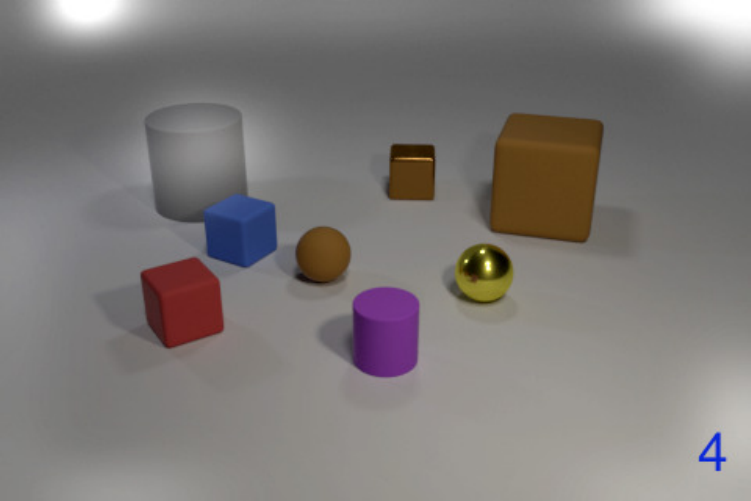}}

\end{minipage}

\caption{Attention maps produced by a MAC network of length 6, providing evidence for the ability of the model to track transitive relations, and perform logical operations, counting and summation. Note how the first iterations focus on the key structural question words ``many" and ``or" that serve as indicators for the model of the required reasoning operation it has to perform. Also note how the model correctly sums up two object groups in the second example, while correctly accounting for the intersection between them.}
\label{vizz2}
\end{figure}

\end{document}